\RequirePackage{fix-cm}
\documentclass[smallextended]{svjour3}       % onecolumn (second format)
\smartqed  % flush right qed marks, e.g. at end of proof
\usepackage[english]{babel}
\usepackage[utf8]{inputenc}
\usepackage{times} % assumes new font selection scheme installed
\usepackage{cite}
\usepackage{balance}

\usepackage{hyperref}
\hypersetup{
	colorlinks=true,
	linkcolor=black,
	urlcolor=blue
}

\usepackage[nolist]{acronym}

\usepackage{multirow}
\usepackage{booktabs}

\DeclareMathAlphabet{\pazocal}{OMS}{zplm}{m}{n}

\usepackage[binary-units]{siunitx}
\usepackage{graphicx} % for pdf, bitmapped graphics files
\usepackage{epsfig} % for postscript graphics files
\usepackage{pgfplots}
\usepackage{pgfplotstable}
\usepgfplotslibrary{statistics} % CRITICAL: This allows LaTeX to calculate the boxplot values
\pgfplotsset{compat=1.18}
\usepackage[font=small]{caption}
\usepackage[font=footnotesize]{subcaption}

\usepackage{float}

\usepackage[export]{adjustbox}

\usepackage{booktabs}

\usepackage{xcolor}
\definecolor{myYellow}{rgb}{0.93,0.69,0.13}
\definecolor{myPurple}{rgb}{0.49,0.18,0.56}
\definecolor{myGreen}{rgb}{0.26 0.72 0.54}
\definecolor{darkgreen}{rgb}{0.272, 0.50, 0.376}
\definecolor{lightgreen}{rgb}{0.585, 0.82, 0.647}

\colorlet{mydarkblue}{blue!30!black}

\usepackage{pifont}
\newcommand{\cmark}{\color{green}{\ding{51}}}%
\newcommand{\xmark}{\color{red}{\ding{55}}}%

\usepackage[markup=bfit, deletedmarkup=sout, authormarkup=brackets]{changes}
\definechangesauthor[name={Pavel S.}, color=blue]{PS}

\usepackage{mathtools}
\usepackage{nicefrac}
\usepackage{amsmath}
\usepackage{amssymb}
\usepackage{bm} % to bold symbols

\DeclareMathOperator*{\minimize}{minimize}

\DeclareMathOperator{\atantwo}{atan2}

\DeclareSIUnit{\litre}{l}

\usepackage{etoolbox}
\makeatletter%
\AfterPreamble{%
	\usepackage{hyperref}%
	\let\oldhypertarget\hypertarget%
	\renewcommand{\hypertarget}[2]{%
		\oldhypertarget{#1}{#2}%
		\protected@write\@mainaux{}{%
			\string\expandafter\string\gdef%
			\string\csname\string\detokenize{#1}\string\endcsname{#2}%
		}%
	}%
	\newcommand{\myhyperlink}[1]{%
		\hyperlink{#1}{\csname #1\endcsname}%
	}%
}
\makeatother%

\newcounter{Remark}
\newcounter{Definition}
\newcounter{Problem}
\newcounter{Example}
\usepackage{siunitx}
\usepackage{algorithm}
\usepackage[noend]{algpseudocode}

\makeatletter
\def\BState{\State\hskip-\ALG@thistlm}
\makeatother

\usepackage{tikz}
\pgfdeclarelayer{foreground}
\usepgfplotslibrary{groupplots}
\pgfsetlayers{background, main, foreground}
\usetikzlibrary{quotes, angles, backgrounds, arrows, automata, shapes, positioning, calc, through, spy, decorations.pathreplacing, decorations.markings, arrows.meta, 3d, perspective, petri, intersections, fillbetween, patterns, pgfplots.fillbetween}
\usepackage[outline]{contour} % halo around text
\contourlength{1.2pt}

\tikzset{
	imglabel/.style={
		rectangle,
		inner sep=2pt,
		text=black,
		minimum height=1em,
		text centered,
		fill=white,
		fill opacity=1.0,
		text opacity=1,
		anchor=south west,
	},
}
\tikzset{
	state/.style={
		rectangle,
		draw=black, very thick,
		minimum height=1.0em,
		text centered,
	},
	smallstate/.style={
		rectangle,
		draw=black, very thick,
		minimum height=0.2em,
		text centered,
	},
}
\tikzset{
	on each segment/.style={
		decorate,
		decoration={
			show path construction,
			moveto code={},
			lineto code={
				\path [#1]
				(\tikzinputsegmentfirst) -- (\tikzinputsegmentlast);
			},
			curveto code={
				\path [#1] (\tikzinputsegmentfirst)
				.. controls
				(\tikzinputsegmentsupporta) and (\tikzinputsegmentsupportb)
				..
				(\tikzinputsegmentlast);
			},
			closepath code={Acp
				\path [#1]
				(\tikzinputsegmentfirst) -- (\tikzinputsegmentlast);
			},
		},
	},
	mid arrow/.style={postaction={decorate,decoration={
				markings,
				mark=at position .5 with {\arrow[#1]{stealth}}
	}}},
}
\usepackage{scalerel}
\usepackage{academicons}
\usetikzlibrary{svg.path}
\definecolor{orcidlogocol}{HTML}{A6CE39}
\tikzset{
	orcidlogo/.pic={
		\fill[orcidlogocol] 
		svg{M256,128c0,70.7-57.3,128-128,128C57.3,256,0,198.7,0,128C0,57.3,57.3,0,128,0C198.7,0,256,57.3,256,128z};
		\fill[white] svg{M86.3,186.2H70.9V79.1h15.4v48.4V186.2z}
		svg{M108.9,79.1h41.6c39.6,0,57,28.3,57,53.6c0,27.5-21.5,53.6-56.8,53.6h-41.8V79.1z 
			M124.3,172.4h24.5c34.9,0,42.9-26.5,42.9-39.7c0-21.5-13.7-39.7-43.7-39.7h-23.7V172.4z}
		svg{M88.7,56.8c0,5.5-4.5,10.1-10.1,10.1c-5.6,0-10.1-4.6-10.1-10.1c0-5.6,4.5-10.1,10.1-10.1C84.2,46.7,88.7,51.3,88.7,56.8z};
	}
}
\newcommand\orcidicon[1]{\href{https://orcid.org/#1}{\mbox{\scalerel*{
				\begin{tikzpicture}[yscale=-1,transform shape]
					\pic{orcidlogo};
				\end{tikzpicture}
			}{|}}}}

\graphicspath{{./figures/}}
\usepackage{enumitem}

\begin{document}
	
	% === Title of the paper ==============
	\title{STL-Based Motion Planning and Uncertainty-Aware Risk Analysis for Human-Robot Collaboration with a Multi-Rotor Aerial Vehicle}
	
	\titlerunning{STL-Based Motion Planning and Uncertainty-Aware Risk Analysis for Human-Robot Collaboration with a Multi-Rotor Aerial Vehicle}        % if too long for running head
	
	% === Author list =====================
	\author{Giuseppe Silano$^{1,2\star}$ \and Amr Afifi$^{3}$ \and Martin Saska$^{1}$ \and Antonio Franchi$^{3,4}$
		% First Author         \and        Second Author %etc.
	}
	
	\institute{$^\star$Corresponding author: Giuseppe Silano \at
		$^1$Giuseppe Silano and Martin Saska are with the Department of Cybernetics, Czech Technical University in Prague, 12135 Prague, Czech Republic (emails: {\tt\footnotesize \{silangiu, martin.saska\}@fel.cvut.cz}).\\
		$^1$Giuseppe Silano is with the Department of Power Generation Technologies and Materials, Ricerca sul Sistema Energetico S.p.A., 20134 Milan, Italy (email: {\tt\footnotesize giuseppe.silano@rse-web.it}).\\
		$^3$Amr Afifi and Antonio Franchi are with the Robotics and Mechatronics Department, Electrical Engineering,  Mathematics, and Computer Science Faculty, University of Twente, 7500 AE Enschede, The Netherlands (emails: {\tt\footnotesize a.n.m.g.afifi@utwente.nl, schol@r-franchi.eu}). \\
		$^4$Antonio Franchi is with Department of Computer, Control and Management Engineering, Sapienza University of Rome, 00185 Rome, Italy (email:  {\tt\footnotesize schol@r-franchi.eu}). \\
	}
	
	\date{Received: April 20, 2025 / Accepted: 20 April 2026}
	
	\maketitle
	
	%%% START SECTION ==========================================================
	
	\begin{acronym}
		\acro{ACW}[ACW]{Aerial Co-Worker}
		\acro{AGM}[AGM]{Arithmetic-Geometric Mean}
		\acro{AR}[AR]{Aerial Robot}
		\acro{ASFM}[ASFM]{Aerial Social Force Model}
		\acro{CBF}[CBF]{Control Barrier Function}
		\acro{CDF}[CDF]{Cumulative Distribution Function}
		\acro{CoM}[CoM]{Center of Mass}
		\acro{DoF}[DoF]{Degree of Freedom}
		\acro{GTMR}[GTMR]{Generically-Tilted Multi-Rotor}
		\acro{HRI}[HRI]{Human-Robot Interaction}
		\acro{ILP}[ILP]{Integer Linear Programming}
		\acro{LSE}[LSE]{Log-Sum-Exponential}
		\acro{MILP}[MILP]{Mixed-Integer Linear Programming}
		\acro{MRAV}[MRAV]{Multi-Rotor Aerial Vehicle}
		\acro{NLP}[NLP]{Nonlinear Programming}
		\acro{NMPC}[NMPC]{Nonlinear Model Predictive Control}
		\acro{pHRI}[pHRI]{physical Human Robot Interaction}
		\acro{ROS}[ROS]{Robot Operating System}
		\acro{SITL}[SITL]{Software-In-The-Loop}
		\acro{STL}[STL]{Signal Temporal Logic}
		\acro{TL}[TL]{Temporal Logic}
		\acro{UAV}[UAV]{Unmanned Aerial Vehicle}
		\acro{UAM}[UAM]{Unmanned Aerial Manipulator}
		\acro{VRP}[VRP]{Vehicle Routing Problem}
		\acro{wrt}[w.r.t.]{with respect to}
	\end{acronym}
	
	%%% END SECTION ============================================================
	
	% Between 150 to 250 words.
	\begin{abstract}
		
		This paper presents a motion planning and risk analysis framework for enhancing human-robot collaboration with a \acl{MRAV}. The proposed method employs \acl{STL} to encode key mission objectives, including safety, temporal requirements, and human preferences, with particular emphasis on ergonomics and comfort. An optimization-based planner generates dynamically feasible trajectories while explicitly accounting for the vehicle's nonlinear dynamics and actuation constraints. To address the resulting non-convex and non-smooth optimization problem, smooth robustness approximations and gradient-based techniques are adopted. In addition, an uncertainty-aware risk analysis is introduced to quantify the likelihood of specification violations under human-pose uncertainty. A robustness-aware event-triggered replanning strategy further enables online recovery from disturbances and unforeseen events by preserving safety margins during execution. The framework is validated through MATLAB and Gazebo simulations on an object handover task inspired by power line maintenance scenarios. Results demonstrate the ability of the proposed method to achieve safe, efficient, and resilient human–robot collaboration under realistic operating conditions.
		
		
	\end{abstract}
	
	%%% END SECTION ============================================================
	
	%%% START SECTION ==========================================================
	
	\section{Introduction}
	\label{sec:introduction}
	
	\begin{sloppypar}
		
		\acp{AR}, and in particular \acfp{MRAV}, have attracted increasing attention due to their agility, maneuverability, and ability to accommodate a wide range of onboard sensors \cite{OlleroTRO2022, ShakhatrehAccess2019}. Their modular design enables applications such as contactless interactions \cite{KratkyRAL2025}, physical engagements with the environment \cite{Tognon2019RAL}, wireless communications \cite{LiceaComMag2025}, aerial filming \cite{KratkyRAL2021}, and surveillance and search-and-rescue missions \cite{PetracekRAL2021}. These platforms are particularly attractive for operations in hazardous or inaccessible environments, including high-altitude workspaces \cite{Afifi2022ICRA}, wind turbine maintenance \cite{Silano2024RAS}, construction sites \cite{Loianno2018IJRR}, and power line inspection \cite{CaballeroIEEEAccess2023}, where human intervention is costly and risky. 
		
		Deploying \acp{AR} as \textit{robotic co-workers} \cite{Haddadin2011Springer, Truc2022ICRA} in such scenarios can reduce physical and cognitive workload while improving operational safety \cite{TognonTRO2021, BenziRAL2022}. However, close-proximity aerial \ac{HRI} introduces stringent safety, comfort, and ergonomic requirements, which are further exacerbated by operations at height and in cluttered environments \cite{Wojciechowska2019ACM}. Effective collaboration therefore demands motion-planning strategies that explicitly go beyond geometric feasibility and encode safety, approach direction, comfort, timing, and interaction preferences while limiting operator burden. 
		
		Despite extensive work on collaborative \ac{HRI} with ground-based platforms, comparatively fewer studies address \acp{AR} \cite{Ajoudani2018AR}. Prior contributions have investigated robotic manipulation assistance \cite{Gienger2018IROS, Staub2018RAM} and handover behaviors \cite{Ortenzi2021TRO, HerdelCHI2022}.  
		For aerial systems operating near humans, however, planners must jointly reason about nonlinear dynamics, actuation limits, ergonomic constraints, and temporal mission requirements. These needs motivate the adoption of formal specification frameworks capable of encoding safety-, ergonomics-, and time-critical objectives while enabling quantitative assessment of their satisfaction under uncertainty. Among such formalisms, \ac{TL} has emerged as a powerful tool for robotic mission specification \cite{Belta2007RAM}. In particular, \acf{STL} \cite{maler2004FTMATFTS} introduces a quantitative \textit{robustness} metric that measures not only whether a specification is satisfied, but also by how much. This property enables the formulation of trajectory-optimization problems that maximize satisfaction margins (i.e., the \textit{robustness score}) while enforcing dynamic feasibility.
		
		Conventional \ac{AR} planners typically rely on waypoint-based \cite{Tan2021Access, Richter2016Polynomial} or geometric \cite{Corsini2022IROS, Wojciechowska2019ACM} formulations combined with low-level tracking controllers \cite{Mellinger2011ICRA}, and rarely encode comfort, approach direction, or timing through formal logic. Uncertainty is often treated through conservative margins rather than robustness- or risk-based analysis. Consequently, existing approaches in aerial \ac{HRI} tend to address only subsets of these challenges, such as safety or perception, without jointly considering formal temporal specifications, full nonlinear dynamics, ergonomics, and uncertainty.
		
		The present work departs from these approaches by integrating \ac{STL} specifications, full nonlinear \ac{MRAV} dynamics, collaboration-oriented constraints, uncertainty-aware risk metrics, and robustness-aware replanning within a unified optimization framework. This difference in problem formulation motivates the use of task-consistent performance indicators rather than direct numerical comparison with planners addressing fundamentally different objectives.
		
		A preliminary version of this work was presented at the 2023 International Conference on Unmanned Aircraft Systems (ICUAS’23) \cite{SilanoICUAS2023}. That conference paper focused on \ac{STL}-based motion planning for aerial human–robot handover using simplified spline-based trajectory representations \cite{MuellerTRO2015} and limited vehicle modeling. The present manuscript substantially advances that study by embedding full nonlinear \ac{MRAV} dynamics and actuator-level constraints into the \ac{STL}-constrained optimization problem, introducing energy-aware regularization and explicit robustness thresholds, expanding the \ac{STL} formulation to include comfort, visibility, and timing requirements, and proposing a systematic uncertainty-aware risk analysis based on Value-at-Risk (VaR) and Conditional-VaR (CVaR) metrics. In addition, a robustness-aware event-triggered replanning strategy is introduced, enforcing only safety-critical \ac{STL} predicates during recovery and quantitatively evaluating recovery behavior through robustness and risk measures. Finally, the experimental evaluation is significantly broadened through extensive numerical and physics-based simulation studies.
		
\begin{figure}[tb]
			\centering
			% left - bottom - right - top
			\adjincludegraphics[width=\columnwidth, trim={{0.1\width} {0.50\height} {0.1\width} {.315\height}}, clip]{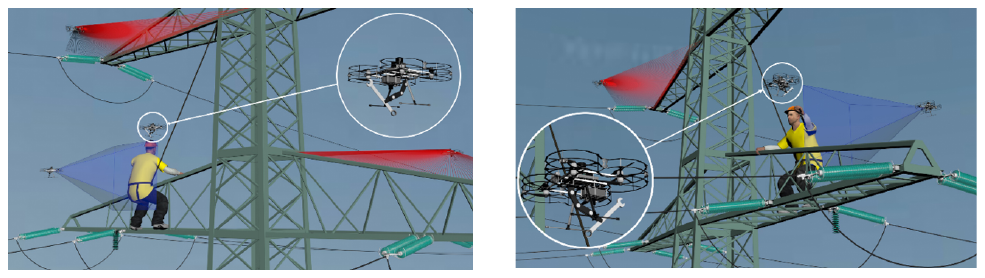}
			\vspace{-4.7em}
			\caption{Illustration of an \ac{MRAV} facilitating tool delivery to a human worker in a power line scenario. Courtesy of the AERIAL-CORE project\textsuperscript{1}.}
			\label{fig:handover_representation}
		\end{figure}

		In this paper, we present a motion planning framework that leverages \ac{STL} to encode collaborative missions involving humans, with a particular focus on ergonomic and comfortable interaction. Object handover \cite{Duan2024BIR} in a power line maintenance scenario is used as a motivating example (see Figure \ref{fig:handover_representation}). Mission requirements are expressed through \ac{STL} formulae and enforced by solving a nonlinear, non-smooth, and non-convex optimization problem that explicitly incorporates vehicle dynamics and actuation limits. The framework further integrates uncertainty-aware risk analysis and robustness-driven online recovery to ensure safe and repeatable execution in realistic collaborative settings.
		
	\end{sloppypar}
	
	%%% START SECTION ==========================================================
	
	\subsection{Related work}
	\label{sec:relatedWork}
	
	\begin{sloppypar}
		
		% Presenting the requirements for human-robot navigation and co-working as well as ergonomy
		To enable effective collaboration between \acp{AR} and humans in shared workspaces, motion planners must explicitly incorporate \ac{HRI} principles. As outlined in \cite{Sisbot2007TRO, Sisbot2007IROS, Cauchard2024ACM}, such planners address three core aspects: (i) \textit{safety}, ensuring that the robot's motion does not endanger nearby humans; (ii) \textit{reliability and effectiveness}, requiring compliance with vehicle dynamics and task completion objectives; and (iii) \textit{social acceptability}, which entails generating behaviors consistent with human preferences and social norms to promote smooth and predictable interactions. 
		
		These elements are crucial for developing a successful \ac{HRI} planner that fosters seamless collaboration and coexistence in shared spaces. For example, even in relatively straightforward tasks like object handovers, the \ac{AR} must be capable of \textit{reaching} the handover location and \textit{approaching} the human in a controlled, easily understandable, and comfortable manner. This approach enhances the human operator's comfort while promoting efficient and safe collaboration, ensuring that the robot's intentions are easily recognizable and acceptable to the operator. Ergonomics considerations are key to improving the overall interaction experience.
		
		The following review therefore focuses on work most closely related to aerial \ac{HRI}, human-aware motion planning, and control synthesis under \ac{TL} specifications, in order to position the proposed framework \ac{wrt} methods addressing similar system classes and objectives.
		
		% Here some of the SOTA solutions: no comfort, ergonomy and robot model
		While the field of aerial \ac{HRI} is growing \cite{Wojciechowska2019ACM, HerdelCHI2022}, many human-aware navigation planners are still rooted in proxemics-based criteria \cite{Kruse2013RAS, Rios-Martinez2015IJSR}. Some alternative methods include the application of \ac{ASFM}, which allow robots to approach humans safely \cite{Garrell2017IROS, Garrell2019IROS}. Additionally, numerous efforts focus on ensuring physical safety of \acp{AR} operating in close proximity to humans. Safety mechanisms include imposing constraints on motion planners \cite{Nageli2017ACMTG}, fine-tuning controller parameters \cite{Berkenkamp2016ICRA}, implementing control barrier functions \cite{Cortez2021ICRA}, or applying formal verification techniques \cite{Barbosa2021CFMATS}. In a related direction, van Waveren et al. \cite{VanWaveren2023HRI} address perceived safety in motion planning for human-drone interaction by proposing a parameterized \ac{CBF} that constrains the drone's maximum deceleration and minimum distance to the human, adapting its parameters to individual users' subjective safety ratings. While this work provides important evidence that physical safety alone does not imply perceived safety, it does not model full \ac{MRAV} actuation dynamics, does not encode ergonomic or temporal mission requirements through formal specifications, and provides no uncertainty-aware risk analysis. However, these solutions generally overlook comfort and ergonomics concerns and often simplify robot dynamics, both of which are critical in human-\ac{AR} collaboration frameworks, especially in high-risk environments. 
		
		% Something about human ergonomics and comfort
		Human ergonomics and comfort have been explored in several studies, particularly for ground-based robots. For instance, \cite{Sisbot2012TRO} introduced a manipulation planner that considers factors such as ergonomics and the human's field of view. Similarly, \cite{Peternel2017Humanoids} proposed a method for computing human joint torques based on a dynamic whole-body model, allowing ground-based manipulators to minimize human joint overload. More recently, \cite{Proia2025TASE} proposed a safety-compliant, ergonomic, and time-optimal trajectory planner for collaborative cobot arms, jointly optimizing ergonomics, safety, and cycle time within a single trajectory optimization problem. While directly relevant to the ergonomics thread of this paper, that work targets ground-based manipulators and does not consider formal temporal logic specifications, aerial vehicle dynamics, or probabilistic risk evaluation under human pose uncertainty. Extending to the aerial domain, \cite{Proia2025TSMC} introduce an integrated control framework for safe and ergonomic human-drone interaction in industrial warehouses via speed-and-separation monitoring; however, this approach does not employ \ac{STL} specifications, does not account for nonlinear \ac{MRAV} actuation dynamics, and lacks systematic risk analysis under human pose uncertainty. However, few of these studies \cite{Truc2022ICRA, Cauchard2024ACM} specifically address the unique challenges associated with \acp{AR} interacting with humans within a unified formal specification and risk-aware framework.
		
		% Something about perception
		In addition to addressing strictly planning aspects, research has also focused on enhancing perception systems for \acp{AR} to detect and respond to human collaborators. 
		For instance, \cite{Corsini2022IROS} proposed a \ac{NMPC} approach that integrates human ergonomics while also enforcing perception and actuation constraints. This system models the human collaborator within the \ac{NMPC} to predict future poses and optimize control actions accordingly. Another approach, detailed in \cite{Afifi2022ICRA}, formulated a quadratic programming problem to control an \ac{AR} interacting physically with a human. This framework used admittance control to prioritize ergonomics and safety while an interaction supervisor adjusted compliance based on predefined interaction zones around the human operator. 
		While these studies enable \acp{AR} to interact safely and ergonomically with humans, they often fail to address situations requiring the explicit specification of time-based requirements, comfort preferences, and complex mission objectives. For instance, in certain scenarios, human safety may require that the robot approaches slowly from the front rather than from the sides \cite{Wojciechowska2019ACM, Dautenhahn2006SIGART}, or adheres to specific behaviors like maintaining reduced speed \cite{Butler2001AR} or staying within a defined area for a set time before reaching the human \cite{Rubagotti2022RAS}. In such cases, expressing mission and comfort requirements through formal specification languages proves beneficial.
		
		% Signal Temporal Logic planners and risk-aware approaches
		In the field of robot controller synthesis based on formal specifications, significant advancements have been made. For instance, \cite{Kshirsagar2019IROS} presented a controller generated from high-level \ac{STL} specifications for human-robot handovers, ensuring the precise timing of each handover phase. This approach allows end-users to specify robot behaviors using high-level goals rather than low-level control parameters. However, this method does not address crucial aspects related to \acp{AR}, such as the dynamic feasibility of aerial platforms and the need to account for physical actuation constraints. Similarly, \cite{Webster2019ArXiv} employed probabilistic model checking to ensure safety and liveness compliance in human-robot handover tasks \cite{Alpern1985IPL}. However, this approach lacks the capacity to manage the continuous, nonlinear dynamics and actuation limits of \acp{AR}, which are essential for safe and stable operations in unpredictable environments. In addition, \cite{Li2014ICTACAS} proposed a formalism for human-in-the-loop control synthesis, developing a semi-autonomous controller based on \ac{TL} specifications. However, this approach does not explicitly address the unique ergonomic and comfort challenges involved in close-proximity \acp{HRI}. This limitation is especially critical in high-risk environments like power line maintenance, where \acp{AR} must work seamlessly with human operators while meeting strict safety and ergonomic standards.
		
		To the best of the authors' knowledge, this paper is the first to address the trajectory planning problem with a specific focus on \acp{MRAV} for ergonomic and comfortable \ac{HRI}. This approach leverages \ac{STL} specifications while fully accounting for the nonlinearity of the \ac{MRAV} model. Furthermore, a risk-aware framework is introduced to evaluate the risks associated with meeting complex system specifications and trajectory requirements under human pose uncertainty. An event-driven replanning strategy is also incorporated to manage unforeseen events and external disturbances, ensuring mission continuity. Table~\ref{tab:comparison} provides a comprehensive comparison between related works and the proposed framework, highlighting the inclusion of key features such as system dynamics, ergonomics, comfort, time requirements, safety, and physical actuation limits, while also explicitly reporting the main methodological paradigms adopted by each approach to improve readability and self-containment.
		
		\begin{table*}[tb]
			\centering
			\caption{Comparison of related approaches and the proposed framework. Main methodological paradigms and addressed features are reported: \textit{included} ({\cmark}) and \textit{not included} ({\xmark}).}
			\label{tab:comparison}
			\vspace{-1.2em}
			\scalebox{0.95}{
				\setlength{\tabcolsep}{3pt}
				\begin{tabular}{@{}l c c c c c c c@{}}
					\hline
					\multirow{2}{*}{\textbf{Ref.}} 
					& \multirow{2}{*}{\textbf{Technique}} 
					& \multicolumn{6}{c}{\textbf{Features}} \\ \cline{3-8}
					&  & $\begin{array}{c} \textbf{System} \\ \textbf{Dynamics} \end{array}$ 
					& \textbf{Ergonomy} 
					& \textbf{Comfort} 
					& $\begin{array}{c} \textbf{Time} \\ \textbf{Req.} \end{array}$ 
					& \textbf{Safety} 
					& $\begin{array}{c} \textbf{Physical} \\ \textbf{Limits} \end{array}$ \\ 
					\cite{Kruse2013RAS} & Proxemics-based navigation & {\xmark} & {\xmark} & {\cmark} & {\xmark} & {\cmark} & {\xmark} \\
					\cite{Garrell2017IROS} & Artificial social force model & {\xmark} & {\xmark} & {\cmark} & {\xmark} & {\cmark} & {\xmark} \\
					\cite{Berkenkamp2016ICRA} & Safe learning / barrier methods & {\xmark} & {\xmark} & {\xmark} & {\xmark} & {\cmark} & {\xmark} \\
					\cite{Barbosa2021CFMATS} & Formal verification & {\xmark} & {\xmark} & {\xmark} & {\cmark} & {\cmark} & {\xmark} \\
					\cite{Sisbot2012TRO, Peternel2017Humanoids} & Ergonomic Optimization Problem (OP) & {\xmark} & {\cmark} & {\cmark} & {\xmark} & {\cmark} & {\xmark} \\
					\cite{Afifi2022ICRA, Corsini2022IROS} & \ac{NMPC} with human modeling & {\cmark} & {\cmark} & {\xmark} & {\xmark} & {\cmark} & {\cmark} \\
					\cite{Kshirsagar2019IROS} & \ac{STL}-based synthesis & {\xmark} & {\xmark} & {\cmark} & {\cmark} & {\cmark} & {\xmark} \\
					\cite{Webster2019ArXiv} & Probabilistic model checking & {\xmark} & {\xmark} & {\xmark} & {\cmark} & {\cmark} & {\xmark} \\
					\cite{VanWaveren2023HRI} & \ac{CBF}-based motion planning & {\xmark} & {\xmark} & {\cmark} & {\xmark} & {\cmark} & {\xmark} \\
					\cite{Proia2025TASE} & Ergonomic trajectory OP (cobot) & {\xmark} & {\cmark} & {\xmark} & {\cmark} & {\cmark} & {\cmark} \\
					\cite{Proia2025TSMC} & Speed-sep. monitoring (drone) & {\xmark} & {\cmark} & {\cmark} & {\xmark} & {\cmark} & {\xmark} \\
					\textbf{Ours} & STL + nonlinear OP + risk analysis & {\cmark} & {\cmark} & {\cmark} & {\cmark} & {\cmark} & {\cmark} \\ 
					\hline
			\end{tabular}}
		\end{table*}
		
	\end{sloppypar}
	
	%%% END SECTION ============================================================
	
	%%% START SECTION ==========================================================
	
	\subsection{Contributions}
	\label{sec:contributions}
	
	\begin{sloppypar}
		
		This paper introduces a novel motion planning framework for \ac{HRI} involving \acp{MRAV}, with a primary focus on enhancing ergonomics and operator comfort. The proposed approach leverages \ac{STL} to formally encode mission objectives, encompassing safety, temporal requirements, and human preferences. By utilizing the expressive capabilities of \ac{STL}, the proposed method ensures that key aspects of \ac{HRI} are systematically addressed. An optimization problem is formulated to generate dynamically feasible trajectories that satisfy these specifications while accounting for physical actuation limits and dynamics constraints of the aerial platform. Solving this problem requires addressing a complex nonlinear, non-smooth, and non-convex optimization challenge. To tackle these complexities, we employ smooth approximations, enabling the use of gradient-based optimization techniques. The effectiveness of this approach is demonstrated through an illustrative task: object handovers by an \ac{MRAV} in a power line maintenance scenario.
		
		This work builds directly upon our preliminary conference paper \cite{SilanoICUAS2023}, which presented an early version of the aerial human–robot handover problem using \ac{STL}-based planning with simplified trajectory representations and limited vehicle modeling. The present manuscript substantially extends that formulation and is also informed by our broader line of research on \ac{STL}-based \ac{MRAV} task assignment and trajectory optimization \cite{SilanoICUAS2023, SilanoRAL2021, CaballeroIEEEAccess2023, Silano2024RAS}. Those earlier works addressed infrastructure inspection \cite{SilanoRAL2021}, payload deployment \cite{CaballeroIEEEAccess2023}, recharging logistics, and heterogeneous timing constraints \cite{Silano2024RAS}, but did not consider close-proximity \ac{HRI} or object handover scenarios. The present manuscript extends that research line to explicitly account for \ac{HRI} requirements, including ergonomic and comfort-oriented specifications, uncertainty in human pose, and robustness-aware online recovery within \ac{STL}-constrained missions. The formulation also embeds full nonlinear \ac{MRAV} dynamics and actuator-level constraints and supports both underactuated and fully actuated platforms \cite{HamandiIJRR2021}.
		
		The main contributions of this paper can be summarized as follows:
		
		\begin{itemize}
			
			\item The problem of object handovers in a power line maintenance scenario, serving as a motivating example for human-robot collaboration using an \ac{MRAV}, is formulated in Section \ref{sec:motivatingExample}. Mission specifications for this problem are established in Section \ref{sec:specificationEncoding}, and the corresponding \ac{STL} formula is derived. An optimization problem is then formulated to determine dynamically feasible trajectories that satisfy safety constraints, temporal requirements, and human preferences while adhering to the \ac{STL} formula (see Section \ref{sec:motionPlanner}).
			
			\item The proposed \ac{STL} optimization problem (see Section \ref{sec:motionPlanner}) seeks  globally optimal solutions but is inherently nonlinear, non-smooth, and non-convex, presenting computational challenges \cite{Gilpin2021LCSS, LindemannAutomatica2019}. To address  these complexities, smooth approximations (see Section \ref{sec:smoothApproximation}) are introduced, enabling the use of gradient-based optimization techniques. While gradient-based methods can be sensitive to local optima \cite{Bertsekas2012Book}, various practical strategies exist to mitigate this issue, as explored in previous work \cite{CaballeroIEEEAccess2023, Silano2024RAS}. Additionally, an energy minimization term is incorporated to implicitly extend the \ac{MRAV}'s endurance during the mission, contributing to overall efficiency (see Section \ref{sec:comparativeAnalysis}).
			
			\item The proposed method includes a risk-aware analysis (see Section \ref{sec:uncertaintyAwareRiskAnalysis}) that considers uncertainties in human pose. This analysis provides a systematic approach to assess and quantify the risks associated with potential deviations from \ac{STL} specifications in the obtained trajectories.
			It enables the determination of whether a specified success rate for meeting the given \ac{STL} specifications (e.g., 80\%) can be reliably achieved or not under uncertain conditions.
			
			\item A robustness-aware triggered replanning strategy (see Section \ref{sec:replanner}) is introduced to handle disturbances and unforeseen events during the mission. This approach reshapes the optimization problem to compute a feasible trajectory that reconnects the drone to the previously computed optimal offline trajectory.
			
			\item Numerical simulations in MATLAB (see Section \ref{sec:simulationResults}) evaluate the method's overall performance in terms of mission specification fulfillment. Additionally, Gazebo simulations (see Section \ref{sec:gazeboSimulations}), conducted in a mock-up setting, demonstrate the method's validity and feasibility under conditions resembling real-world scenarios. Conclusions are presented in Section \ref{sec:conclusions}.
			
		\end{itemize}
		
	\end{sloppypar}
	
	%%% END SECTION ============================================================
	
	%%% START SECTION ==========================================================
	
	\section{Problem Setup and Objectives}
	\label{sec:motivatingExample}
	
	This section formalizes the problem addressed in this paper and outlines the overall objectives pursued by the proposed framework. We first state the high-level goal of enabling safe, ergonomic, and comfortable human–aerial robot collaboration under dynamic and actuation constraints (see Section \ref{sec:objective}), and then introduce the power line handover scenario used throughout the paper as a motivating example (see Section \ref{sec:scenarioDescription}).
	
	%%% END SECTION ============================================================
	
	%%% START SECTION ==========================================================
	
	\subsection{Objective}
	\label{sec:objective}
	
	The objective of this work is to improve ergonomic and comfortable collaboration between humans and \acp{AR} in high-risk environments, with a particular focus on high-altitude maintenance operations. This research was conducted within the AERIAL-CORE European project\footnote{\url{https://aerial-core.eu}}, motivated by the high incidence of safety regulation violations during maintenance of electric power transmission infrastructures and the associated risks for human operators.
	
	Specifically, the goal is to compute dynamically feasible \ac{MRAV} trajectories that complete collaborative tasks within prescribed time windows while satisfying safety constraints, physical actuation limits, and human-centered requirements. Safety requirements include workspace containment, obstacle avoidance, and prohibiting approaches from behind the operator. The environment is assumed to be known in advance through a geometric map providing polyhedral representations of obstacles and restricted regions.
	
	%%% END SECTION ============================================================
	
	%%% START SECTION ==========================================================

	\subsection{Motivating Example}
	\label{sec:scenarioDescription}
	
	The motivating scenario considered in this paper involves an \ac{MRAV} equipped with a rigidly attached stick carrying a tool, tasked with performing repetitive object handovers to a human operator during power line maintenance (see Figure~\ref{fig:handover_representation}). Because the same maneuver is executed multiple times under similar environmental conditions, the task is well suited to offline planning, enabling optimization of ergonomic and comfort requirements while explicitly accounting for vehicle dynamics and actuation limits.
	
	Variations in human pose between successive deliveries motivate the inclusion of robustness and risk analysis to evaluate how deviations affect safety margins and mission satisfaction. The rigid stick attachment eliminates pendulum effects, simplifying the system dynamics during flight. Although power line maintenance serves as the primary example, the same principles apply to a broad class of collaborative aerial manipulation tasks in hazardous environments \cite{Gienger2018IROS}.
	
	In such handover operations, the final configuration must satisfy multiple human-centered criteria, including \textit{safety}, \textit{visibility}, \textit{ergonomics}, and \textit{comfort}, as outlined in prior studies \cite{Sisbot2007TRO, Sisbot2007IROS}. The \ac{MRAV} may approach the operator from different directions -- front, lateral, above, or below -- according to operator preferences \cite{Wojciechowska2019ACM}. To promote comfort and predictability, approach velocities are constrained near the operator \cite{Butler2001AR, Rubagotti2022RAS}.
	
	The handover is modeled within a three-dimensional workspace in which the \ac{MRAV} starts the mission already carrying the tool, and only one object is delivered at a time. Visibility requirements impose that the vehicle first reaches a designated location in front of the operator and remains there for a prescribed duration before initiating the handover. Once the \ac{MRAV} reaches the operator, the physical exchange is handled by a low-level onboard controller, following the approach developed in \cite{Corsini2022IROS, Afifi2022ICRA}. 
	
	The \ac{MRAV} is subject to physical actuation limits, including bounds on propeller speed, rotor-generated forces, and torques, which directly influence stability and human comfort. Limiting airflow and noise produced by the propellers is therefore essential during close-proximity interaction \cite{Cauchard2015ACM}. 
	
	\begin{figure}[tb]
		\centering
		\adjincludegraphics[width=\columnwidth, trim={{0.07\width} {0.08\height} {0.11\width} {.1\height}}, clip]{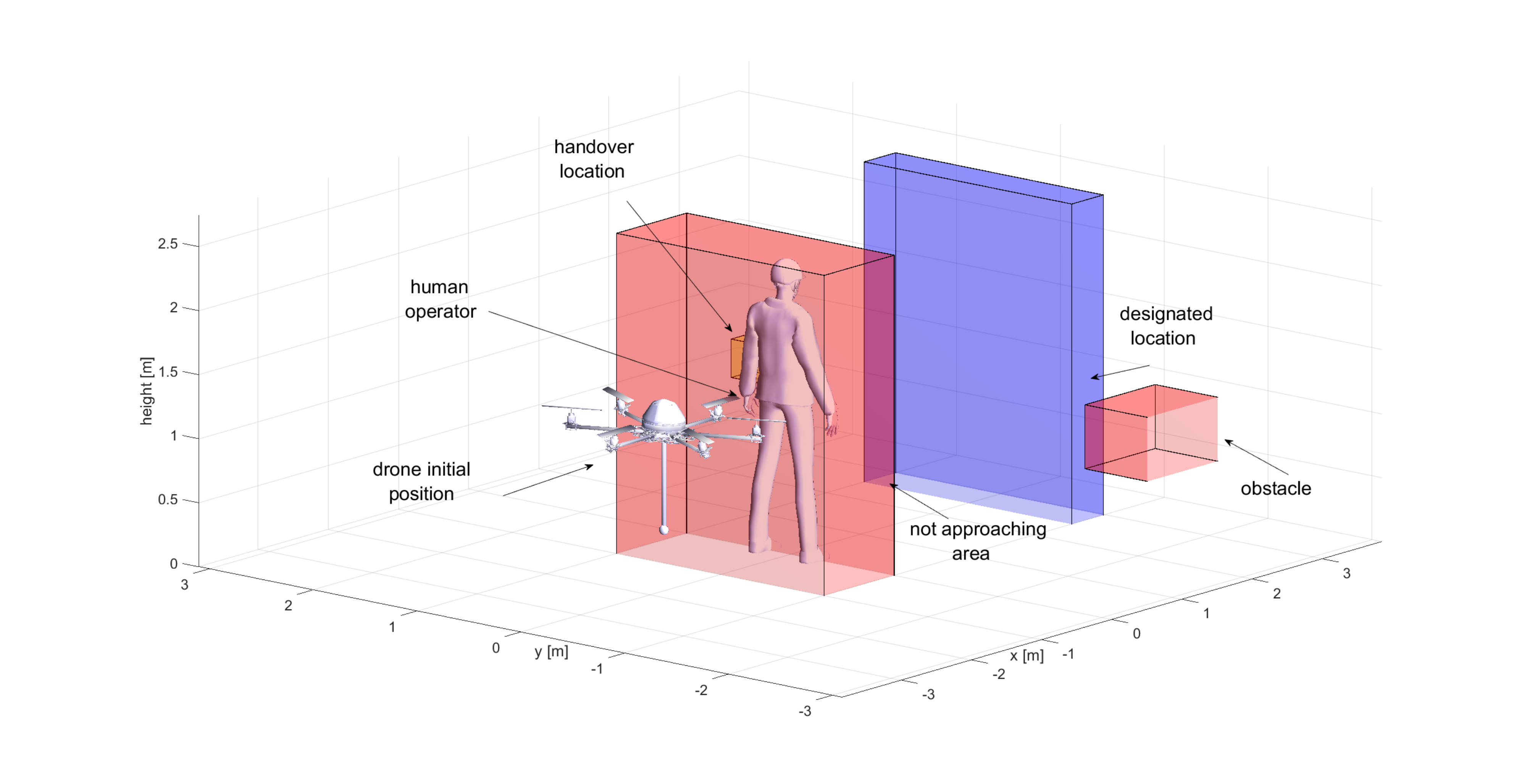}
		\vspace{-2.25em}
		\caption{Schematic depiction of the object handover scenario, highlighting the operator's preferred handover region (yellow), the designated location (blue), and obstacles along with the restricted area behind the operator (red).}
		\label{fig:handover_scenario}
	\end{figure}
	
	Although Figure~\ref{fig:handover_scenario} abstracts away the specific geometry of power transmission lines, it captures the essential components of the collaborative task and the associated constraints. The same structure applies to realistic maintenance settings, illustrating the generality of the proposed formulation for close-proximity human–robot collaboration in hazardous environments.
	
	%%% END SECTION ============================================================
	
	%%% START SECTION ==========================================================
	
	\section{Technical Preliminaries}
	\label{sec:preliminaries}
	
	This section introduces the technical preliminaries necessary to understand the contributions of this paper. It provides an overview of the fundamental concepts related to system dynamics, signal temporal logic, stochastic processes, and risk measures, which form the basis of the proposed approach. Additionally, to facilitate understanding and improve readability, the notation used throughout the paper is summarized in Table \ref{tab:tableOfNotation}.
	
	\begin{table}[tb]
		\centering
		\caption{Notation - System variables, general symbols and reference frames.}
		\vspace{-1.2em}
		\label{tab:tableOfNotation}
		\begin{tabular}{r p{7.25cm}}
			\toprule
			$\mathcal{F}_W$, $\mathcal{F}_B$, $\mathbf{p}$, $\mathbf{v}$, $\bm{\eta}$, $\bm{\omega}$ & reference frames; MRAV position, velocity, orientation, and angular velocity \\
			$\mathbf{p}_0$, $\bm{\eta}_0$, $N_p$ & \ac{MRAV} initial position and orientation, number of actuators \\
			$m$, $\mathbf{J}$, $\mathbf{g}$ & \ac{MRAV} mass and inertia tensor, and gravity vector \\
			$c_{\xi_i}$, $c_{\tau_i}$, $\mathbf{p}_{m_i}$ & force and torque constant parameters related to the $i$-th propeller's design, and $i$-th motor position expressed in $\mathcal{F}_B$  \\
			$\bm{\xi}$, $\bm{\tau}$, $\dot{\bm{\xi}}$, $\Omega_i$, $\mathbf{z}_{P_i}$ & \ac{MRAV} motor forces and torques, force derivatives, squared motor speed, and unit axis vector of the $i$-th propeller \\
			$\mathbf{T}$, $\mathbf{R}$, $\mathbf{G}$ & Jacobian matrix mapping $\bm{\omega}$ to $\dot{\bm{\eta}}$, rotation matrix mapping from 
			$\mathcal{F}_B$ to $\mathcal{F}_W$, and force/torque allocation matrix \\
			$\mathbf{p}_\mathrm{hum}$, $\bm{\eta}_\mathrm{hum}$ & human operator position and orientation \\
			$\mathbf{t}$, $N$, $T_s$, $\bullet_k$ & time vector, number of samples, sampling period, generic $k$-th element of a sequence \\
			$\pi$, $I$, $p_i$, $\mu_i$, $\lambda$ & \acs{STL} formula, generic time interval, $i$-th predicate and its real-valued function, tunable parameter for $\tilde{\rho}_\pi(\mathbf{x})$ \\
			$\neg$, $\wedge$, $\vee$, $\implies$, $\lozenge$, $\square$, $\pazocal{U}$, $\bigcirc$ & Boolean and temporal \acs{STL} operators: negation, conjunction, disjunction, implication, eventually, always, until, and next \\
			M, AP & set of real-valued functions and the corresponding predicates \\
			$\rho_\pi(\mathbf{x})$, $\tilde{\rho}_\pi(\mathbf{x})$ & robustness and smooth robustness values of the \ac{STL} formula $\pi$ \\
			$\rho_\pi(\mathbf{x}, Y(\cdot, \varepsilon))$, $\tilde{\rho}_\pi(\mathbf{x}, Y(\cdot, \varepsilon))$ & robustness and smooth robustness values of the \ac{STL} formula $\pi$ considering the realization of the stochastic process $Y$ \\
			$\Sigma$, $\mathcal{G}$, $P$ & sample space, $\sigma$-algebra of $\Sigma$, and probability measure \\
			$Z$, $\varepsilon$, $F_Z$, $Y$, $Y(\cdot,\varepsilon)$ & random variable, sample realization, cumulative distribution, stochastic process and its realization \\    
			$\mathrm{VaR}_\beta(Z)$, $\mathrm{CVaR}_\beta(Z)$ & $\beta$-Value-at-Risk and Conditional $\beta$-Value-at-Risk \\
			$\pi_\mathrm{ws}$, $\pi_\mathrm{obs}$, $\pi_\mathrm{beh}$ & \ac{STL} safety requirements \\
			$\pi_\mathrm{vr}$, $\pi_\mathrm{vis}$, $\pi_\mathrm{pr}$ & \ac{STL} visibility and ergonomic requirements \\
			$\pi_\mathrm{vel}$, $\pi_\mathrm{pro}$, $\pi_\mathrm{ho}$ & \ac{STL} comfort and mission requirements \\
			$T_N$, $T_\mathrm{vr}$, $N_\mathrm{obs}$, $N_\mathrm{pr}$ & mission duration and visibility intervals, number of obstacles, and number of preference regions \\
			$\underline{\mathrm{\Gamma}}_\mathrm{vel}$, $\bar{\mathrm{\Gamma}}_\mathrm{vel}$ $\underline{\mathrm{\Gamma}}_\mathrm{pro}$, $\bar{\mathrm{\Gamma}}_\mathrm{pro}$, $\gamma$ & bounds on linear velocity, propeller speed, and heading margin \\ 
			$\underline{p}^{(j)}_{\bullet}$, $\bar{p}^{(j)}_{\bullet}$ & generic vertices of the rectangular regions defining safety, ergonomic, comfort, and mission requirements \\
			$\mathcal{L}(\mathbf{x})$, $w$, $\kappa$ & energy term and relative weight, and robustness threshold \\
			$\bar{\bm{\xi}}$, $\underline{\bm{\xi}}$ &  maximum and minimum values for the \ac{MRAV} motor forces \\
			$\mu_{z}$, $\sigma_{z}$, $\mathbb{I}$, $K$, $\delta$ & mean and covariance of the normal distribution, indicator function, number of realizations, and level of confidence \\
			$\underline{\mathrm{VaR}}_\beta$, $\overline{\mathrm{VaR}}_\beta$, $\underline{\mathrm{CVaR}}_\beta$, $\overline{\mathrm{CVaR}}_\beta$ & lower and upper bounds of $\mathrm{VaR}_\beta$ and $\mathrm{CVaR}_\beta$ \\
			$T_e$, $T_g$, $\bar{\mathbf{t}}$, $\hat{\mathbf{t}}$, $t_c$ & event-driven period, ``waypoint'' period, corresponding time vectors, and maximum expected computation time for replanning\\
			$\zeta$, $\tilde{\mathbf{p}}$, $\mathbf{p}^\star$ & replanning threshold, actual and optimal \ac{MRAV} positions \\
			$\pi_\mathrm{crit}$, $\pi_\mathrm{soft}$ & safety-critical and soft subsets of the STL specification $\pi$ \\
			$\hat{\rho}_{\pi_\mathrm{crit}}$, $\kappa_\mathrm{crit}$ & robustness surrogate and minimum safety margin for the critical STL subset \\
			$\pazocal{I}_\mathrm{crit}$, $\mathbf{a}_{ij}$, $b_{ij}$ & index set of safety-critical predicates, half-space parameters defining polyhedral safety regions for predicate $i$ \\
			$w_r$, $w_u$ & weights on state deviation and control effort \\
			\bottomrule
		\end{tabular}
	\end{table}

	%%% END SECTION ============================================================
	
	%%% START SECTION ==========================================================
	
	\subsection{System modeling}
	\label{sec:systemModeling}
	
	%%% SYSTEM DEFINITION
	Let us consider the discrete-time dynamical model of the \ac{MRAV}, described as a \ac{GTMR} system \cite{Afifi2022ICRA, Corsini2022IROS}, expressed in the general form $x_{k+1} = f(x_k, u_k)$. Here, $x_{k+1}, x_k \in \mathcal{X} \subset \mathbb{R}^n$ represent the next and current states of the system at time step $k$, respectively, and $u_k \in \mathcal{U} \subset \mathbb{R}^m$ is the control input. The system is actuated by $N_p$ motor-propeller units, and its dynamics are derived using the Newton-Euler formalism. These actuators are placed arbitrarily and oriented \ac{wrt} the vehicle's main body. The number of actuators and their orientation \ac{wrt} the vehicle body determines whether the system is an underactuated or a fully-actuated platform \cite{HamandiIJRR2021}.
	
\begin{figure}[tb]
		\vspace*{-0.75em}
		\centering
		\scalebox{0.875}{
			\begin{tikzpicture}[scale=2.0, line cap=round, line join=round, >=Triangle]
				%% ELLIPSE
				\draw[color=black!60, rotate around={-110:(0,0)}, fill=darkgreen!80, line width=1pt] (0,0) ellipse (0.95cm and 1.50cm);
				\draw[color=black!60, rotate around={-20:(0,0)}, fill=lightgreen!80, dashed, line width=1.5pt] (0,0) ellipse (1.5cm and 0.75cm);
				
				%% CENTER OF MASS
				\fill (0,0) -- ++(0.2em,0) arc [start angle=0, end angle=90, radius=0.2em] -- ++(0,-0.4em) arc [start angle=270, end angle=180, radius=0.2em];
				\draw (0,0) [radius=0.2em] circle;
				
				%% BODY FRAME
				\draw (-0.05,0) node[left]{$O_B$}; % O_W
				\draw [->] (0,0) -- ({0.5*cos(65)},{0.5*sin(65)}) node[left]{$\mathbf{z}_B$}; % z_B
				\draw [->] (0,0) -- ({0.5*cos(20)},{0.5*sin(20)}) node[above]{$\mathbf{y}_B$}; % y_B
				\draw [->] (0,0) -- ({0.5*cos(-25)},{0.5*sin(-25)}) node[below]{$\mathbf{x}_B$}; % x_B
				
				%% GLOBAL FRAME
				\draw (-1.5,-1.5) node[left]{$O_W$}; % O_W
				\draw [->] (-1.5,-1.5) -- (-1.5,-1.0) node[left]{$\mathbf{z}_W$}; % z_W
				\draw [->] (-1.5,-1.5) -- (-1.15,-1.25) node[right]{$\mathbf{y}_W$}; % y_W
				\draw [->] (-1.5,-1.5) -- (-1.0,-1.5) node[below]{$\mathbf{x}_W$}; % x_W
				
				%% POSITION AND ROTATION VECTORS
				\draw[->, red] (-1.5,-1.5) -- node[above]{$\mathbf{p}$} (0,0);
				\draw[->, red] (-1.0,-1.75) to [out=-20, in=-70] node[left]{$\mathbf{R}$} (0.15, -0.15);
				
				%% MOTORS
				\tikzset
				{%
					pics/cylinder/.style n args={3}{% #1 = radius, #2 = height, #3 = angle
						code={%
							\draw[pic actions] (135:#1) arc (135:315:#1) --++ (0,0,#2) arc (315:135:#1) -- cycle;
							\draw[pic actions] (0,0,#2) circle (#1);
							\foreach\z in {0,1}
							{
								\begin{scope}[canvas is xy plane at z=\z*\h]
									\coordinate (-cen\z) at       (0,0);
									\coordinate (-ESE\z) at    (-#3:#1);
									\coordinate (-ENE\z) at     (#3:#1);
									\coordinate (-NNE\z) at  (90-#3:#1);
									\coordinate (-NNW\z) at  (90+#3:#1);
									\coordinate (-WNW\z) at (180-#3:#1);
									\coordinate (-WSW\z) at (180+#3:#1);
									\coordinate (-SSW\z) at (270-#3:#1);
									\coordinate (-SSE\z) at (270+#3:#1);
								\end{scope}
							}
					}},
				}
				\def\l{2}    % boom length
				\def\R{1}    % body radius
				\def\r{0.15} % motor radius
				\def\h{0.3}  % boom and body height
				%%%% PROPELLER 1
				\pic[fill=gray!30, rotate=-15, shift={(0,0,-\h)}] at (1.25,0.10) {cylinder={\r}{2*\h}{45}};
				\draw[->] (1.25,0.1225) -- (1.55,0.30) node[above]{$\mathbf{z}_P$};
				% propellers
				\begin{scope}[shift={(1.405,0.25)},rotate around z=120,canvas is xy plane at z=\h]
					\draw[fill=gray!30] (0.0,0) sin  (0.25,0.1) cos  (0.65,0) sin  (0.25,-0.1) cos (0,0)sin (-0.25,0.1) cos (-0.65,0) sin (-0.25,-0.1) cos (0,0);
					\fill (0,0) circle (0.05);
					\draw[latex-] (0.115,-0.175) arc (020:160:0.1) node [below] {$\tau$};
				\end{scope}
				%%%% PROPELLER 2
				\pic[fill=gray!30, rotate=175, shift={(0,0,-\h)}] at (-1.25,-0.10) {cylinder={\r}{2*\h}{45}};
				\draw[->] (-1.275,-0.11) -- (-1.545,-0.345) node[above]{$\mathbf{z}_P$};
				\begin{scope}[shift={(-1.18,-0.025)},rotate around z=130,canvas is xy plane at z=\h]
					\draw[fill=gray!30] (0.0,0) sin  (0.25,0.1) cos  (0.65,0) sin  (0.25,-0.1) cos (0,0)sin (-0.25,0.1) cos (-0.65,0) sin (-0.25,-0.1) cos (0,0);
					\draw[-latex] (0.1,0.075) arc (020:160:0.1) node [below] {$\tau$};
					\fill (0,0) circle (0.05);
				\end{scope}
				%%%% PROPELLER 3
				\pic[fill=gray!30, rotate=70, shift={(0,0,-\h)}] at (-0.835,1.035) {cylinder={\r}{2*\h}{45}};
				\draw[->] (-0.845,1.045) -- (-0.965,1.335) node[right]{$\mathbf{z}_P$};
				\begin{scope}[shift={(-0.75,1.20)},rotate around z=20,canvas is xy plane at z=\h]
					\draw[fill=gray!30] (0.0,0) sin  (0.25,0.1) cos  (0.65,0) sin  (0.25,-0.1) cos (0,0)sin (-0.25,0.1) cos (-0.65,0) sin (-0.25,-0.1) cos (0,0);
					\draw[latex-] (0.075,0.075) arc (00:140:0.1) node [left] {$\tau$};
					\fill (0,0) circle (0.05);
				\end{scope}
				%%%% PROPELLER 4
				\pic[fill=gray!30, rotate=240, shift={(0,0,-\h)}] at (0.85,-0.80) {cylinder={\r}{2*\h}{45}};
				\draw[->] (0.865,-0.85) -- (0.95,-1.10) node[right]{$\mathbf{z}_P$};
				\begin{scope}[shift={(0.985,-0.74)},rotate around z=15,canvas is xy plane at z=\h]
					\draw[fill=gray!30] (0.0,0) sin  (0.25,0.1) cos  (0.65,0) sin  (0.25,-0.1) cos (0,0)sin (-0.25,0.1) cos (-0.65,0) sin (-0.25,-0.1) cos (0,0);
					\draw[-latex] (0.1,-0.2) arc (00:140:0.1) node [below left] {$\tau$};
					\fill (0,0) circle (0.05);
				\end{scope}
			\end{tikzpicture}
		} 
		\vspace*{-2.4em}
		\caption{Schematic representation of a \ac{GTMR} system with its world $\mathcal{F}_W = \{O_W, \mathbf{x}_W, \mathbf{y}_W, \mathbf{z}_W\}$ and body $\mathcal{F}_B = \{O_B, \mathbf{x}_B, \mathbf{y}_B, \mathbf{z}_B\}$ reference frames.}
		\label{fig:systemModeling}
	\end{figure}
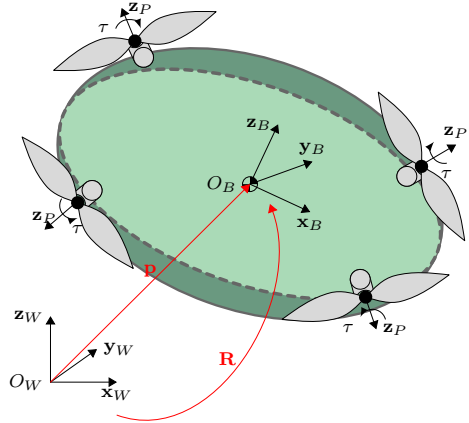

	The \ac{MRAV} is described in two reference frames: the world frame $\mathcal{F}_W$ and the body frame $\mathcal{F}_B$. The body frame is attached to the vehicle's \ac{CoM}, denoted by $O_B$, as shown in Figure \ref{fig:systemModeling}. The system's key parameters include mass $m \in \mathbb{R}_{>0}$, inertia tensor $\mathbf{J} \in \mathbb{R}^{3 \times 3}$, position $\mathbf{p} \in \mathbb{R}^3$, and linear velocity $\mathbf{v} \in \mathbb{R}^3$ expressed in $\mathcal{F}_W$. The orientation of the vehicle is described using Euler angles $\bm{\eta} = (\varphi, \vartheta, \psi)^\top$ -- roll, pitch, and yaw -- while its angular velocity is given by $\bm{\omega} \in \mathbb{R}^3$ in $\mathcal{F}_B$. The forces generated by the motors are represented as $\bm{\xi} \in \mathbb{R}^{N_p}$, acting on the vehicle's \ac{CoM}. 
	
	The force $\xi_i$ and torque $\tau_i$ exerted by the $i$-th motor, with $i=\{1, \dots, N_p\}$, on the vehicle are given by the relations $\xi_i = c_{\xi_i} \Omega_i \mathbf{z}_{P_i}$ and $\tau_i = ( c_{\xi_i} \mathbf{p}_{m_i} \times \mathbf{z}_{P_i} + c_{\tau_i} \mathbf{z}_{P_i}) \Omega_i$, where $c_{\xi_i}$ and $c_{\tau_i}$ are constants related to the propeller design, $\mathbf{p}_{m_i} \in \mathbb{R}^3$ is the motor position in $\mathcal{F}_B$, $\mathbf{z}_{P_i} \in \mathbb{S}^2$ is the motor's axis of rotation, and $\Omega_i \in \mathbb{R}_{\geq 0}$ is the squared motor speed \cite{Afifi2022ICRA, Corsini2022IROS}.
	
	To describe the time evolution of the system, a time vector $\mathbf{t} = (t_0, \dots, t_N)^\top \in \mathbb{R}^{N+1}$ is defined, where $N \in \mathbb{N}_{>0}$ represents the number of time steps, and the system evolves with a sampling period $T_s \in \mathbb{R}_{>0}$.  The state $\mathbf{x}$ and control input $\mathbf{u}$ sequences are denoted as $\mathbf{x}=(\mathbf{p}, \bm{\eta}, \mathbf{v}, \bm{\omega}, \bm{\xi})^\top \in \mathbb{R}^{(12+N_p) \times N}$ and $\mathbf{u}= \dot{\bm{\xi}} \in \mathbb{R}^{N_p \times N}$, with each $k$-th element of the sequences (i.e., $\mathbf{p}$, $\mathbf{v}$, etc.) denoted as $\bullet_k$ (i.e., $p_k$, $v_k$, etc.). Using the Newton-Euler approach, the \ac{MRAV} dynamics can be described by the set of equations:
	\begin{equation}\label{eq:multirotorDynamics}
		\left\{
		\begin{array}{l}
			\Dot{\mathbf{p}} = \mathbf{v} \\
			\Dot{\bm{\eta}} = \mathbf{T}(\bm{\eta}) \bm{\omega} \\
			m\dot{\mathbf{v}} = m\mathbf{g} + \mathbf{R}(\bm{\eta}) \mathbf{G} \bm{\xi} \\ 
			\mathbf{J}\Dot{\bm{\omega}} = -\bm{\omega} \times \mathbf{J}\bm{\omega}  + \mathbf{G} \bm{\xi}
		\end{array}
		\right.,
	\end{equation}
	where $\mathbf{T}(\bm{\eta}) \in \mathbb{R}^{3 \times 3}$ is the Jacobian matrix mapping $\bm{\omega}$ to $\dot{\bm{\eta}}$, $\mathbf{R}(\bm{\eta}) \in \mathbb{R}^{3 \times 3}$ is the rotation matrix from $\mathcal{F}_B$ to $\mathcal{F}_W$, $\mathbf{g} = (0, 0, -g)^\top$ is the gravitational acceleration vector, and $\mathbf{G} \in \mathbb{R}^{6 \times N_p}$ is the force/torque allocation matrix \cite{Afifi2022ICRA, Corsini2022IROS} mapping the forces produced by each actuator to the total force and torque acting on the vehicle's \ac{CoM}.
	
	%%% END SECTION ============================================================
	
	%%% START SECTION ==========================================================
	
	\subsection{Signal temporal logic}
	\label{sec:signalTemporalLogic}
	
	%%% SIGNAL TEMPORAL LOGIC
	\ac{STL} was first introduced in \cite{maler2004FTMATFTS} to monitor and specify the behavior of real-valued signals over time. It provides a powerful, compact, and unambiguous way to represent complex system behaviors by encoding mission specifications into a single formula $\pi$ \cite{maler2004FTMATFTS}. For example, it can capture requirements such as: ``At least two vehicles must survey regions A and B, one must visit region C within the time interval $[t_1, t_2]$, and all vehicles must adhere to safety constraints''. The formal syntax and semantics of STL are described in detail in \cite{maler2004FTMATFTS, donze2010ICFMATS}, but they are not included here for brevity.
	
	In essence, an \ac{STL} formula $\pi$ is built from a set of predicates $p_i$, where $i \in \mathbb{N}_0$. These predicates serve as atomic propositions that can represent simple system conditions, such as whether a state variable belongs to a certain region or satisfies a threshold condition. Formally, let $M = \{\mu_1, \dots, \mu_L \}$ be a set of real-valued functions of the system state, where $\mu_i \colon \mathcal{X} \rightarrow \mathbb{R}$. Each predicate $p_i$ corresponds to a subset of the state space $\mathcal{X}$, specifically defined as $p_i = \{x \in \mathcal{X} \vert \mu_i(\mathbf{x}) \geq 0 \}$.  Together, these predicates form the atomic propositions $AP \coloneqq \{p_1, \dots, p_L \}$ used to define more complex \ac{STL} specifications.
	
	These predicates can be combined using standard Boolean operators such as \textit{negation} ($\neg$), \textit{conjunction} ($\wedge$), \textit{disjunction} ($\vee$), and \textit{implication} ($\hspace{-0.45em} \implies \hspace{-0.45em}$), as well as temporal operators. Temporal operators like \textit{eventually} ($\lozenge$), \textit{always} ($\square$), \textit{until} ($\pazocal{U}$), and \textit{next} ($\bigcirc$), enable \ac{STL} to specify constraints over non-singleton intervals $I \subset \mathbb{R}$. Thus, an \ac{STL} formula $\pi$ is recursively constructed from predicates $p_i$ using the following grammar:
	\begin{equation}
		\pi \coloneqq \top \vert p \vert \neg \pi \vert \pi_1 \wedge \pi_2 \vert \pi_1 \vee \pi_2 \vert \square_I \pi \vert \lozenge_I \pi \vert \bigcirc_I \pi \vert \pi_1 \pazocal{U}_I \pi_2,
	\end{equation}
	where $\pi_1$ and $\pi_2$ are \ac{STL} formulae. In the above grammar, the symbol “$|$” denotes alternative productions (logical “or”), not an absolute value or a norm.
	
	An \ac{STL} formula $\pi$ is considered satisfied if it evaluates to true ($\top$), and violated if it evaluates to false ($\bot$). For instance, the temporal operator $\pi_1 \pazocal{U}_I \pi_2$ requires that $\pi_2$ becomes true at some time within the interval $I$, while $\pi_1$ remains continuously satisfied up to that time.
	
	%%% END SECTION ============================================================
	
	%%% START SECTION ==========================================================
	
	\subsection{Robust signal temporal logic}
	\label{sec:robustSignalTemporalLogic}
	
	%%% ROBUST SIGNAL TEMPORAL LOGIC %%%
	System uncertainties, dynamic environmental changes, and unforeseen events can all impact the satisfaction of an \ac{STL} formula $\pi$. To introduce flexibility in how well $\pi$ is satisfied and to quantify how effectively a given specification is met, the concept of \textit{robust semantics} for \ac{STL} formulae was developed \cite{donze2010ICFMATS, maler2004FTMATFTS, Fainekos2009TCS}. This \textit{robustness}, denoted as $\rho$, provides a quantitative measure to guide optimization towards the most feasible solution for satisfying the mission specifications. Robustness is defined recursively using the following formulae:
	\begin{equation}
		\begin{array}{rll}
			\rho_{p_i} (\mathbf{x}, t_k) & = & \mu_i (\mathbf{x}, t_k), \\ 
			\rho_{\neg \pi} (\mathbf{x}, t_k) & = & - \rho_\pi (\mathbf{x}, t_k), \\
			\rho_{\pi_1 \wedge \pi_2} (\mathbf{x}, t_k) & = & \min \left(\rho_{\pi_1} (\mathbf{x}, t_k), \rho_{\pi_2} (\mathbf{x}, t_k) \right), \\
			\rho_{\pi_1 \vee \pi_2} (\mathbf{x}, t_k) & = & \max \left(\rho_{\pi_1} (\mathbf{x}, t_k), \rho_{\pi_2} (\mathbf{x}, t_k) \right), \\
			\rho_{\square_I \pi} (\mathbf{x}, t_k) & = & \min\limits_{t_k^\prime \in [t_k + I]} \rho_\pi (\mathbf{x}, t_k^\prime), \\
			\rho_{\lozenge_I \pi} (\mathbf{x}, t_k) & = & \max\limits_{t_k^\prime \in [t_k + I]} \rho_\pi (\mathbf{x}, t_k^\prime), \\
			\rho_{\bigcirc_I \pi} (\mathbf{x}, t_k) & = & \rho_\pi (\mathbf{x}, t_k^\prime), \text{with} \; t_k^\prime \in [t_k+I], \\
			\rho_{\pi_1 \mathcal{U}_I \pi_2} (\mathbf{x}, t_k) & = & \max\limits_{t_k^\prime \in [t_k + I]} \Bigl( \min \left( \rho_{\pi_2} (\mathbf{x}, t_k^\prime) \right), \hfill \min\limits_{ t_k^{\prime\prime} \in [t_k, t_k^\prime] } \left( \rho_{\pi_1} (\mathbf{x}, t_k^{\prime \prime} \right)  \Bigr).
		\end{array}
	\end{equation}
	
	In this context, $t_k + I$ denotes the Minkowski sum of the scalar $t_k$ and the interval $I$. These formulae, as said, are recursively defined from \textit{predicates} $p_i$ and their corresponding real-valued functions $\mu_i(\mathbf{x}, t_k)$. Predicates are considered true if their robustness value is greater than zero ($\mu_i(\mathbf{x}, t_k) > 0$) and false otherwise ($\mu_i(\mathbf{x}, t_k) \leq 0$).
	
	The entire formula behaves as a logical expression, evaluating to false if at least one predicate is false. In simple terms, the \ac{STL} formula $\pi_1 \wedge \pi_2$ is satisfied if either $\pi_1$ or $\pi_2$ is true. Evaluation follows the application of logical and temporal operators (e.g., \textit{always}, \textit{eventually}, \textit{conjunction}) from the innermost part to the outermost part of the formula. For instance, the robustness of the formula could determine whether and ``how well'' a system remains inside a target region or avoids an obstacle at a particular interval $I$. Further details can be found in \cite{Souza2007JSTTT, maler2004FTMATFTS, donze2010ICFMATS}. In this context, we say that $\mathbf{x}$ satisfies the \ac{STL} formula $\pi$ at time $t_k$ if $\rho_\pi(\mathbf{x}, t_k) > 0$ (denoted as $\mathbf{x}(t_k)\models\pi$) and $\mathbf{x}$ violates $\pi$ if $\rho_\pi(\mathbf{x}, t_k) \leq 0$. To simplify notation, we use $\rho_\pi(\mathbf{x})$ instead of $\rho_\pi(\mathbf{x},0)$ when $t_k = 0$. Moreover, the value of $\rho_\pi(\mathbf{x}, t_k)$ represents ``how well'' the formula $\pi$ is satisfied (if $\rho_\pi(\mathbf{x}, t_k) > 0)$ or ``how much'' is violated (if $\rho_\pi(\mathbf{x}, t_k) \leq 0$), implicitly introducing a robustness criterion.
	
	With this understanding, control inputs $\mathbf{u}$ can be optimized to maximize the robustness $\rho_\pi(\mathbf{x})$ over a set of finite state and input sequences $\mathbf{x}$ and $\mathbf{u}$. An optimal  control sequence $\mathbf{u}^\star$ is considered valid if the resulting robustness $\rho_\pi (\mathbf{x}^\star)$ is positive, where $\mathbf{x}^\star$ and $\mathbf{u}^\star$ adhere to the dynamical system. A higher robustness value $\rho_\pi (\mathbf{x}^\star)$ indicates that the system can tolerate greater disturbances without violating the \ac{STL} specification.
	
	%%% END SECTION ============================================================
	
	%%% START SECTION ==========================================================
	
	\subsection{Smooth approximation}
	\label{sec:smoothApproximation}
	
	%%% SMOOTH APPROXIMATION %%%
	The computation of $\rho_\pi(\mathbf{x})$ involves non-differentiable functions like $\min$ and $\max$ (see Section \ref{sec:robustSignalTemporalLogic}), which can significantly increase the computational complexity, especially when used in optimization routines. To address this challenge, it is advantageous to utilize a \textit{smooth approximation}, denoted as $\tilde{\rho}_\pi(\mathbf{x})$, of the robustness function $\rho_\pi(\mathbf{x})$. This approach facilitates more efficient computations by replacing the non-differentiable $\min$ and $\max$ operations with smooth, differentiable alternatives. Hence, considering $\lambda \in \mathbb{R}_{>0}$ as a tunable parameter, we can express the smooth approximation of the $\min$ and $\max$ operators with $\beta$-arguments as follows:
	\begin{equation}
		\begin{split}
			&\max(\rho_{\pi_1}, \dots, \rho_{\pi_\beta}) \approx \frac{ \sum_{i=1}^\beta  \rho_{\pi_i} e^{\lambda \rho_{\pi_i}} }{ \sum_{i=1}^\beta  e^{\lambda \rho_{\pi_i}} }, 
			\\
			&\min(\rho_{\pi_1}, \dots, \rho_{\pi_\beta}) \approx -\frac{1}{\lambda} \log \left( \sum_{i=1}^\beta e^{-\lambda \rho_{\pi_i}} \right). 
		\end{split}
	\end{equation}
	
	In contrast to our earlier work \cite{SilanoRAL2021, SilanoICUAS2023}, this paper adopts a smooth robustness formulation proposed in \cite{Gilpin2021LCSS} as a numerical design choice to enable gradient-based optimization of the \ac{STL}-constrained problem. This formulation shares key properties with the widely used \ac{LSE} approximation \cite{donze2010ICFMATS}, including \textit{asymptotic completeness} and \textit{smoothness everywhere}\footnote{The property of \textit{asymptotic completeness} ensures that as the parameter $\lambda$ increases ($\lambda \rightarrow \infty$), the smooth approximation $\tilde{\rho}_\pi(\mathbf{x})$ converges to the true robustness $\rho_\pi(\mathbf{x})$. Moreover, \textit{smoothness everywhere} guarantees that the approximation is infinitely differentiable, making it compatible with gradient-based optimization methods, which are more computationally efficient for solving complex problems \cite{Gilpin2021LCSS}.}, and additionally provides a \textit{soundness} guarantee: a trajectory with strictly positive smooth robustness ($\tilde{\rho}_\pi(\mathbf{x}) > 0$) satisfies $\pi$, whereas a strictly negative value ($\tilde{\rho}_\pi(\mathbf{x}) < 0$) implies violation. As $\lambda \rightarrow \infty$, the approximation $\tilde{\rho}_\pi(\mathbf{x})$ converges to the exact robustness $\rho_\pi(\mathbf{x})$ (see Section \ref{sec:robustSignalTemporalLogic}), while remaining infinitely differentiable for finite $\lambda$, making it suitable for nonlinear, non-convex trajectory optimization. The choice of this particular smoothing technique is not claimed as a contribution of this paper; rather, it follows established practice in the literature, where different smooth robustness approximations and their trade-offs have already been analyzed \cite{Welikala2023Smooth, Dhonthi2021STLrobustness}. Systematic frameworks for computing \ac{STL} robustness through differentiable computation graphs, such as \textsc{STLCG} proposed in \cite{Leung2023IJRR}, provide 
	further evidence that gradient-based \ac{STL} optimization is a well-established numerical methodology rather than a novel methodological contribution of this work.
	
	%%% END SECTION ============================================================
	
	%%% START SECTION ==========================================================
	
	\subsection{Stochastic processes and risk measure}
	\label{sec:randomVariableStocasticProcess}
	
	%%% RANDOM VARIABLES %%%
	In addition to interpreting \ac{STL} formulae over deterministic signals, it is equally important to extend their interpretation to stochastic processes. To do so, we consider a probability space $(\Sigma, \mathcal{G}, P)$, where $\Sigma$ represents the sample space, $\mathcal{G}$ is a $\sigma$-algebra over $\Sigma$, and $P$ is a probability measure mapping from $\mathcal{G}$ to the interval $[0, 1]$. In this framework, we define a real-valued \textit{random vector} as $Z$, which is a measurable function $Z \colon \Sigma \rightarrow \mathbb{R}^n$. For $n=1$, $Z$ is referred to as a random variable. Each realization $Z(\varepsilon)$ corresponds to a specific outcome $Z$, where $\varepsilon \in \Sigma$ \cite{DurrettBook}.
	
	Since $Z$ is measurable, we can construct a probability space specifically for it, which allows us to define a \ac{CDF}, denoted as $F_Z(z)$, for the random vector $Z$. If a measurable function $g \colon \mathbb{R}^n \rightarrow \mathbb{R}$ is applied to $Z$, the result $g(Z(\varepsilon))$ becomes a \textit{derived random variable}, because function composition preserves measurability, as detailed in \cite{DurrettBook}.
	
	%%% STOCHASTIC PROCESSES %%%
	Hence, a \textit{stochastic process} is defined as a function $Y \colon T \times \Sigma \rightarrow \mathbb{R}^n$, where $T$ represents the time domain, and for each fixed time $t_k \in T$, $Y(t_k, \cdot)$ is a random vector. Essentially, a stochastic process is a collection of random vectors $\{Y(t_k, \cdot) \lvert t_k \in T \}$, with each vector defined on the probability space $(\Sigma, \mathcal{G}, P)$ and indexed by time. When $\varepsilon \in \Sigma$ is fixed, the function $Y(\cdot, \varepsilon)$ becomes a \textit{realization} of the process. Alternatively, a stochastic process can be viewed as a collection of deterministic time functions $\{ Y(\cdot, \varepsilon) \lvert \varepsilon \in \Sigma \}$, each indexed by elements from $\Sigma$ \cite{DurrettBook}. This duality is particularly useful for applying \ac{STL} formulae to systems governed by stochastic processes \cite{Lindemann2023ACM}. 
	
	%%% RISK MEASURES %%%
	\begin{sloppypar}
		A \textit{risk measure} is a function $R : \mathcal{H}(\Sigma, \mathbb{R}) \rightarrow \mathbb{R}$, which maps real-valued random variables (often referred to as \textit{cost random variables}) to real numbers. These measures are critical for assessing risks associated with uncertain outcomes, particularly in safety-critical applications. 
		
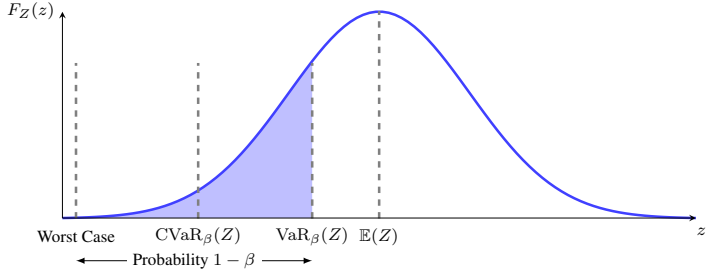
\begin{figure}[tb]
			\centering
			% GUASSIAN FUNCTION
			\pgfmathdeclarefunction{gauss}{2}{%
				\pgfmathparse{1/(#2*sqrt(2*pi))*exp(-((x-#1)^2)/(2*#2^2))}%
			}
			\scalebox{0.8}{
				\begin{tikzpicture}
					
					\begin{axis}[
						no markers, domain=0:7, samples=100,
						every axis y label/.style={at=(current axis.above origin),anchor=south},
						every axis x label/.style={at=(current axis.right of origin),anchor=west},
						axis lines=middle,
						height=5cm, width=12cm,
						xtick=\empty, ytick=\empty,
						enlargelimits=false, clip=false, axis on top,
						grid = major
						]
						% FUNCTIONS
						\addplot[fill=blue!25, draw=none, domain=0:2.76] {gauss(3.5,1)} \closedcycle;
						\addplot[very thick, blue!75] {gauss(3.5,1)};
						
						% ADDING TEXT
						\draw[yshift=-0.3cm](axis cs:3.5,0) node [fill=white] {$\mathbb{E}(Z)$};
						\draw[yshift=-0.3cm](axis cs:2.76,0) node [fill=white] {$\mathrm{VaR}_\beta(Z)$};
						\draw[yshift=-0.3cm](axis cs:1.5,0) node [fill=white] {$\mathrm{CVaR}_\beta(Z)$};
						\draw[yshift=-0.3cm](axis cs:0.15,0) node [fill=white] {Worst Case};
						
						% VERTICAL BARS
						\draw[very thick, dashed, black!50] (axis cs:3.5,0) -- (axis cs:3.5,0.40);
						\draw[very thick, dashed, black!50] (axis cs:2.76,0) -- (axis cs:2.76,0.30);
						\draw[very thick, dashed, black!50] (axis cs:1.5,0) -- (axis cs:1.5,0.30);
						\draw[very thick, dashed, black!50] (axis cs:0.15,0) -- (axis cs:0.15,0.30);
						
						% BAR UNDER TEXT
						\draw[yshift=-0.7cm, latex-latex](axis cs:0.15,0) -- node [fill=white] {Probability $1 - \beta$} (axis cs:2.76,0);
						
						% X AND Y LABELS
						\draw(axis cs:-0.35,0.40) node[text centered]{$F_Z(z)$};
						\draw(axis cs:7.06,-0.025) node[text centered]{$z$};
						
					\end{axis}
					
				\end{tikzpicture}
			}
			\caption{Illustration of the expected value $\mathbb{E}(Z)$, $\beta$-Value-at-Risk $\mathrm{VaR}_\beta(Z)$, and Conditional $\beta$-Value-at-Risk $\mathrm{CVaR}_\beta(Z)$ for a specified risk level $\beta \in (0,1)$. The axes represent the stochastic variable $z$ and its \ac{CDF} $F_Z(z)$. The shaded area corresponds to $\%\beta$ of the total area under $F_Z(z)$. $\mathrm{VaR}_\beta(Z)$ represents the value of $z$ at the $\beta$-tail of the distribution, while $\mathrm{CVaR}_\beta(Z)$ averages the worst-case values of $z$ in the $\beta$-tail. A negative $\mathrm{CVaR}_\beta(Z)$ indicates unsafe behavior.}
			\label{fig:expectedValue}
		\end{figure}

		In this paper, we focus on common risk measures, such as the expected value, $\beta$-Value-at-Risk ($\mathrm{VaR}_\beta$), and Conditional $\beta$-Value-at-Risk ($\mathrm{CVaR}_\beta$) at a specified risk level $\beta \in (0, 1)$, as shown in Figure \ref{fig:expectedValue}. Hence, the $\mathrm{VaR}_\beta$ of a random variable $Z \colon \Sigma \rightarrow \mathbb{R}$ is defined as:
		\begin{equation}
			\mathrm{VaR}_\beta(Z) \coloneqq \inf\{\alpha \in \mathbb{R} \lvert F_Z(\alpha) \geq \beta \},
		\end{equation}
		in other words, $\mathrm{VaR}_\beta(Z)$ corresponds to the $\beta$-quantile of the random variable $Z$. The $\mathrm{CVaR}_\beta$ of the random variable $Z$ is defined as:
		\begin{equation}
			\mathrm{CVaR}_\beta(Z) \coloneqq \inf_{\alpha \in \mathbb{R}} \left( \alpha + (1 - \beta)^{-1} \, \mathbb{E}([Z-\alpha]^+) \right),
		\end{equation}
		where $[Z - \alpha]^+ \coloneqq \max(Z - \alpha, 0)$ and 
		$\mathbb{E}(\cdot)$ represents the expected value. In cases where the \ac{CDF} $F_Z$ of $Z$ is continuous, $\mathrm{CVaR}_\beta(Z)$ can be expressed as $\mathbb{E}(Z | Z \geq \mathrm{VaR}_\beta(Z))$. In simpler terms, $\mathrm{CVaR}_\beta(Z)$ represents the expected value of $Z$ under the condition that $Z$ is greater than or equal to $\mathrm{VaR}_\beta(Z)$ \cite{Majumdar2020RR}. 
		
	\end{sloppypar}
	
	%%% START SECTION ==========================================================
	
	\section{Methodology}
	\label{sec:methodology}
	
	\begin{sloppypar}
		
		In this section, we present the motion planning framework designed to address the mission specifications outlined in Section \ref{sec:motivatingExample} and expressed as an \ac{STL} formula $\pi$ (see Section \ref{sec:specificationEncoding}). The motivating example (see Section \ref{sec:scenarioDescription}) focuses on the challenge of object handovers in a power line maintenance scenario, which highlights the application of human-robot collaboration with \acp{MRAV}. This task requires solving a complex nonlinear, non-smooth, and non-convex optimization problem to generate dynamically feasible trajectories that satisfy the \ac{STL} formula $\pi$, which encodes the safety constraints, temporal requirements, and human ergonomic and comfort preferences.
		
		To address the computational challenges posed by the robustness function $\rho_\pi(\mathbf{x})$, we employ smooth approximations (see Section \ref{sec:smoothApproximation}), which enable the use of gradient-based optimization techniques. These approximations make the problem more tractable by facilitating the optimization of the mission's robustness \ac{wrt} the desired system behaviors. Additionally, we enhance the planner by incorporating an energy minimization term, which implicitly extends the endurance of the \ac{MRAV} during the mission (see Section \ref{sec:motionPlanner}). This addition not only improves the vehicle's energy efficiency but also ensures that the planned trajectory remains feasible under the vehicle's physical and actuation constraints. 
		
		Additionally, we augment the framework with a risk-aware analysis (see Section \ref{sec:uncertaintyAwareRiskAnalysis}) that accounts for uncertainties in human pose. This systematic approach evaluates and quantifies the risks associated with potential deviations from \ac{STL} specifications, ensuring that the planned trajectories maintain a high probability of mission success even in uncertain environments. Lastly, a robustness-aware replanner is embedded to cope with disturbances or unforeseen events (see Section \ref{sec:replanner}).
		
	\end{sloppypar}
	
	%%% END SECTION ============================================================
	
	%%% START SECTION ==========================================================
	
	\subsection{Mission specifications encoding}
	\label{sec:specificationEncoding}
	
	\begin{sloppypar}
		
		This section outlines the mission specifications for the problem introduced in Section \ref{sec:motivatingExample} and defines the corresponding \ac{STL} formula $\pi$. The scenario involves a \ac{MRAV} equipped with a rigidly attached stick carrying a small object, tasked with performing object handovers in a power line maintenance setting, as depicted in Figure \ref{fig:handover_scenario}. 
		The mission specifications are categorized into four key areas: safety, visibility, ergonomic, and comfort requirements. 
		
		The \textit{safety requirements} are crucial for ensuring safe operation throughout the mission duration, $T_N$. These include keeping the \ac{MRAV} within the designated workspace ($\pi_\mathrm{ws}$), avoiding collisions with surrounding objects ($\pi_\mathrm{obs}$), and preventing the drone from approaching the operator from behind ($\pi_\mathrm{beh}$).
		
		The \textit{visibility requirements} ensure that during the mission, the \ac{MRAV} reaches a designated location in front of the human and remains there for a specific duration $T_\mathrm{vr}$ ($\pi_\mathrm{vr}$) before proceeding further. While approaching the operator, the drone must align its heading with the direction of movement ($\pi_\mathrm{vis}$), allowing the operator to maintain continuous visual contact with the small object on the stick. Once the \ac{MRAV} reaches the operator, a low-level onboard controller manages the handover process ($\pi_\mathrm{ho}$). 
		
		The \textit{ergonomic requirements} ensure that the \ac{MRAV} approaches the operator from their preferred direction -- whether from the front, left, right, above, or below -- based on the operator's communicated preferences before the mission ($\pi_\mathrm{pr}$), reducing strain on the operator \cite{Wojciechowska2019ACM}. 
		
		The \textit{comfort requirements} aim to enhance operator comfort during the mission by limiting the drone's approach speed ($\pi_\mathrm{vel}$), as a slower speed increases perceived safety, as discussed in \cite{Butler2001AR, Rubagotti2022RAS}. Additionally, the maximum propeller velocity is restricted ($\pi_\mathrm{pro}$) to reduce noise and wind from the propellers, mitigating any discomfort for the operator \cite{Cauchard2015ACM, Cauchard2024ACM}. This is especially important in environments where the operator may be working in close proximity to the drone for extended periods, as in power line maintenance scenarios.
		
		Importantly, the role of the proposed planner is to compute the optimal trajectory for the \ac{MRAV} from the start to the handover location, leaving the execution of the handover itself to the onboard system. Example such as those from previous work by the authors \cite{Corsini2022IROS, Afifi2022ICRA} offer viable options, but any appropriate handover algorithm can be applied based on the specific mission requirements. 
		
		The resulting \ac{STL} specification combines these requirements into a single \ac{TL} formula that enforces: (i) persistent safety over the entire mission horizon, (ii) a staged approach in which the vehicle first reaches and maintains a visibility region in front of the operator, (iii) a preference-consistent and comfort-aware approach phase, and (iv) temporal consistency during the handover operation. These behaviors are encoded through nested temporal operators that capture ordering, persistence, and timing constraints across the mission timeline. 
		These requirements are formally encoded in the following \ac{STL} formula:
		\begin{equation}\label{eq:stlFormulaProblem}
			%\resizebox{0.885\hsize}{!}{$%
				\begin{split}
					\pi =& \square_{[0,T_N]} (\pi_\mathrm{ws} \wedge \pi_\mathrm{obs} \wedge \pi_\mathrm{beh}) \wedge \\
					&\left(\lozenge_{[0, T_N - T_\mathrm{vr}]} \square_{[0,T_\mathrm{vr}]} \pi_\mathrm{vr} \right) \pazocal{U}_{[0, T_N - T_\mathrm{vr}]} \Bigl( \left( \pi_\mathrm{pr} \wedge \pi_\mathrm{vel} \wedge \pi_\mathrm{pro} \wedge \pi_\mathrm{vis} \right) \wedge \\
					&\square_{[1, T_N - T_\mathrm{vr} - 1]} \left( \pi_\mathrm{ho} \implies \bigcirc_{[0,t_{k+1}]} \pi_\mathrm{ho} \right) \Bigr)
				\end{split}\hspace{0.25em}.
				%$}%
		\end{equation}
		
		The \ac{STL} formula $\pi$ consists of nine specifications ($\pi_\mathrm{ws}$, $\pi_\mathrm{obs}$, $\pi_\mathrm{beh}$, $\pi_\mathrm{vr}$, $\pi_\mathrm{pr}$, $\pi_\mathrm{vel}$, $\pi_\mathrm{pro}$, $\pi_\mathrm{vis}$, and $\pi_\mathrm{ho}$) and two time intervals ($T_N$ and $T_\mathrm{vr}$). The following equations describe each of these specifications:
		\begin{subequations}\label{eq:STLcomponents}
			\begin{align}
				\textstyle{\pi_\mathrm{ws}} &= \textstyle{\bigwedge_{j=1}^3 \mathbf{p}^{(j)} \in (\underline{p}^{(j)}_\mathrm{ws}, \bar{p}^{(j)}_\mathrm{ws})}, \label{subeq:belongWorkspace} \\
				\textstyle{\pi_\mathrm{obs}} &= \textstyle{\bigwedge_{j=1}^3 \bigvee_{q=1}^{N_\mathrm{obs}} \mathbf{p}^{(j)} \hspace{-0.25em} \not\in (\underline{p}^{(j)}_{\mathrm{obs,q}}, \bar{p}^{(j)}_{\mathrm{obs,q}})}, \label{subeq:avoidObostacles} \\
				\textstyle{\pi_\mathrm{beh}} &=  \textstyle{\bigvee_{j=1}^3 \mathbf{p}^{(j)} \hspace{-0.25em} \not\in (\underline{p}^{(j)}_{\mathrm{beh}}, \bar{p}^{(j)}_{\mathrm{beh}})}, \label{subeq:notBehind} \\
				\textstyle{\pi_\mathrm{vr}} &= \textstyle{\bigwedge_{j=1}^3 \mathbf{p}^{(j)} \hspace{-0.25em} \in (\underline{p}^{(j)}_\mathrm{vr}, \bar{p}^{(j)}_\mathrm{vr})}, \label{subeq:safeSpot} \\
				\textstyle{\pi_\mathrm{pr}} &= \textstyle{\bigwedge_{j=1}^3 \bigvee_{q=1}^{N_\mathrm{pr}} \left( \mathbf{p}^{(j)} \hspace{-0.25em} \in (\underline{p}^{(j)}_{\mathrm{pr,q}}, \bar{p}^{(j)}_{\mathrm{pr,q}}) \right),} \label{subeq:visitTarget} \\
				%
				% \bigwedge_{k=0}^N
				\textstyle{\pi_\mathrm{vel}} &= \textstyle{ \lVert \mathbf{v}(t_k) \rVert \hspace{-0.25em} \in (\underline{\mathrm{\Gamma}}_\mathrm{vel}, \bar{\mathrm{\Gamma}}_\mathrm{vel})}, \label{subeq:velLimit} \\
				%
				% \bigwedge_{k=0}^N
				\textstyle{\pi_\mathrm{pro}} &= \textstyle{\bigwedge_{q=1}^{N_p} \Omega_q(t_k) \hspace{-0.25em} \in (\underline{\mathrm{\Gamma}}_\mathrm{pro}, \bar{\mathrm{\Gamma}}_\mathrm{pro})}, \label{subeq:propLimit}  \\
				%
				% \bigwedge_{k=0}^N
				\textstyle{\pi_{\mathrm{vis}}} &=  \textstyle{\psi(t_k) \hspace{-0.25em} \in (\psi_\mathrm{vis}(t_k) - \gamma, \psi_\mathrm{vis}(t_k) + \gamma)}, \text{with} \nonumber \\ & \textstyle{\psi_\mathrm{vis} = \atantwo(p^{(2)}_k - p^{(2)}_{k-1}, p^{(1)}_k - p^{(1)}_{k-1})}, \label{subeq:visibility} \\ % \tag{3h} \\
				\textstyle{\pi_\mathrm{ho}} &= \textstyle{\bigwedge_{j=1}^3 \mathbf{p}^{(j)} \hspace{-0.25em} \in (\underline{p}^{(j)}_\mathrm{ho}, \bar{p}^{(j)}_\mathrm{ho})}.   \label{subeq:handOver} % \tag{3i}
			\end{align}
		\end{subequations} %\setcounter{equation}{3}
		
		In \eqref{subeq:belongWorkspace}, the \ac{MRAV}'s position $\mathbf{p}^{(j)}$, where $j=\{1,2,3\}$, along the $j$-axis of the world frame $\mathcal{F}_W$, is constrained to remain within the workspace boundaries, defined by $\underline{p}^{(j)}_\mathrm{ws}$ and $\bar{p}^{(j)}_\mathrm{ws}$. Obstacle avoidance and the restriction preventing the \ac{MRAV} from approaching the operator from behind are captured by \eqref{subeq:avoidObostacles} and \eqref{subeq:notBehind}, respectively. Here, $N_\mathrm{obs}$ represents the number of obstacles in the environment, with rectangular regions having vertices $(\underline{p}^{(j)}_{\mathrm{obs,q}}$, $\bar{p}^{(j)}_{\mathrm{obs,q}})$ and $(\underline{p}^{(j)}_{\mathrm{beh}}$, $\bar{p}^{(j)}_{\mathrm{beh}})$ defining obstacle areas and the region behind the operator, respectively. 
		
\begin{figure}[tb]
			\centering
			% left - bottom - right - top
			\adjincludegraphics[width=\columnwidth, trim={{0.07\width} {0.08\height} {0.11\width} {.1\height}}, clip]{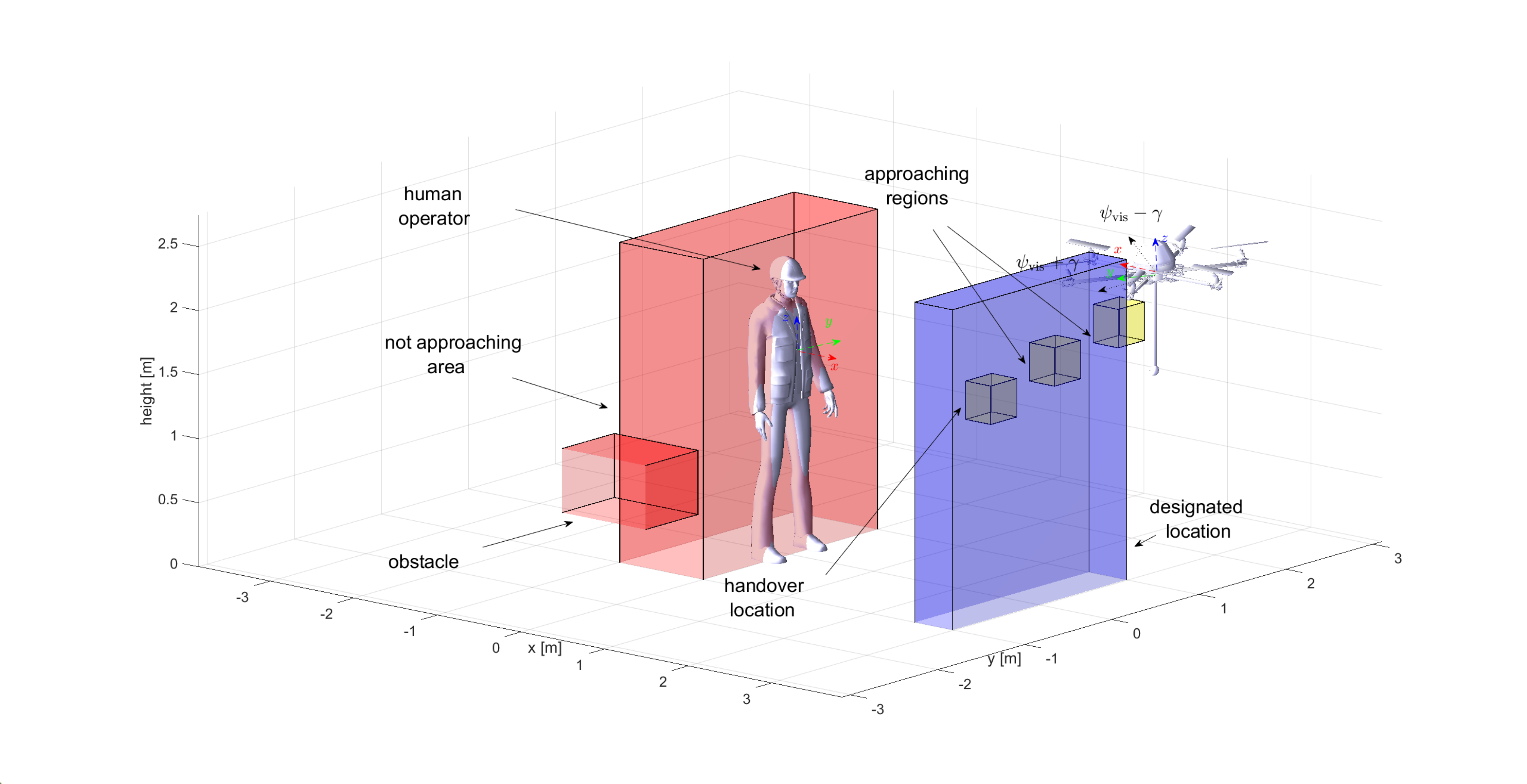}
			\vspace{-2.25em}
			\caption{Object handover scenario with highlighted approaching regions (yellow) representing the ergonomic preferences from top-to-bottom and left-to-right. Reference axes aid in visualizing the drone's maneuverability margin ($\psi_\mathrm{vis} \pm \gamma$) in the displacement direction.}
			\label{fig:schematicRepresentation}
		\end{figure}

		The visibility requirements, described in \eqref{subeq:safeSpot} and \eqref{subeq:visibility}, ensure that the \ac{MRAV} reaches a designated location in front of the operator and aligns its heading with the direction of movement. The margin for maneuverability, denoted by $\gamma \in \mathbb{R}_{>0}$, aids the optimization process by allowing the \ac{MRAV} to adjust its heading within a defined range, while still meeting visibility and approach requirements. The operator's preferences for the drone's approach direction (front, left, right, above, or below), as detailed in \cite{Wojciechowska2019ACM}, are encoded in \eqref{subeq:visitTarget}. Specifically, the vertices $(\underline{p}^{(j)}_{\mathrm{pr,q}}$,  $\bar{p}^{(j)}_{\mathrm{pr,q}})$ define in order the regions within $\mathcal{F}_W$ where the drone is allowed to approach, based on both the operator's pose and preferences. These regions are geometrically derived by extending from the handover location to the stopping area in front of the operator ($\pi_\mathrm{vr}$), based on the operator's approach preferences (left, right, above, or below). This creates well-defined approach corridors for the drone to follow. A schematic representation of this setup is shown in Figure \ref{fig:schematicRepresentation}.
		
		Comfort requirements, which include limiting the drone's approach speed and setting a maximum propeller velocity, are described in \eqref{subeq:velLimit} and \eqref{subeq:propLimit}, respectively. The parameters $(\underline{\mathrm{\Gamma}}_\mathrm{vel}, \bar{\mathrm{\Gamma}}_\mathrm{vel}) \in \mathbb{R}_{>0}$ and $(\underline{\mathrm{\Gamma}}_\mathrm{pro}, \bar{\mathrm{\Gamma}}_\mathrm{pro}) \in \mathbb{R}_{>0}$ set thresholds for the linear velocity of the drone and the rotational speed of the actuators. The relationship between the propeller speed $\Omega_q$, with $q=\{1, \dots, N_p\}$, and the system state $\mathbf{x}$ is detailed in Section \ref{sec:systemModeling}, where the dynamics governing this interaction are thoroughly explained.
		
		Lastly, \eqref{subeq:handOver} provides guidelines for completing the mission, defining the object handover position through the boundaries $\underline{p}^{(j)}_{\mathrm{ho}}$ and $\bar{p}^{(j)}_{\mathrm{ho}}$. These mission specifications \eqref{eq:STLcomponents} ensure that the \ac{MRAV} performs the task safely while considering for operator comfort and preferences.
		
		The \textit{always} operators ($\square$) in \eqref{eq:stlFormulaProblem} guarantees compliance with time requirements $T_N$ and $T_\mathrm{vr}$, corresponding to the mission duration and visibility requirement, respectively. The \textit{eventually} operator ($\lozenge$) ensures that the $\pi_\mathrm{vr}$ specification is satisfied within the time frame $T_N - T_\mathrm{vr}$. Additionally, the \textit{until} operator ($\pazocal{U}$) ensures that the \ac{MRAV} does not approach the operator for handover before reaching the designated location ($\pi_\mathrm{vr}$). 
		
		To solve the optimization problem for satisfying these mission specifications, it is necessary to compute the robustness score $\rho_\pi(\mathbf{x})$ associated with the \ac{STL} formula $\pi$. The robustness values, calculated as Euclidean distances in $\mathbb{R}^n$, indicate how well the \ac{MRAV} adheres to the mission requirements. For instance, a positive robustness score indicates that the \ac{MRAV} is within the specified region, while a negative score indicates that it has violated constraints, such as entering obstacle areas or approaching the operator from behind. Further details are provided in the authors' previous work \cite{CaballeroIEEEAccess2023, Silano2024RAS}, here omitted for brevity.
			
		\end{sloppypar}
		
		%%% END SECTION ============================================================
		
		%%% START SECTION ==========================================================
		
		\subsection{Motion planner}
		\label{sec:motionPlanner}
		
		\begin{sloppypar}
			
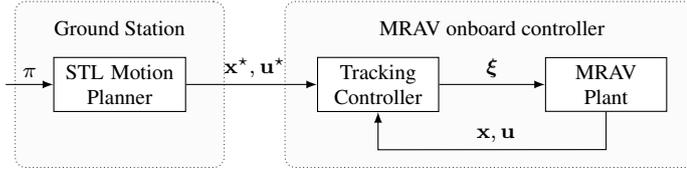
\begin{figure}[tb]
				\centering
				\begin{tikzpicture}
					
					%%%%%%%%%%%%%%%%%%%%% Motion Planner
					% Ground station block
					\node (MultiRobots-Box) at (0.25,0) [fill=gray!3,rounded       corners, draw=black!70, densely dotted, minimum height=2.2cm, minimum width=2.75cm]{}; 
					
					% Motion Planner
					\node (MotionPlanner) at (0.25,0) [text centered, fill=white, draw, rectangle, minimum	width=1.5cm, text width=5.5em]{\ac{STL} Motion\\Planner};
					
					% Links
					\draw[-latex] ($(MotionPlanner) - (1.5,0)$) -- node[above]{$\pi$} (MotionPlanner);
					
					% Multi-robots box
					\node (GroundStation) at (0.25,0.75) [text centered]{\small Ground Station};
					
					%%%%%%%%%%%%%%%%%%%
					% MRAV 
					\node (MRAV-Box) at (5.15,0) [fill=gray!3,rounded corners, draw=black!70, densely dotted, minimum height=2.2cm, minimum width=5.45cm]{};
					\node (TrackingController) at (3.65,0) [fill=white, draw, rectangle, text centered, text width=5em]{Tracking\\Controller};
					\node (MRAVPlant) at (6.65,0) [fill=white, draw, rectangle, text centered, text width=5em]{\ac{MRAV}\\Plant};
					\node (MRAV-Box-Text) at (5.15,0.75) [text centered]{\small \ac{MRAV} onboard controller};		
					
					% MRAV - Links
					\draw[-latex] (TrackingController) -- node[above]{$\bm{\xi}$} (MRAVPlant);	 
					\draw[-latex] ($ (MotionPlanner.east) $) -- node[above]{\hspace{0.3em}$\mathbf{x}^\star, \mathbf{u}^\star$} 
					(TrackingController.west);
					\draw[-latex] (MRAVPlant.south) -- ($(MRAVPlant.south) + (0,-0.5)$) -- ($(TrackingController.south) + (0,-0.5)$) node[above right]{$\hspace*{4.25em} \mathbf{x}, \mathbf{u}$} -- (TrackingController.south);
					
				\end{tikzpicture}
				\vspace*{-0.75em}
				\caption{System Architecture: The \textit{STL Motion Planner} at the ground station generates optimal trajectories ($\mathbf{x}^\star, \mathbf{u}^\star$) for the \ac{MRAV}. These trajectories are then fed into the \textit{Tracking Controller}, which operates in closed loop to compute the motor forces $\bm{\xi}$, ensuring precise flight maneuvers.}
				\label{fig:controlScheme}
			\end{figure}

			In this section, we explain how to generate the trajectory for the \ac{MRAV} based on the mission specifications $\pi$. The motion planner is designed as an optimization problem that produces a feasible trajectory while accounting for the vehicle's physical constraints. These trajectories are then used by the tracking controller to execute the handover task. Figure \ref{fig:controlScheme} illustrates the overall system architecture.
			
			Leveraging the robust semantics of \ac{STL} (see Section \ref{sec:robustSignalTemporalLogic}), the motion planner is synthesized by framing the task of determining a dynamically feasible control sequence for the \ac{MRAV} as an optimization problem. This problem seeks to satisfy a given \ac{STL} formula $\pi$ (see Section \ref{sec:specificationEncoding}) by optimizing over the control sequence $\mathbf{u} = (u_0, \dots, u_{N-1})^\top \in \mathbb{R}^N$ and the corresponding state sequence $\mathbf{x} = (x_0, \dots, x_{N-1})^\top \in \mathbb{R}^{N+1}$ (see Section \ref{sec:systemModeling}), as follows:
			\begin{subequations}\label{eq:optimizationProblem}
				\begin{align}
					&\minimize_{\mathbf{x}, \, \mathbf{u}} \;\; w \pazocal{L}(\mathbf{x}) - \rho_\pi (\mathbf{x}) \label{eq:objectiveFunctionOP} \\
					&\qquad \text{s.t.} \quad\; \mathbf{x}_0 = \mathbf{x}(t_0), \label{eq:initialStateOP} \\
					&\,\;\;\, \qquad \quad\;\;\, \mathbf{x}_{k+1} = \mathbf{f}(\mathbf{x}_k, \mathbf{u}_k), \label{eq:dynamicsOP} \\
					&\,\;\;\, \qquad \quad\;\;\, \rho_\pi(\mathbf{x}) \geq \kappa, \label{eq:robustTresholdOP}\\
					&\,\;\;\, \qquad \quad\;\;\, \underline{\bm{\xi}} \leq \bm{\xi}_k \leq \bar{\bm{\xi}}, \forall k=\{0,1, \dots, N-1\}. \label{eq:constraintOP} % \\
					%
					%&\,\;\;\, \qquad \quad\;\;\, \underline{\bm{\zeta}} \leq \bm{\zeta}_k \leq \bar{\bm{\zeta}},  
				\end{align}.
			\end{subequations}
			
			The problem is solved over a finite prediction horizon discretized into $N$ time steps, with dynamics enforced at each sampling instant. 
			
			In the above equations, \eqref{eq:objectiveFunctionOP} defines the objective function, which consists of two terms: the energy minimization term $\mathcal{L}(\mathbf{x})$ and the robustness degree $\rho_{\pi}(\mathbf{x})$. \eqref{eq:initialStateOP} specifies the system's initial state $\mathbf{x}(t_0)$, while \eqref{eq:dynamicsOP} models the system dynamics as $\mathbf{x}_{k+1} = \mathbf{f}(\mathbf{x}_k, \mathbf{u}_k)$, following the \ac{GTMR} system framework (see Section \ref{sec:systemModeling}). This ensures that the generated trajectories account for the nonlinear dynamics of the \ac{MRAV}, whether it operates in an underactuated configuration or a fully-actuated one \cite{HamandiIJRR2021}. \eqref{eq:robustTresholdOP} sets a minimum robustness threshold, acting as a safety margin to ensure satisfaction of the \ac{STL} formula $\pi$ even if the energy prioritization leads to a trade-off. Lastly, \eqref{eq:constraintOP} imposes the motor force constraints, with $\underline{\bm{\xi}}$ and $\bar{\bm{\xi}}$, representing the vehicle's minimum and maximum physical actuation limits.
			
			The energy minimization term $\mathcal{L}(\mathbf{x})$, part of the objective function \eqref{eq:objectiveFunctionOP}, is designed to extend the \ac{MRAV}'s endurance by reducing energy consumption. This is achieved through a weight coefficient $w \in \mathbb{R}_{>0}$, which allows for tuning the balance between energy efficiency and the satisfaction of mission objectives, represented by the robustness degrees $\rho_\pi(\mathbf{x})$ of the \ac{STL} formula $\pi$. A higher $w$ increases the emphasis on energy minimization, while a lower $w$ focuses more on meeting mission objectives. The energy minimization term $\mathcal{L}(\mathbf{x})$ includes the power consumption and kinetic energy, as expressed by the following equation:
			\begin{equation}\label{eq:energyTermMotionPlanner}
				\pazocal{L}(\mathbf{x}) = \sum_{i=1}^{N_p} \; c_{\xi_i} \bm{\Omega}_{i|k}^{3/2} + \frac{1}{2} \begin{bmatrix} \mathbf{v}_k^\top &  \bm{\omega}_k^\top \end{bmatrix} \begin{bmatrix} m\mathbf{I}_3 & \mathbf{0}_3 \\ \mathbf{0}_3 & \mathbf{J} \end{bmatrix} \begin{bmatrix} \mathbf{v}_k \\ \bm{\omega}_k \end{bmatrix},
			\end{equation}
			where $\mathbf{I}_3 \in \mathbb{R}^{3 \times 3}$ and $\mathbf{0}_3 \in \mathbb{R}^{3 \times 3}$ are the identity and zero matrices, respectively. 
			
			It is crucial to distinguish that, unlike the specification $\pi_\mathrm{pro}$ in \eqref{subeq:propLimit}, which restricts propeller velocity (and thus motor forces, see Section \ref{sec:systemModeling}) only during the approach phase to ensure comfort, the motor force constraints in this optimization guarantee apply throughout the entire mission. This ensures that the vehicle operates within its physical limits at all times. Typically, the allowable propeller velocity range ($\underline{\mathrm{\Gamma}}_\mathrm{pro}$ and $\bar{\mathrm{\Gamma}}_\mathrm{pro}$) is significantly narrower than the range of physically actuable propeller speeds ($\underline{\bm{\xi}}$ and $\bar{\bm{\xi}}$). This limitation minimizes noise and wind, enhancing operator comfort during close-range interactions.
			
			As described in Section \ref{sec:smoothApproximation}, the robustness function $\rho_\pi(\mathbf{x})$ involves non-differentiable functions like $\max$ and $\min$, making it non-differentiable itself \ac{wrt} the state $\mathbf{x}$ and control inputs $\mathbf{u}$. While various approaches like mixed-integer programming solvers \cite{Raman2014CDC}, non-smooth optimizers \cite{AbbasACC2013}, or stochastic heuristics \cite{Abbas2013ACM} can be employed to find a solution for this problem, it is crucial to acknowledge that the problem is inherently NP-hard, and these methods may encounter difficulties, particularly as the number of variables increases \cite{Bertsekas2012Book}. However, as demonstrated in \cite{Pant2017CCTA}, one effective strategy for managing the computational complexity is to adopt a smooth approximation $\tilde{\rho}_\pi(\mathbf{x})$ of the robust function $\rho_\pi(\mathbf{x})$. This smoothing procedure is adopted as a numerical enabler for gradient-based optimization rather than as a methodological contribution of this work. In this scenario, the resulting optimization problem remains nonlinear and non-convex, but it becomes amenable to smooth optimization techniques such as sequential quadratic programming, which can help identify a local minimum \cite{Bertsekas2012Book}. 
			
			Thus, the problem \eqref{eq:optimizationProblem} can be reformulated by replacing $\rho_\pi(\mathbf{x})$ with its smooth counterpart $\tilde{\rho}_\pi(\mathbf{x})$ as follows:
			\begin{equation}\label{eq:optimizationProblemSmooth}
				\begin{split}
					&\minimize_{\mathbf{x}, \, \mathbf{u}} \;\; w\pazocal{L}(\mathbf{x}) - \tilde{\rho}_\pi (\mathbf{x}) \\
					&\qquad \text{s.t.} \quad\; \mathbf{x}_0 = \mathbf{x}(t_0), \\
					&\,\;\;\, \qquad \quad\;\;\, \mathbf{x}_{k+1} = \mathbf{f}(\mathbf{x}_k, \mathbf{u}_k), \\
					&\,\;\;\, \qquad \quad\;\;\, \tilde{\rho}_\pi(\mathbf{x}) \geq \kappa, \\
					&\,\;\;\, \qquad \quad\;\;\, \underline{\bm{\xi}} \leq \bm{\xi}_k \leq \bar{\bm{\xi}}, \forall k=\{0,1, \dots, N-1\}. \; % \\
				\end{split}.
			\end{equation}
			
			It is important to note that the choice of the parameter $\lambda$ in Section \ref{sec:smoothApproximation} significantly impacts the problem's computational complexity. Higher values of $\lambda$  tighten the smooth approximation, which increases numerical conditioning and solver difficulty. Conversely, smaller values of $\lambda$ impose constraints on the motion planner, potentially resulting in trajectories that do not satisfy the \ac{STL} specification. Furthermore, as demonstrated in \cite{Pant2017CCTA}, an appropriate value of $\kappa$ can be chosen to ensure that the approximation error remains within a desired bound, such that $\lvert \rho_\pi(\mathbf{x}) - \tilde{\rho}_\pi(\mathbf{x}) \rvert \leq \kappa$.
			
			After the successful completion of the optimization process \eqref{eq:optimizationProblemSmooth}, the obtained solution can be transmitted to the onboard tracking controller (such as those described in \cite{Corsini2022IROS, Afifi2022ICRA}) of the \ac{MRAV} for precise tracking and execution (see Figure \ref{fig:controlScheme}). 
			
		\end{sloppypar}
		
		%%% END SECTION ============================================================
		
		%%% START SECTION ==========================================================
		
		\subsection{Uncertainty-aware risk analysis}
		\label{sec:uncertaintyAwareRiskAnalysis}
		
		The motion planner introduced in this paper generates feasible trajectories for an \ac{MRAV}, enhancing ergonomic, safety, and comfort considerations in \ac{HRI}, especially for tasks performed at heights, such as power line maintenance. Using an object handover task as a motivating example, we address the challenge of uncertainties in human pose, which can affect the satisfaction of the \ac{STL} formula $\pi$ and potentially lead to mission failure. To mitigate this, we propose a risk-aware analysis framework that assesses and quantifies the risks associated with deviations from \ac{STL} specifications \eqref{eq:stlFormulaProblem} in the trajectory derived from the optimization problem \eqref{eq:optimizationProblemSmooth}. This framework evaluates whether a given success rate (e.g., 80\%) for satisfying \ac{STL} specifications can still be achieved under uncertainty, drawing on methodologies presented in \cite{Lindemann2021CDC, Lindemann2023ACM, Lindemann2022ACM}.
		
		While the \ac{STL} formula $\pi$, as defined in Section \ref{sec:specificationEncoding}, traditionally applies to deterministic signals represented by the state sequence $\mathbf{x}$, we extend its interpretation to handle uncertainties using a stochastic process, denoted as $Y$ (see Section \ref{sec:randomVariableStocasticProcess}). This extension accounts for variability in human pose, which directly influences critical mission elements such as the object handover ($\pi_\mathrm{ho}$), the \ac{MRAV}'s approach direction ($\pi_\mathrm{pr}$), heading alignment  ($\pi_\mathrm{vis}$), and safety requirements like not approaching the operator from behind ($\pi_\mathrm{beh}$).  Human pose is modeled as uncertain, with a random distribution characterized by specific mean ($\mu_z$) and covariance ($\sigma_z$) parameters -- for example, using a normal distribution $\mathcal{N}(\mu_{z}, \sigma_{z})$.
		
		For any given realization $Y(\cdot, \varepsilon)$ of the stochastic process, we can evaluate whether the realization satisfies the \ac{STL} formula $\pi$. However, in this stochastic context, satisfaction of $\pi$ becomes probabilistic -- some realizations of $Y$ may satisfy $\pi$, while others may not. To manage this uncertainty, we apply risk measures, as described in Section \ref{sec:randomVariableStocasticProcess}, to quantify the likelihood that the stochastic process $Y$ fails to satisfy the \ac{STL} specification $\pi$. This approach provides a systematic way to handle uncertainty in \ac{HRI} and ensures that mission objectives are met with a quantifiable level of confidence. 
		
		To compute the level of risk of not satisfying the \ac{STL} formula $\pi$, the following steps are taken:
		\begin{itemize}
			\item[(i)] \textit{Identify the stochastic elements} $Y$ in the mission specifications (see Section \ref{sec:specificationEncoding}) that introduce uncertainty, such as human pose or environmental conditions.
			\item[(ii)] \textit{Compute the robustness score} $\rho_\pi(\mathbf{x})$, which evaluates the degree to which the \ac{MRAV} satisfies the \ac{STL} formula $\pi$ under a specific realization $Y(\cdot, \varepsilon)$ of the stochastic process. The robustness score indicates how far the system is from violating or satisfying the mission objectives.
			\item[(iii)] \textit{Apply an appropriate risk measure}, such as $\mathrm{VaR}_\beta$ (see Section \ref{sec:randomVariableStocasticProcess}), to quantify the risk of not meeting the \ac{STL} specification $\pi$. This provides a statistical threshold for the level of risk posed by uncertainties in the system.
		\end{itemize}
		
		Once the stochastic elements -- such as human pose -- are identified, we proceed to compute the robustness score $\rho_\pi(\mathbf{x})$ for each realization of the stochastic process $Y$. The \ac{STL} formula $\pi$ in \eqref{eq:stlFormulaProblem} consists of two types of predicates: those evaluating whether the \ac{MRAV}'s position belongs to a specific region ($\pi_\mathrm{ws}$, $\pi_\mathrm{vr}$, $\pi_\mathrm{pr}$, $\pi_\mathrm{vel}$, $\pi_\mathrm{pro}$, $\pi_\mathrm{vis}$, and $\pi_\mathrm{ho}$) and those ensuring it avoids restricted regions ($\pi_\mathrm{obs}$ and $\pi_\mathrm{beh}$). Given the structural similarities among these predicates, the analysis can focus on two key ones: $\pi_\mathrm{ho}$, which defines the handover location (in \eqref{subeq:handOver}), and $\pi_\mathrm{beh}$, which ensures the \ac{MRAV} does not approach the human from behind (in \eqref{subeq:notBehind}). These predicates are critical because they establish the spatial boundaries that the \ac{MRAV} must either remain within or avoid for safe and efficient collaboration.
		
		For the handover location, the robustness score $\rho_{\pi_\mathrm{ho}}$ is calculated by measuring how close the \ac{MRAV}'s current position is to the defined boundaries of the handover region. We compute the robustness score as follows:
		\begin{equation}\label{eq:positiveRobustnessStochastic}
			\resizebox{0.915\hsize}{!}{$%
				\begin{split}
					\rho_{\pi_\mathrm{ho}}(\mathbf{x}, Y(\cdot, \varepsilon)) &= \min \Bigl( \rho_{\underline{\pi}^{(1)}}(\mathbf{x}, Y(\cdot, \varepsilon)), \rho_{\underline{\pi}^{(2)}}(\mathbf{x}, Y(\cdot, \varepsilon)), \rho_{\underline{\pi}^{(3)}}(\mathbf{x}, Y(\cdot, \varepsilon)), \\ 
					&  \rho_{\bar{\pi}^{(1)}}(\mathbf{x}, Y(\cdot, \varepsilon)), \rho_{\bar{\pi}^{(2)}}(\mathbf{x}, Y(\cdot, \varepsilon)), \rho_{\bar{\pi}^{(3)}}(\mathbf{x}, Y(\cdot, \varepsilon)) \Bigr), 
				\end{split}
				$}%
		\end{equation}
		with
		\begin{equation*}
			\resizebox{0.925\hsize}{!}{$%
				\rho_{\underline{\pi}^{(j)}}(\mathbf{x}, Y(\cdot, \varepsilon)) = \mathrm{p}_k^{(j)} - \underline{p}^{(j)}_\mathrm{ho}(Y(\cdot, \varepsilon)), \quad
				\rho_{\bar{\pi}^{(j)}}(\mathbf{x}, Y(\cdot, \varepsilon)) = \bar{p}^{(j)}_\mathrm{ho}(Y(\cdot, \varepsilon)) - \mathrm{p}_k^{(j)}. 
				$}%
		\end{equation*}
		
		Here, the terms $\underline{p}_\mathrm{ho}^{(j)} (Y(\cdot, \varepsilon))$ and $\bar{p}_\mathrm{ho}^{(j)} (Y(\cdot, \varepsilon))$ represent the lower and upper bounds of the handover location, modeled as stochastic variables. The robustness score measures the Euclidean distance between the \ac{MRAV}'s current position $\mathrm{p}_k^{(j)}$ and these bounds, reflecting how uncertainty in the handover location impacts mission success.
		
		Similarly, for the no-approach-from-behind predicate $\pi_\mathrm{beh}$, the robustness score is calculated by taking the inverse of the minimum distance to the boundary, as shown below:
		\begin{equation}\label{eq:negativeRobustnessStochastic}
			\resizebox{0.915\hsize}{!}{$%
				\begin{split}
					\rho_{\pi_\mathrm{beh}}(\mathbf{x}, Y(\cdot, \varepsilon)) &= -\min \Bigl(\rho_{\underline{\pi}^{(1)}}(\mathbf{x}, Y(\cdot, \varepsilon)), \rho_{\underline{\pi}^{(2)}}(\mathbf{x}, Y(\cdot, \varepsilon)), \rho_{\underline{\pi}^{(3)}}(\mathbf{x}, Y(\cdot, \varepsilon)), \\ 
					&  \rho_{\bar{\pi}^{(1)}}(\mathbf{x}, Y(\cdot, \varepsilon)), \rho_{\bar{\pi}^{(2)}}(\mathbf{x}, Y(\cdot, \varepsilon)), \rho_{\bar{\pi}^{(3)}}(\mathbf{x}, Y(\cdot, \varepsilon)) \Bigr).
				\end{split}
				$}%
		\end{equation}
		
		The terms $\rho_{\underline{\pi}^{(j)}}(\mathbf{x}, Y(\cdot, \varepsilon))$ and $\rho_{\bar{\pi}^{(j)}}(\mathbf{x}, Y(\cdot, \varepsilon))$ in \eqref{eq:negativeRobustnessStochastic} are computed by replacing the handover location bounds in \eqref{eq:positiveRobustnessStochastic} with the corresponding bounds for the region behind the operator $\underline{p}^{(j)}_\mathrm{beh}(\mathbf{x}$, $Y(\cdot, \varepsilon))$ and $\bar{p}^{(j)}_\mathrm{beh}(\mathbf{x}, Y(\cdot, \varepsilon))$, respectively.
		
		The robustness scores \eqref{eq:positiveRobustnessStochastic} and \eqref{eq:negativeRobustnessStochastic} reflect how uncertainty in human pose affects satisfaction of the \ac{STL} specifications. If the robustness score is positive for a given realization $Y(\cdot, \varepsilon)$, the \ac{MRAV} satisfies the \ac{STL} formula $\pi$. A zero or negative score indicates a failure to meet the specification. For simplifying the analysis, we assume that the realization $Y(\cdot, \varepsilon)$ of the stochastic process $Y$ is time-independent, meaning the human pose remains constant throughout the mission. In cases where the position changes, a new trajectory plan is computed for the \ac{MRAV}, as outlined in Section \ref{sec:replanner}. 
		
		To evaluate the risk of mission failure under uncertainty, we apply a risk measure, such as $\mathrm{VaR}_\beta$ or its counterpart, $\mathrm{CVaR}_\beta$, as outlined in Section \ref{sec:randomVariableStocasticProcess}. The $\mathrm{VaR}_\beta$ provides a threshold value, below which the robustness score $\rho_\pi(\mathbf{x})$ is unlikely to fall with a probability of at least $1-\beta$. This measure quantifies the system's tolerance to uncertainty while still ensuring mission success. As described in Section \ref{sec:randomVariableStocasticProcess}, the formal definition of $\mathrm{VaR}_\beta$ is given by:
		\begin{equation}\label{eq:varIdeal}
			\mathrm{VaR}_\beta(\rho_\pi(\mathbf{x})) \coloneqq \inf\{\alpha \in \mathbb{R} \lvert F_{\rho_\pi} (\alpha) \geq \beta \},
		\end{equation}
		where $F_{\rho_\pi} (\alpha)$ represents the \ac{CDF} of the robustness scores $\rho_\pi$. However, since the exact \ac{CDF} is often unknown and may require complex computations, such as high-dimensional integrals \cite{Lindemann2021CDC, Lindemann2023ACM, Lindemann2022ACM}, we adopt a data-driven approach by estimating it using empirical data from $K$ observed realizations of the stochastic process $Y$ (e.g., human pose). The empirical \ac{CDF} is computed as:
		\begin{equation} \label{eq:empiricalCDF}
			\hat{F}_{\rho_\pi} (\alpha, \rho_\pi(\mathbf{x})) \coloneqq \frac{1}{K} \sum^K_{i=1} \mathbb{I}(^i\rho_\pi(\mathbf{x}) \leq \alpha), 
		\end{equation}
		where $\mathbb{I}$ is the indicator function, which returns 1 if the robustness score  for the $i$-th realization ($^i\rho_\pi(\mathbf{x})$) of $Y$ is less than or equal to $\alpha$, and 0 otherwise. Based on this empirical data \eqref{eq:empiricalCDF}, we can calculate upper and lower bounds for $\mathrm{VaR}_\beta$ at a confidence level $\delta \in (0,1)$, as detailed in \cite{Lindemann2021CDC, Lindemann2023ACM, Lindemann2022ACM}, as follows:
		\begin{subequations} \label{eq:confidenceLevelBounds} 
			\begin{align}
				\overline{\mathrm{VaR}}_\beta (\rho_\pi(\mathbf{x}), \delta) &\coloneqq \inf \left\{\alpha \in \mathbb{R} \lvert \hat{F}_{\rho_\pi}(\alpha, \rho_\pi(\mathbf{x})) - \sqrt{\frac{\ln(2/\delta)}{2K}} \geq \beta \right\}, \\
				\underline{\mathrm{VaR}}_\beta (\rho_\pi(\mathbf{x}), \delta) &\coloneqq \inf \left\{\alpha \in \mathbb{R} \lvert \hat{F}_{\rho_\pi}(\alpha, \rho_\pi(\mathbf{x})) + \sqrt{\frac{\ln(2/\delta)}{2K}} \geq \beta \right\}.
			\end{align}
		\end{subequations} 
		
		Thus, by observing $K$ realizations of the stochastic process $Y$, and selecting an appropriate values for $\delta \in (0,1)$, we can calculate upper and lower bounds for $\mathrm{VaR}_\beta$. This method provides an estimate of the $\mathrm{VaR}_\beta$, and as the number of realizations $K$ increases, the estimate becomes more accurate, converging to the true $\mathrm{VaR}_\beta$ value. This approach enables us to quantify the risk of the \ac{MRAV} failing to meet mission objectives under the inherent uncertainty in human pose.
		
		%%% END SECTION ============================================================
		
		%%% START SECTION ==========================================================
		
		\subsection{Event-driven replanner}
		\label{sec:replanner}
		
		\begin{sloppypar}
			
			As described earlier, the offline motion planner \eqref{eq:optimizationProblemSmooth} generates trajectories that satisfy safety, ergonomic, and comfort considerations encoded in the \ac{STL} specification $\pi$. During execution, however, disturbances such as wind gusts, hardware faults, or battery degradation may cause deviations between the nominal plan and the actual \ac{MRAV} state. To ensure mission continuity under such conditions, we introduce a robustness-aware event-driven replanner that recomputes corrective trajectory segments online.
			
			Discrete monitoring instants are defined by the vector $\bar{\mathbf{t}} = (\bar{t}_0, \dots, \bar{t}_L)^\top \in \mathbb{R}^{L+1}$, with sampling period $T_s$ and event-checking period $T_e \in \mathbb{R}_{\geq 0}$. Low-rate ``topic'' waypoint are defined at times $\hat{\mathbf{t}} = (\hat{t}_0, \dots, \hat{t}_G)^\top \in \mathbb{R}^{G+1}$, with update period $T_g \in \mathbb{R}_{\geq 0}$ satisfying $T_g \gg T_s$.
			
			Let $\tilde{\mathbf{p}}_l$ denote the measured \ac{MRAV} position at $\bar{t}_l \in \bar{\mathbf{t}}$, and $\mathbf{p}^\star_l$ the nominal planned position. Replanning is triggered when either a geometric deviation exceeds a threshold $\zeta \in \mathbb{R}_{\geq 0}$ or the predicted robustness of the safety-critical \ac{STL} subset falls below a prescribed margin $\kappa_\mathrm{crit} \in \mathbb{R}_{> 0}$, namely:
			\begin{equation}
				||\tilde{\mathbf{p}}_l - \mathbf{p}^\star_l|| > \zeta\ \vee\ \hat{\rho}_{\pi_\mathrm{crit}}(\tilde{\mathbf{x}}_l) < \kappa_\mathrm{crit}.
			\end{equation}
			
			Here, $\pi_\mathrm{crit}$ collects only safety-critical predicates, such as workspace containment $\pi_\mathrm{ws}$, obstacle avoidance $\pi_\mathrm{obs}$, and no-approach-from-behind $\pi_\mathrm{beh}$. These predicates are encoded geometrically as polyhedral regions in the workspace, as described in Section~\ref{sec:specificationEncoding}. 
			
			During execution, instead of recomputing the full \ac{STL} robustness, a fast surrogate $\hat{\rho}_{\pi_{\mathrm{crit}}}(\tilde{\mathbf{x}}_l)$ is evaluated from signed distances to the faces of these polyhedra. Let $\pazocal{I}_\mathrm{crit}$ denote the index set of safety-critical predicates included in $\pi_\mathrm{crit}$. Specifically, we write:
			\begin{equation}
				\hat{\rho}_{\pi_\mathrm{crit}}(\tilde{\mathbf{x}}_l) = \min_{p_i \in \pazocal{I}_\mathrm{crit}} \min_j\ (b_{ij}-\mathbf{a}_{ij}^\top \mathbf{p}),
			\end{equation}
			where $\mathbf{p}$ denotes the \ac{MRAV} position extracted from the state $\mathbf{x}$. Each safety region associated with predicate $p_i$ (see Section \ref{sec:signalTemporalLogic}) is represented as the intersection of linear half-spaces $\mathbf{a}_{ij}^\top \mathbf{p} \le b_{ij}$, with known constant vectors $\mathbf{a}_{ij}$ and scalars $b_{ij}$ defining the outward normal directions and offsets of the polyhedral faces induced by workspace boundaries, obstacles, and exclusion zones. This geometric surrogate enables efficient online evaluation and avoids recomputation of the full \ac{STL} robustness during recovery.
			
			Upon triggering, the ground station (see Figure \ref{fig:controlScheme}) solves a short-horizon recovery problem over $[\bar{t}_l + t_c, \hat{t}_{g+1}]$, where $t_c$ represents the maximum expected computation time for replanning\footnote{The computation time $t_c$ is estimated by running multiple instances of the \ac{STL} optimization problem under varying conditions, such as different \ac{MRAV} initial positions, obstacle placements ($\pi_\mathrm{obs}$), visibility requirements ($\pi_\mathrm{vis}$), and operator preferences ($\pi_\mathrm{pr}$). This ensures that the \ac{MRAV} computes a feasible trajectory from the triggered position $\mathbf{p}_l$ to the next topic waypoint $\mathbf{p}_{g+1}$, mitigating discrepancies between the planned and actual trajectories.}, and $\hat{t}_{g+1}$ refers to the next ``topic'' waypoint $\mathbf{p}_{g+1}$. The replanning problem minimizes deviation from the nominal trajectory while explicitly preserving a safety margin, e.g.,
			\begin{equation}\label{eq:eventDrivenProblem}
				\begin{split}
					&\minimize_{\mathbf{x}, \, \mathbf{u}} \;\;  w_r ||\mathbf{x} - \mathbf{x}^\star|| - \hat{\rho}_{\pi_\mathrm{crit}}(\mathbf{x}) + w_u ||\mathbf{u}||^2 \\
					&\qquad \text{s.t.} \quad\; \mathbf{x}_0 = \mathbf{x}(t_0), \\
					&\,\;\;\, \qquad \quad\;\;\, \mathbf{x}_{k+1} = \mathbf{f}(\mathbf{x}_k, \mathbf{u}_k), \\
					&\,\;\;\, \qquad \quad\;\;\, \hat{\rho}_{\pi_\mathrm{crit}}(\mathbf{x}) \geq \kappa_\mathrm{crit}, \\
					&\,\;\;\, \qquad \quad\;\;\, \underline{\bm{\xi}} \leq \bm{\xi}_k \leq \bar{\bm{\xi}}, \forall k=\{0,1, \dots, N-1\}. \; % \\
				\end{split}.
			\end{equation}
			
			Here, $w_r$ penalizes deviation from the nominal offline plan $\mathbf{x}^\star$, encouraging rapid reconnection to the reference trajectory, while $w_u$ regularizes control effort to avoid aggressive actuation during recovery. The negative robustness term explicitly drives the optimizer toward states that increase the safety margin of the critical specification subset.
			
			This formulation explicitly optimizes safety robustness during recovery rather than minimizing tracking error alone. Specifically, the \ac{STL} specification is decomposed as $\pi = \pi_\mathrm{crit} \wedge \pi_\mathrm{soft}$, where $\pi_\mathrm{soft}$ includes comfort- and ergonomics-related predicates (visibility $\pi_\mathrm{vis}$, approach preference $\pi_\mathrm{pr}$, velocity $\pi_\mathrm{vel}$ and propeller limits $\pi_\mathrm{pro}$, and handover timing $T_\mathrm{vr}$).
			
			The offline planner \eqref{eq:optimizationProblemSmooth} enforces the full specification $\pi$, whereas the online replanner enforces only $\pi_\mathrm{crit}$ while minimizing deviation from the nominal plan. During recovery, the constraint $\hat{\rho}_{\pi_\mathrm{crit}}(\mathbf{x})\geq \kappa_\mathrm{crit}$ is maintained to guarantee a minimum safety margin. This separation reflects a hierarchical planning strategy: the offline optimization enforces all comfort, ergonomic, and timing objectives under nominal conditions, whereas the online replanner acts as a safety-oriented recovery layer whose primary role is to rapidly restore feasibility after disturbances. Enforcing the full \ac{STL} specification online would substantially increase computational complexity and could hinder timely recovery, while temporarily relaxing non-critical objectives enables fast reconnection to the nominal plan, after which the complete mission specification is again satisfied.
			
			Because replanning is performed over a reduced horizon and enforces only a restricted subset of safety-critical predicates, its computational burden is substantially lower than that of the initial offline planning stage, enabling execution within seconds and supporting online deployment during close-proximity human–robot collaboration. Moreover, the online recovery problem is formulated with a fixed set of safety-critical predicates and a fixed prediction horizon, independent of the number or complexity of the \ac{STL} specifications used in the offline phase; as a result, the size and structure of the replanning optimization remain unchanged across missions, preventing growth of the online computational cost with full specification complexity.
			
			Specifically, recovery is carried out over the shortened interval $[\bar{t}_l + t_c, \hat{t}_{g+1}] \ll [0, T_N]$ and requires satisfaction only of workspace containment ($\pi_\mathrm{ws}$), obstacle avoidance ($\pi_\mathrm{obs}$), and no-approach-from-behind ($\pi_\mathrm{beh}$). Other objectives, such as energy efficiency or trajectory smoothness, are temporarily relaxed to prioritize rapid feasibility restoration and collision avoidance, allowing the \ac{MRAV} to quickly return to a safe operational state.
			
		\end{sloppypar}
		
		%%% END SECTION ============================================================
		
		%%% START SECTION ==========================================================
		
		\section{Simulation Results}
		\label{sec:simulationResults}
		
		\begin{sloppypar}
			
			All parameters required to initialize the proposed planning, risk analysis, and replanning framework are summarized in Table~\ref{tab:tableParamters}. This table collects the \ac{MRAV} physical and actuation limits, workspace geometry, \ac{STL} predicate bounds and timing intervals, robustness and smoothing parameters, replanning thresholds, and the statistical descriptors used to model human pose uncertainty. Together, these quantities fully specify the optimization problem in Section \ref{sec:motionPlanner} and allow reproduction of the reported results.
			
			\begin{table}[tb]
				\centering
				\caption{Parameter values for the optimization problem.}
				\label{tab:tableParamters}
				\vspace{-1.2em}
				\begin{adjustbox}{max width=0.98\columnwidth}
					\begin{tabular}{c|c|c|c|c|c}
						\hline 
						\textbf{Description} & \textbf{Symbol} & \textbf{Value} & \textbf{Description} & \textbf{Symbol} & \textbf{Value}\\
						\hline \hline
						Min motor forces & $\underline{\bm{\xi}}$ & $\SI{0.29}{\newton}$ & Max motor forces & $\bar{\bm{\xi}}$ & $\SI{11.5}{\newton}$ \\
						Weight coefficient & $w$ & $\SI{0.50}{[-]}$ & Number of motors & $N_p$ & $\SI{6}{[-]}$ \\
						\ac{MRAV} mass & $m$ & $\SI{2.25}{\kilogram}$ & Inertia & $\mathbf{J}$ & $\text{diag}\{2.07, 2.10, 3.10\} \times 10^{-2}$ \\
						Force cst. parameter & $c_{\xi_i}$ & $\num{11.5e-4}{[-]}$ & Torque cst. parameter & $c_{\tau_i}$ & $\num{2.38e-05}{[-]}$ \\
						Operator init. position & $\mathbf{p}_\mathrm{hum}$ & $\{0.0,0.0,0.0\}\si{\meter}$ & Operator init. orientation & $\bm{\eta}_\mathrm{hum}$ & $\{0.0,0.0,0.0\}\si{\degree}$ \\
						\ac{MRAV} init. position & $\mathbf{p}_0$ & $\{-1.80,0.0,1.0\}\si{\meter}$ & \ac{MRAV} init. orientation & $\bm{\eta}_0$ & $\{0.0,0.0,0.0\}\si{\degree}$ \\
						Number of samples & $N$ & $\SI{170}{[-]}$ & Sampling period & $T_s$ & $\SI{0.1}{\second}$ \\
						Smooth apprx. parameter & $\lambda$ & $\SI{10}{[-]}$ & Mission duration & $T_N$ & $\SI{17}{\second}$ \\
						Max linear velocity & $\bar{\mathrm{\Gamma}}_\mathrm{vel}$ & $\SI{0.4}{\meter\per\second}$ & Min linear velocity & $\underline{\mathrm{\Gamma}}_\mathrm{vel}$ & $\SI{0}{\meter\per\second}$ \\
						Max prop. speed & $\sqrt{\bar{\mathrm{\Gamma}}_\mathrm{pro}}$ & $\SI{80}{\hertz}$ & Min prop. speed & $\sqrt{\underline{\mathrm{\Gamma}}_\mathrm{pro}}$ & $\SI{40}{\hertz}$\\
						Level of confidence & $\delta$ & $\num{0.01}{[-]}$ & Heading manu. margin & $\gamma$ & $\SI{30}{\degree}$ \\
						Number obstacles & $N_\mathrm{obs}$ & $\SI{1}{[-]}$ & Handover location & $\mathbf{p}_\mathrm{pr,3}$ & $\{1.0,1.0,1.0\}\si{\meter}$ \\
						Number pref. regions & $N_\mathrm{pr}$ & $\SI{3}{[-]}$ & Number of realizations & $K$ & $\num{15000}{[-]}$ \\
						Replanning threshold & $\zeta$ & \SI{1}{\meter} & Event-driven period & $T_e$ & $\SI{0.5}{\second}$ \\
						Waypoint period & $T_g$ & $\SI{1}{\second}$ & Computation time & $t_c$ & $\SI{0.4}{\second}$ \\
						Visibility time & $T_\mathrm{vr}$ & $\SI{5}{\second}$ & Robustness threshold & $\kappa$ & $\num{0.2}{[-]}$ \\
						Critical robustness margin & $\kappa_\mathrm{crit}$ & $\num{0.2}{[-]}$ & Reconnection weight & $w_r$ & $\num{10}{[-]}$ \\
						Control regularization & $w_u$ & $\num{0.1}{[-]}$ & Replanning horizon steps & $N_\mathrm{r}$ & $\num{50}{[-]}$ \\
						\hline
					\end{tabular}
				\end{adjustbox}
			\end{table}
			
			To validate the proposed framework, simulations were conducted at two levels of fidelity. First, numerical experiments in MATLAB were used to analyze the optimization-based planner in isolation, without explicitly modeling the \ac{MRAV} dynamics or the low-level tracking controller. These tests provided insight into the structure of the planner trajectories and into the role of the \ac{STL} specifications and cost terms. The analysis was then extended to the Gazebo robotics simulator in a \ac{SITL} configuration \cite{SilanoSMC19, Cora2024RAM}, where full vehicle dynamics and onboard controllers were included, allowing assessment of closed-loop feasibility and implementation realism. This progression from MATLAB to Gazebo therefore supports a comprehensive evaluation of both high-level planning and execution-level behavior. 
			
			The proposed architecture deliberately separates offline mission planning from online recovery: the full \ac{STL}-constrained nonlinear program is solved prior to execution, whereas only short-horizon, safety-critical replanning problems are solved online during flight.
			
			The evaluation metrics employed in this section are selected to directly reflect the objectives encoded in the \ac{STL} formulation rather than to provide generic benchmarking scores. In particular, robustness margins, energy-related quantities, replanning time, and uncertainty-driven risk indicators are used to assess safety, comfort, feasibility, and operational viability in the considered collaborative task. These indicators are also used to provide a quantitative basis for comparison with conventional planning approaches that do not explicitly encode such specifications, without forcing misleading equivalences between planners addressing fundamentally different problem formulations.
			
			It is worth emphasizing that the collaboration-oriented specifications and risk-based indicators introduced in this paper are not strictly required to obtain dynamically feasible trajectories. Even in their absence, the planner can satisfy vehicle dynamics, actuator limits, workspace constraints, and basic mission timing. Their role is instead to shape the resulting behavior toward interaction-aware solutions by enforcing ergonomic approach directions, limiting approach speed and propeller activity near the operator, preserving visibility, and quantifying sensitivity to human-pose uncertainty. The analyses in Sections~\ref{sec:comparativeAnalysis} and~\ref{sec:riskAssessment} illustrate that removing or relaxing these terms does not prevent task execution, but leads to trajectories with less desirable interaction characteristics and eliminates quantitative insight into robustness margins in collaborative settings.
			
			The simulations were designed to validate four key aspects: (i) satisfaction margins \ac{wrt} to mission requirements (see Section \ref{sec:objectHandover}); (ii) the influence of the energy-aware term on trajectory structure and endurance (see Section \ref{sec:comparativeAnalysis}); (iii) the ability of the system to recover from disturbances through robustness-aware replanning (see Section \ref{sec:comparativeAnalysis}); and (iv) the effectiveness of the uncertainty-aware risk analysis framework in quantifying degradation of \ac{STL} satisfaction under human pose variability (see Section \ref{sec:riskAssessment}).
			
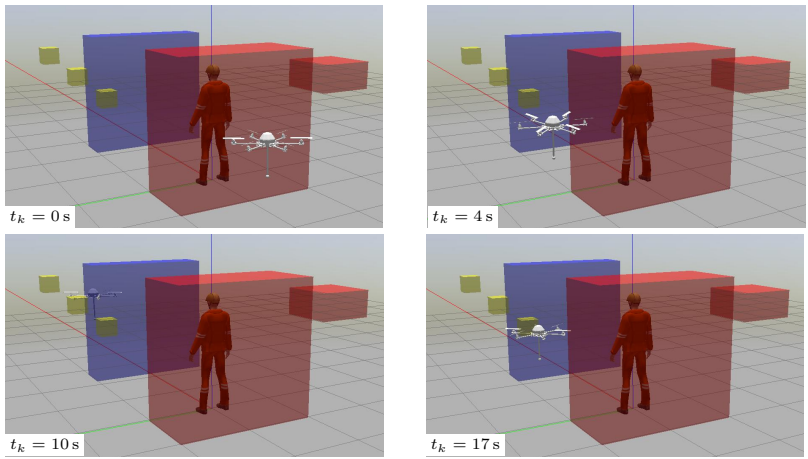
\begin{figure*}
				\centering%\hspace*{-0.20cm}
\begin{subfigure}{0.45\columnwidth}
  \centering
  % left - bottom - right - top
  \scalebox{0.85}{
  \begin{tikzpicture}
    \node[anchor=south west,inner sep=0] (img) at (0,0) { 
    % left - bottom - right - top
    \adjincludegraphics[trim={{.175\width} {.20\height} {.25\width} 
    {.20\height}},clip,scale=0.185]{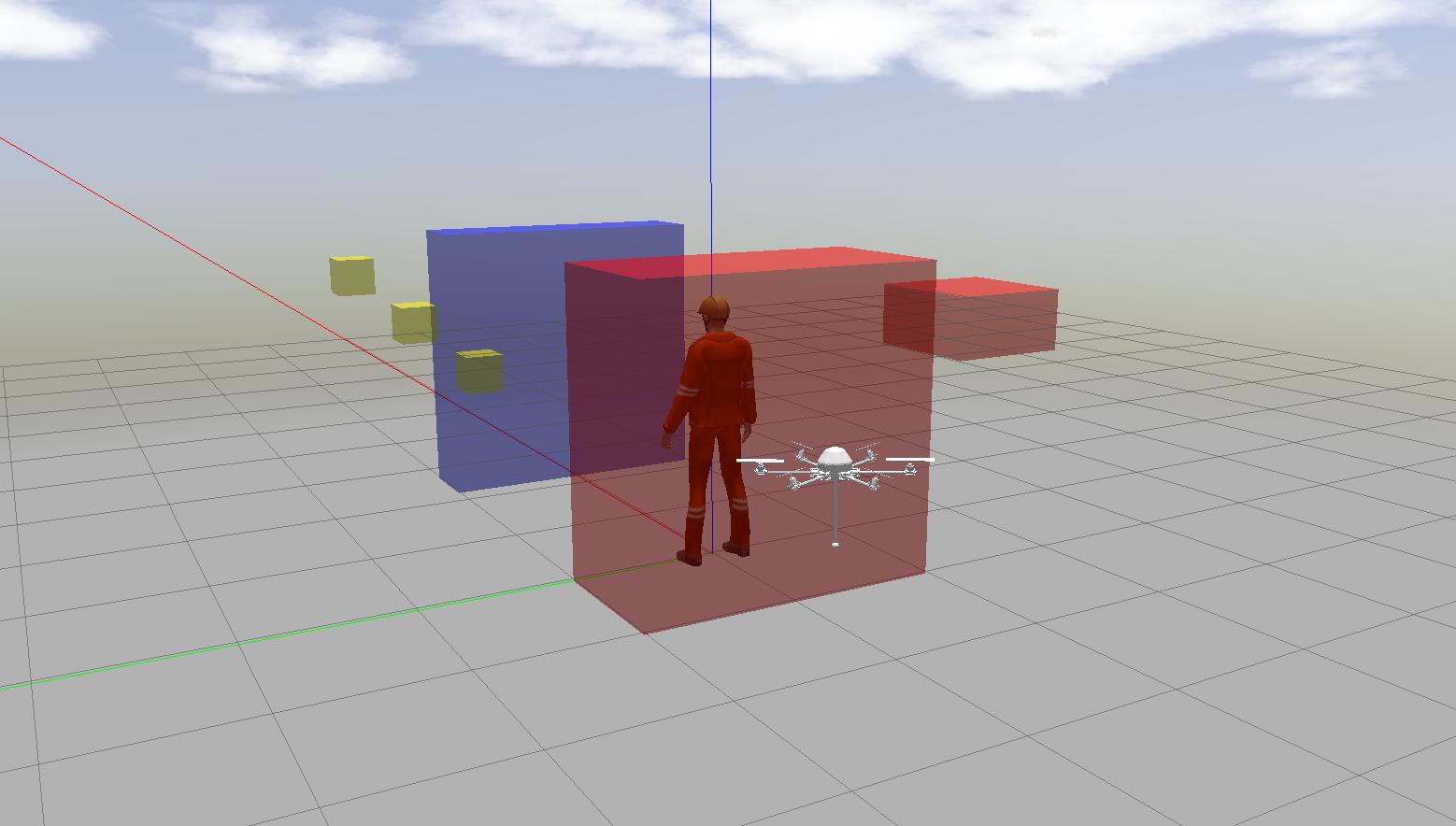}};
    \begin{scope}[x={(img.south east)},y={(img.north west)}]
      % plot some stuff over the image
      %\draw [white, solid, ultra thick] (0.05, 0.275) circle (0.05); % drone1
      %\draw [white, dashed, ultra thick] (0.2250, 0.285) circle (0.05); % drone2
      %\draw [white, dotted, thick] (0.75, 0.61) circle (0.05); % drone3
      \node[imglabel,text=black] (label) at (img.south west) {\scriptsize $t_k=\SI{0}{\second}$};
    \end{scope}
  \end{tikzpicture}
  }
  % \caption{$\mathbf{t}_k=\SI{24}{\second}$}
\end{subfigure}
\hspace*{0.075cm}
\begin{subfigure}{0.45\columnwidth}
  \centering
  % left - bottom - right - top
  \scalebox{0.85}{
  \begin{tikzpicture}
    \node[anchor=south west,inner sep=0] (img) at (0,0) { 
    % left - bottom - right - top
    \adjincludegraphics[trim={{.175\width} {.20\height} {.25\width} 
    {.20\height}},clip,scale=0.185]{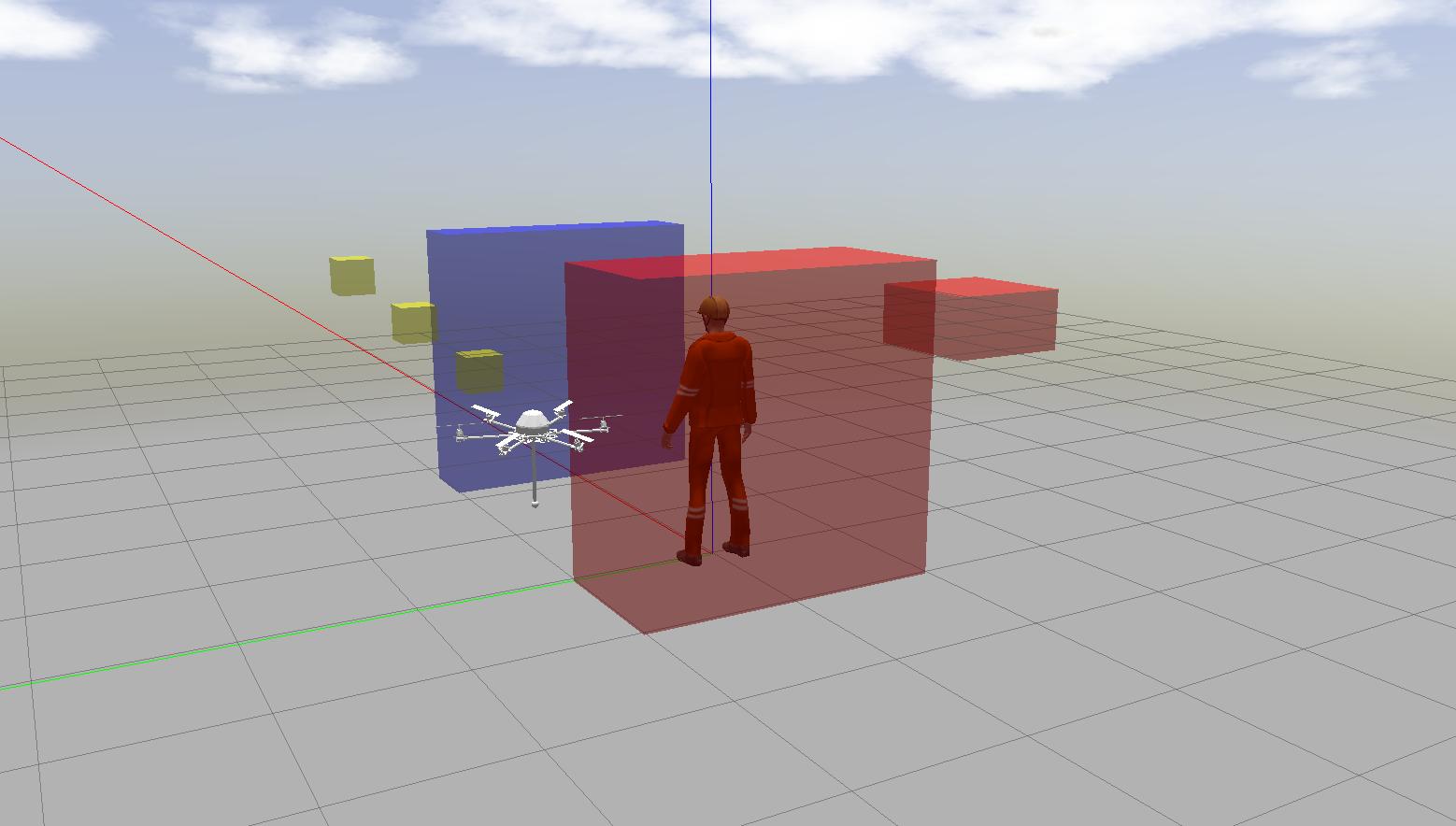}};
    \begin{scope}[x={(img.south east)},y={(img.north west)}]
      % plot some stuff over the image
      %\draw [white, solid, ultra thick] (0.18, 0.625) circle (0.05); % drone1
      %\draw [white, dashed, ultra thick] (0.47, 0.595) circle (0.05); % drone2
      %\draw [white, dotted, thick] (0.525, 0.82) circle (0.05); % drone3
      \node[imglabel,text=black] (label) at (img.south west) {\scriptsize $t_k=\SI{4}{\second}$};
    \end{scope}
  \end{tikzpicture}
  }
  % \caption{$\mathbf{t}_k=\SI{36}{\second}$}
\end{subfigure}
\\
\vspace{0.05cm}
\hspace{0.05cm}
\begin{subfigure}{0.45\columnwidth}
  \centering
  % left - bottom - right - top
  \scalebox{0.85}{
  \begin{tikzpicture}
    \node[anchor=south west,inner sep=0] (img) at (0,0) { 
    % left - bottom - right - top
    \adjincludegraphics[trim={{.175\width} {.20\height} {.25\width} 
    {.20\height}},clip,scale=0.185]{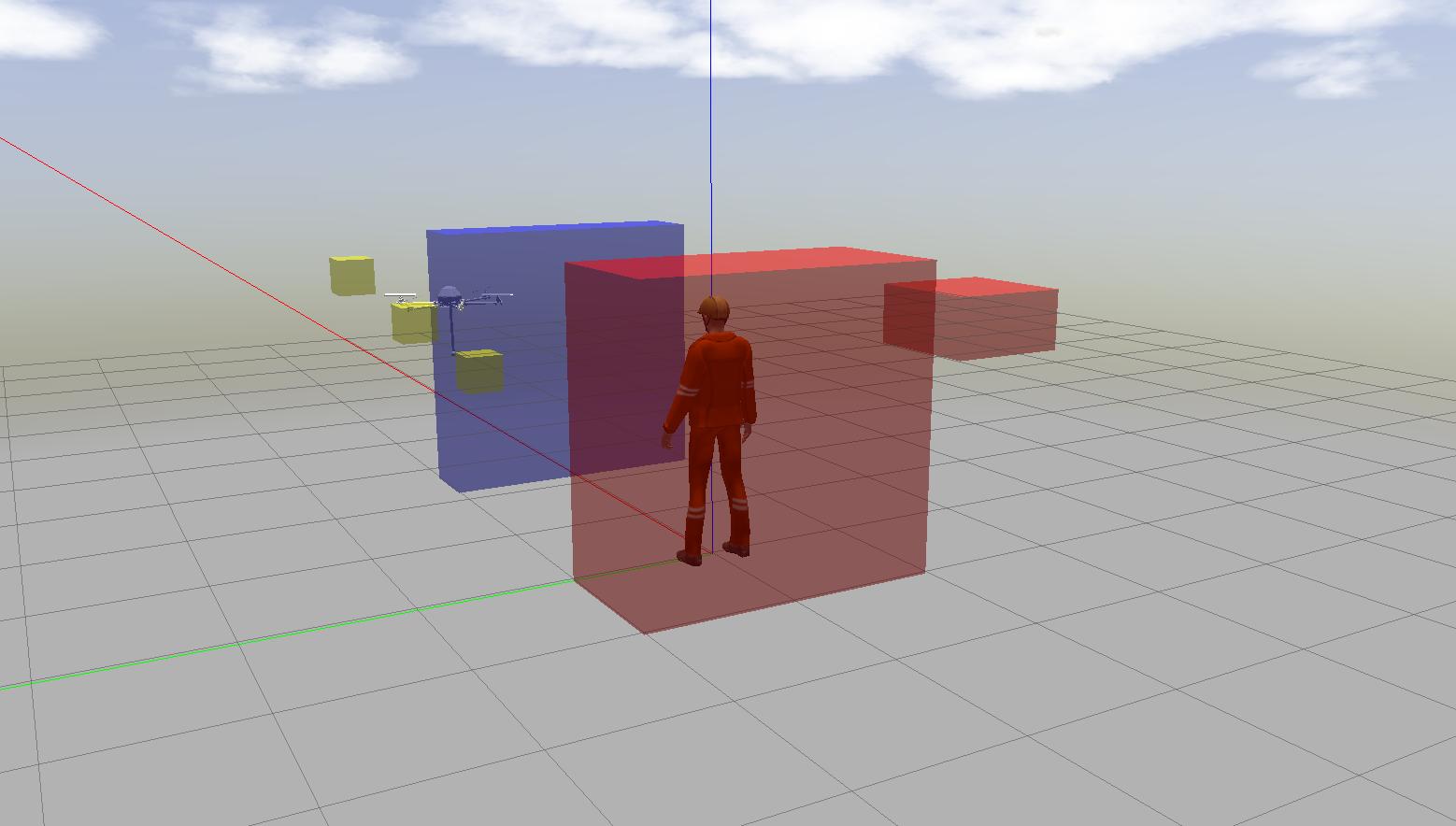}};
    \begin{scope}[x={(img.south east)},y={(img.north west)}]
      % plot some stuff over the image
      %\draw [white, solid, ultra thick] (0.24, 0.64) circle (0.05); % drone1
      %\draw [white, dashed, ultra thick] (0.305, 0.445) circle (0.05); % drone2
      %\draw [white, dotted, thick] (0.94, 0.145) circle (0.05); % drone3
      \node[imglabel,text=black] (label) at (img.south west) {\scriptsize $t_k=\SI{10}{\second}$};
    \end{scope}
  \end{tikzpicture}
  }
  % \caption{$\mathbf{t}_k=\SI{60}{\second}$}
\end{subfigure}
\hspace*{0.05cm}
\begin{subfigure}{0.45\columnwidth}
  \centering
  % left - bottom - right - top
  \scalebox{0.85}{
  \begin{tikzpicture}
    \node[anchor=south west,inner sep=0] (img) at (0,0) { 
    % left - bottom - right - top
    \adjincludegraphics[trim={{.175\width} {.20\height} {.25\width} 
    {.20\height}},clip,scale=0.185]{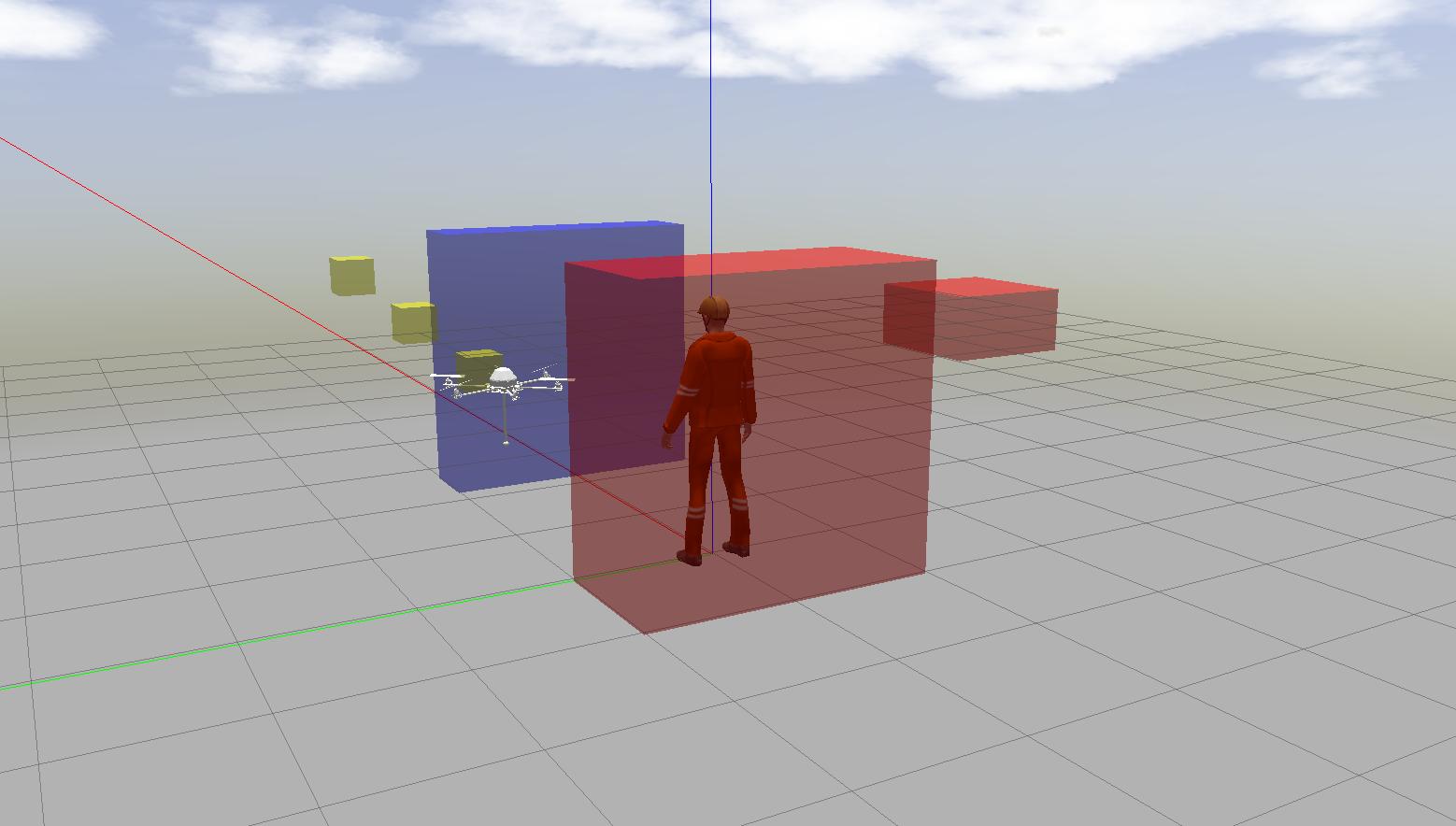}};
    \begin{scope}[x={(img.south east)},y={(img.north west)}]
      % plot some stuff over the image
      %\draw [white, solid, ultra thick] (0.24, 0.61) circle (0.05); % drone1
      %\draw [white, dashed, ultra thick] (0.48, 0.295) circle (0.05); % drone2
      %\draw [white, dotted, thick] (0.94, 0.145) circle (0.05); % drone3
      \node [imglabel,text=black] (label) at (img.south west) {\scriptsize $t_k=\SI{17}{\second}$};
    \end{scope}
  \end{tikzpicture}
  }
  % \caption{$\mathbf{t}_k=\SI{60}{\second}$}
\end{subfigure}
%\vspace{-2mm}
				\vspace{-1.0mm}
				\caption{Snapshots from Gazebo simulations illustrating a left-handed, top-to-bottom preferred approach direction. The blue, red, and yellow regions represent the designated location, no-fly zones, and approach areas, respectively.}
				\label{fig:experimentsTimeLine}
			\end{figure*}

			The optimization algorithm was implemented in MATLAB R2019b using CasADi\footnote{\url{https://web.casadi.org}} and solved with IPOPT\footnote{\url{https://coin-or.github.io/Ipopt/}}. Gazebo simulations relied on the GenoM robotics middleware \cite{Mallet2010ICRA} together with the TeleKyb3 software available through the OpenRobots platform\footnote{\url{https://git.openrobots.org/projects/telekyb3}}. All experiments were executed on a computer equipped with an i7-8565U processor (1.80 GHz) and 32GB of RAM running Ubuntu 20.04. Additional visual material is available at  \url{https://mrs.felk.cvut.cz/stl-ergonomy-risk-analysis}. Figure \ref{fig:experimentsTimeLine} provides representative snapshots from the Gazebo simulations.
			
		\end{sloppypar}
		
		%%% END SECTION ============================================================
		
		%%% START SECTION ==========================================================
		
		\subsection{Object handover scenario}
		\label{sec:objectHandover}
		
		\begin{sloppypar}
			
\begin{figure}[tb]
				\centering
				% left - bottom - right - top
				\adjincludegraphics[width=\columnwidth, trim={{0.07\width} {0.05\height} {0.06\width} {.1\height}}, clip]{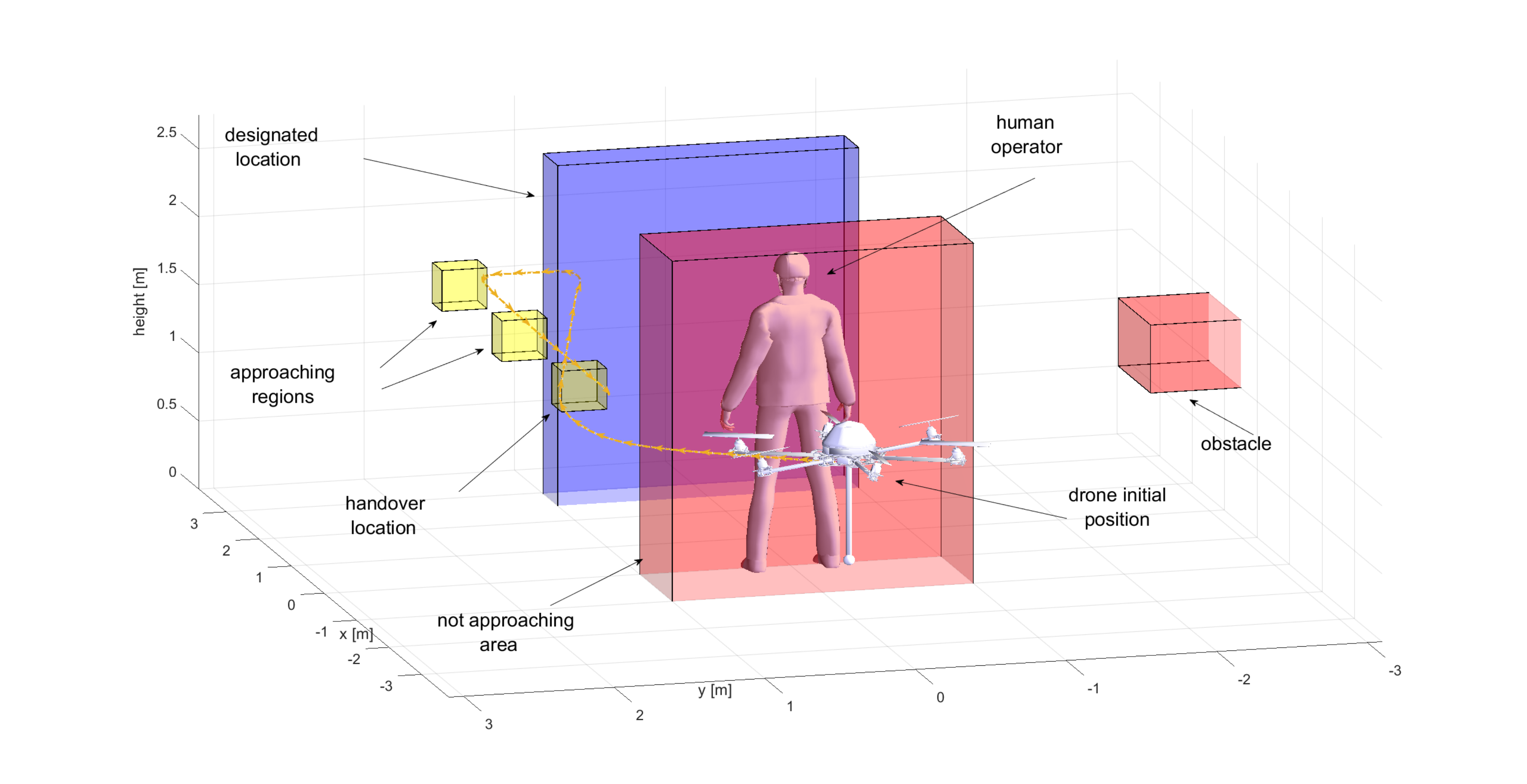}
				\vspace{-2.25em}
				\caption{Planned trajectory for the \ac{MRAV}, showing a left-handed, top-to-bottom preferred approach direction. Arrows indicate the paths followed by the \ac{MRAV} throughout the mission.}
				\label{fig:3Dtrajectory}
			\end{figure}

			The proposed planning strategy was evaluated on the object handover scenario introduced in Section \ref{sec:motivatingExample}. The simulation environment, shown in Figure \ref{fig:3Dtrajectory}, spans $\SI{7.2}{\meter} \times \SI{5.7}{\meter} \times \SI{2.5}{\meter}$ and includes a human operator, an obstacle, and an \ac{MRAV} equipped with a rigidly attached stick for performing the handover task. 
			The rigid attachment prevents pendulum effects during flight, thereby simplifying the planning problem and allowing the analysis to focus on ergonomic collaboration, vehicle dynamics, physical limits, and temporal mission requirements.
			The objective of this section is to verify, on a representative handover mission, that the offline solution simultaneously (i) satisfies safety-critical predicates (workspace containment $\pi_\mathrm{ws}$, obstacle avoidance $\pi_\mathrm{obs}$, and no-approach-from-behind $\pi_\mathrm{beh}$), (ii) fulfills collaboration-oriented specifications (visibility $\pi_\mathrm{vis}$, approach preference $\pi_\mathrm{pr}$, and comfort $\pi_\mathrm{vel}$), and (iii) remains dynamically feasible under the \ac{GTMR} model and actuator bounds. The discussion therefore reports both qualitative proprieties of the planned path and quantitative margins \ac{wrt} the imposed limits.
			
			The parameters used to initialize the optimization problem, summarized in Table~\ref{tab:tableParamters}, serve different roles within the proposed framework. Physical parameters such as mass, inertia, actuator limits, and velocity bounds define hard constraints and directly determine feasibility. Geometric and temporal mission parameters specify the handover location, approach regions, and timing requirements encoded in the \ac{STL} formula $\pi$. The mission duration ($T_N$) and visibility time ($T_\mathrm{vr}$) were intentionally kept short to focus on the critical phases of the trajectory and avoid unnecessary analysis of extended waiting periods. By contrast, parameters such as the robustness threshold $\kappa$, the smoothing factor $\lambda$, and the energy weight $w$ act as design variables that influence conservativeness, smoothness, and robustness margins without altering the system's physical limits. This distinction is important when adapting the framework to new vehicles or mission configurations.
			
			Solving the offline optimization problem \eqref{eq:optimizationProblemSmooth} that embeds the \ac{GTMR} dynamics, actuator bounds, and the full \ac{STL} specification $\pi$ required approximately $\SI{11}{\minute}$ on the reported hardware. This runtime reflects an offline setting aligned with repetitive handover operations: the solution is computed once for a given environment map and operator-preference configuration, whereas online execution relies only on trajectory tracking and, when needed, short-horizon recovery (see Section \ref{sec:comparativeAnalysis}). The use of a smooth robustness approximation is essential to achieve numerical tractability in this nonlinear program; however, it comes at the cost of relinquishing the strict satisfaction guarantees associated with exact \ac{STL} robustness evaluation, thereby introducing a deliberate trade-off between computational feasibility and formal correctness.
			
			The offline runtime is primarily affected by the planning horizon, discretization resolution, and the temporal structure of the \ac{STL} formula, rather than by the sheer number of predicates alone \cite{Raman2014CDC, Raman2015Hybrid}. In particular, deeply nested temporal operators and long horizons dominate computational effort, whereas conjunctions of simple spatial predicates introduce comparatively moderate overhead. The proposed framework targets repetitive and structured missions, and does not aim to provide real-time guarantees or scalability claims for arbitrarily large or unstructured environments.
			
			Figure \ref{fig:3Dtrajectory} illustrates the resulting plan for a left-handed, top-to-bottom approach. The trajectory avoids obstacles ($\pi_\mathrm{obs}$) and the no-approach-from-behind region ($\pi_\mathrm{beh}$), while traversing the preferred approach corridors in the prescribed order ($\pi_\mathrm{pr}$), thereby satisfying both safety and ergonomic constraints.
			
\begin{figure}[H]
				\centering	
				\input{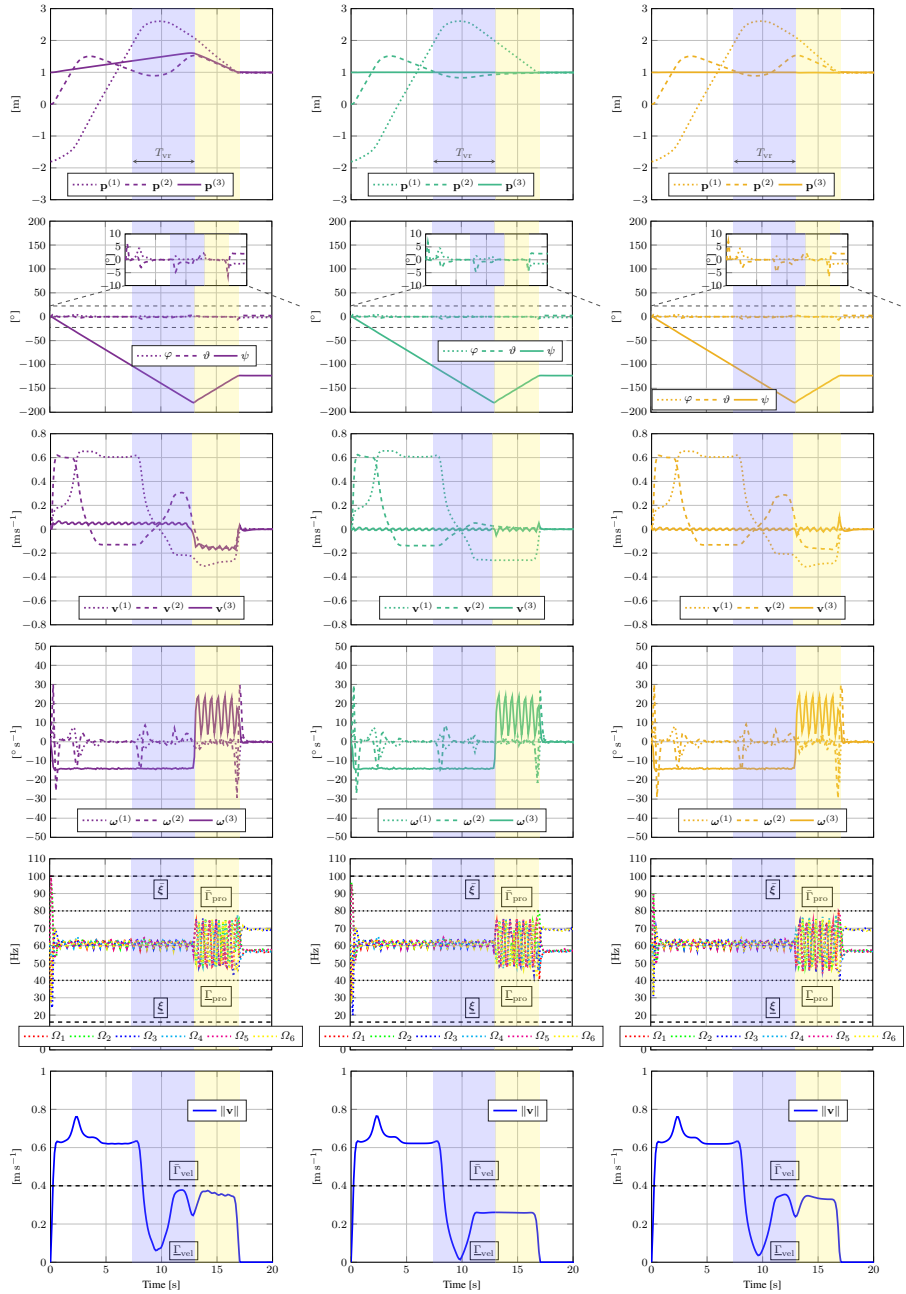}
				\vspace{-1.65mm}
				\caption{Position, orientation, linear and angular velocities, propeller speed, and velocity magnitude for the planned trajectories based on different operator approach preferences: left-handed top-to-bottom approach (left), front approach (center), and left-side approach (right). Blue and yellow segments indicate the time frames for meeting the visibility ($T_\mathrm{vr}$) and handover requirements, respectively.}
				\label{fig:simulationResults}
			\end{figure}

			Figure~\ref{fig:simulationResults} provides a specification-level validation of the same plan in closed-loop Gazebo simulation. The velocity magnitude $\|\mathbf{v}(t_k)\|$ remains within its prescribed bounds $(\underline{\mathrm{\Gamma}}_{\mathrm{vel}}, \bar{\mathrm{\Gamma}}_{\mathrm{vel}})$, supporting the comfort predicate $\pi_{\mathrm{vel}}$, while the propeller-speed traces respect limits derived from both motor forces ($\underline{\bm{\xi}}$ and $\bar{\bm{\xi}}$) and speed constraints ($\underline{\mathrm{\Gamma}}_{\mathrm{pro}}$ and $\bar{\mathrm{\Gamma}}_{\mathrm{pro}}$)\footnote{To improve readability, the graph displays the propeller rotation speeds, incorporating bounds derived from motor force constraints ($\underline{\bm{\xi}}$ and $\bar{\bm{\xi}}$). Although these constraints are originally expressed in terms of forces, the corresponding propeller speeds were calculated by inverting the relationship between motor rotation and forces (see Section \ref{sec:systemModeling}), and the resulting bounds were plotted accordingly. Similarly, for the propeller speed constraints ($\underline{\mathrm{\Gamma}}_\mathrm{pro}$ and $\bar{\mathrm{\Gamma}}_\mathrm{pro}$), originally defined as limits on squared motor speeds, the bounds were converted to equivalent propeller rotation speeds. With a slight abuse of notation, both sets of constraints are represented on the same plot to simplify interpretation and enhance clarity.}, confirming compliance with actuation and comfort specifications $\pi_\mathrm{pro}$. The blue and yellow shaded intervals correspond to the temporal operators in the \ac{STL} formula: the designated-location requirement $\pi_{\mathrm{vr}}$ is satisfied within its allotted time window, whereas the handover $\pi_{\mathrm{ho}}$ is achieved afterward, consistent with the encoded sequencing. The latter interval varies across approach preferences because no strict temporal bound is imposed, allowing the planner to adapt the final phase of the maneuver to improve ergonomics. Throughout execution, heading alignment remains within the maneuverability margin $\gamma$, supporting $\pi_{\mathrm{vis}}$ and ensuring continuous visual contact between the operator and the end-effector, while all motor forces stay within admissible bounds $(\underline{\bm{\xi}},\bar{\bm{\xi}})$, confirming dynamic feasibility under the \ac{GTMR} model.
			
			A comparison across approach preferences (see Figure~\ref{fig:simulationResults}) further indicates that the planner adapts the transient motion mainly through changes in the approach corridors $\pi_{\mathrm{pr}}$, while preserving similar comfort and actuation envelopes. In particular, velocity and propeller-speed profiles remain below their limits for all preferences, whereas the time spent inside the designated region varies with corridor geometry. This behavior is consistent with the role of the approach-preference $\pi_{\mathrm{pr}}$ as an ergonomics-driven constraint that shapes the path without relaxing safety or feasibility requirements.
			
			These results highlight the successful integration of ergonomic objectives, nonlinear dynamics, and actuator constraints within the \ac{STL}-based planning framework, ensuring mission completion while preserving operator comfort and system safety.
			
			Finally, as discussed in Sections \ref{sec:smoothApproximation} and \ref{sec:motionPlanner}, the \ac{STL}-constrained optimization problem \eqref{eq:optimizationProblem} was solvable only through the smooth approximation $\tilde{\rho}_\pi(\mathbf{x})$, which mitigates the non-differentiability introduced by Boolean and temporal operators and enables local optimization.
			
			While a systematic sensitivity analysis is beyond the scope of this paper, the reported simulations illustrate how modifying selected design parameters, such as the energy weight $w$ or the uncertainty covariance used in the risk analysis, affects trajectory shape, robustness margins, and replanning behavior, whereas feasibility remains governed primarily by physical and safety constraints.
			
		\end{sloppypar}
		
		%%% END SECTION ============================================================
		
		%%% START SECTION ==========================================================
		
		\subsection{Energy-aware analysis and replanning strategy}
		\label{sec:comparativeAnalysis}
		
\begin{figure}[tb]
			\centering
			% left - bottom - right - top
			\adjincludegraphics[width=\columnwidth, trim={{0.07\width} {0.05\height} {0.06\width} {.1\height}}, clip]{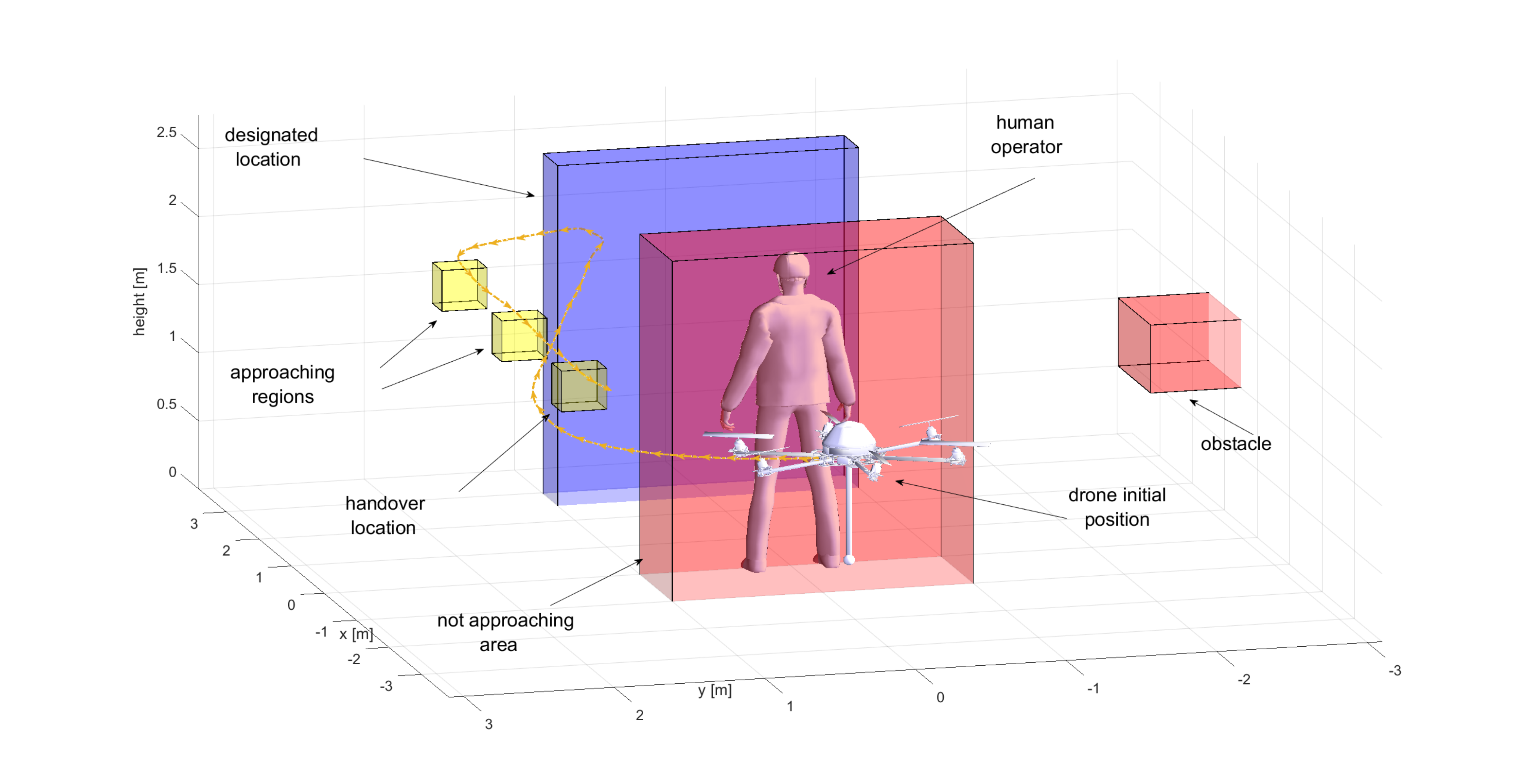}
			\vspace{-2.25em}
			\caption{Planned trajectory for the \ac{MRAV}, showing a left-handed, top-to-bottom preferred approach direction, with the energy term excluded from the optimization problem (i.e., setting $w=0$ in \eqref{eq:optimizationProblemSmooth}). Arrows indicate the paths followed by the \ac{MRAV} throughout the mission.}
			\label{fig:energyAware3D}
		\end{figure}

\begin{figure}[tb]
			\centering
			% left - bottom - right - top
			\adjincludegraphics[width=\columnwidth, trim={{0.07\width} {0.05\height} {0.06\width} {.1\height}}, clip]{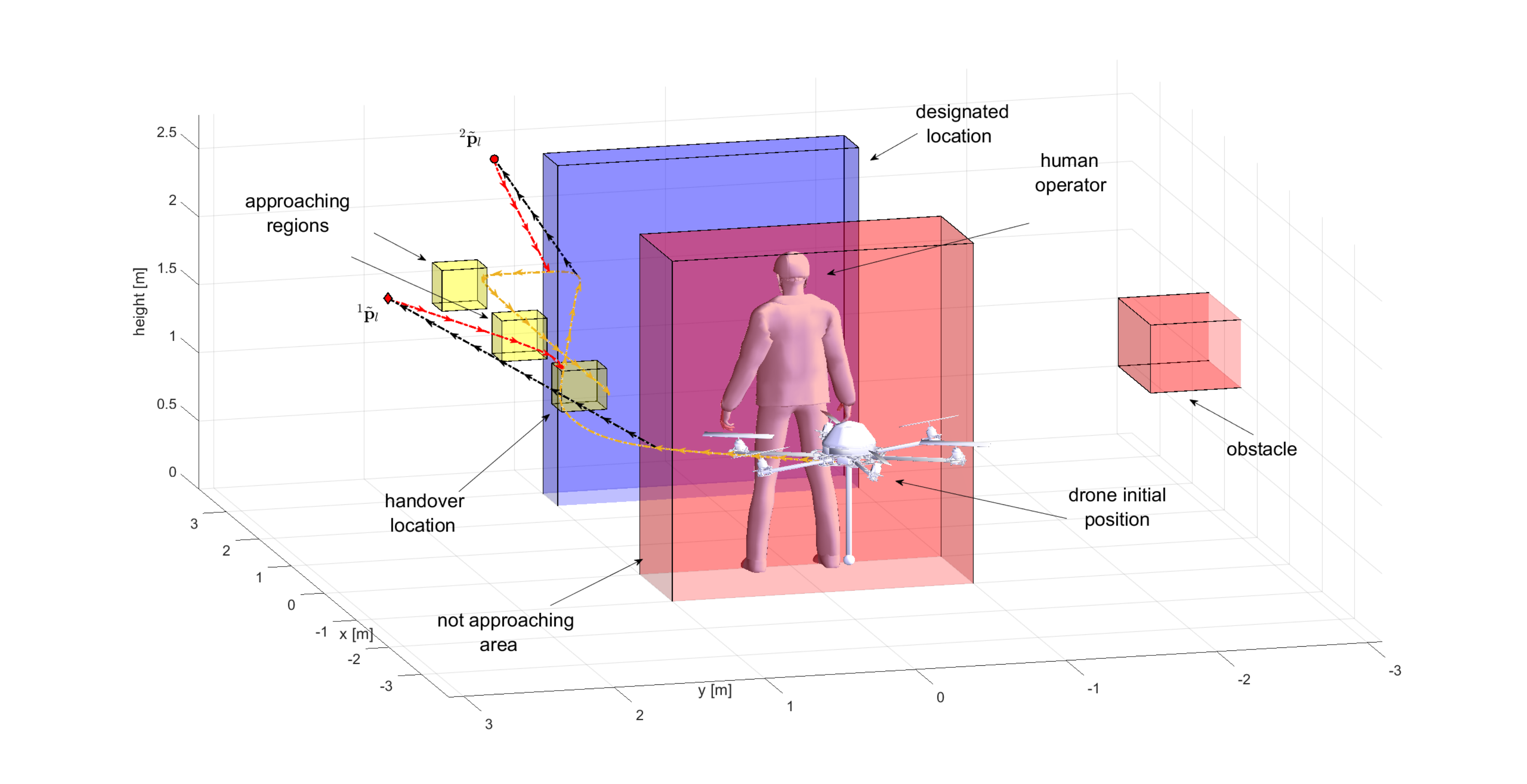}
			\vspace{-2.25em}
			\caption{Event-driven replanner trajectories for two disturbance scenarios, showing the \ac{MRAV}'s left-handed approach. Black paths indicate deviations from the original trajectory, red paths show updated trajectories from the replanner, and red markers denote the \ac{MRAV}'s position ($^1\tilde{\mathbf{p}}_l$ and $^2\tilde{\mathbf{p}}_l$) when replanning is triggered.}
			\label{fig:replanner3D}
		\end{figure}

		This section evaluates two complementary aspects of the proposed framework. First, it analyzes the effect of the energy-related term in \eqref{eq:optimizationProblemSmooth} on the structure of the planned trajectories by comparing energy-aware solutions with those defined for $w=0$, while verifying that satisfaction of the \ac{STL} specification $\pi$ is preserved. Second, it assesses whether the proposed robustness-aware event-driven replanner is able to restore feasibility after disturbances while enforcing only the safety-critical subset of \ac{STL} constraints, as described in Section~\ref{sec:replanner}. Results from MATLAB numerical simulations illustrating both aspects are reported in Figures~\ref{fig:energyAware3D} and~\ref{fig:replanner3D}.
		
		Figure \ref{fig:energyAware3D} illustrates that trajectories planned without the energy term $\mathcal{L}(\mathbf{x})$ in \eqref{eq:optimizationProblemSmooth} tend to be longer and exhibit sharper direction changes compared to those incorporating energy (as shown in Figure \ref{fig:3Dtrajectory}).  This difference arises because, with the inclusion of the energy term, the optimization problem \eqref{eq:optimizationProblemSmooth} enforces stricter constraints on allowable actuator rotational speeds, as well as on linear and angular velocities, than in cases where energy considerations are omitted ($w=0$). Quantitatively, the energy-aware solution exhibits lower peak propeller speeds and reduced velocity transients compared to the case $w=0$, while maintaining $\tilde{\rho}_\pi(\mathbf{x}) \ge \kappa$ for the full mission horizon. This confirms that the energy regularization reshapes the solution toward smoother actuation profiles without sacrificing satisfaction of the STL specification. This approach enhances \ac{MRAV} endurance by reducing overall energy consumption while still meeting mission objectives. 
		
		\begin{figure}[tb]
			\centering	
			\input{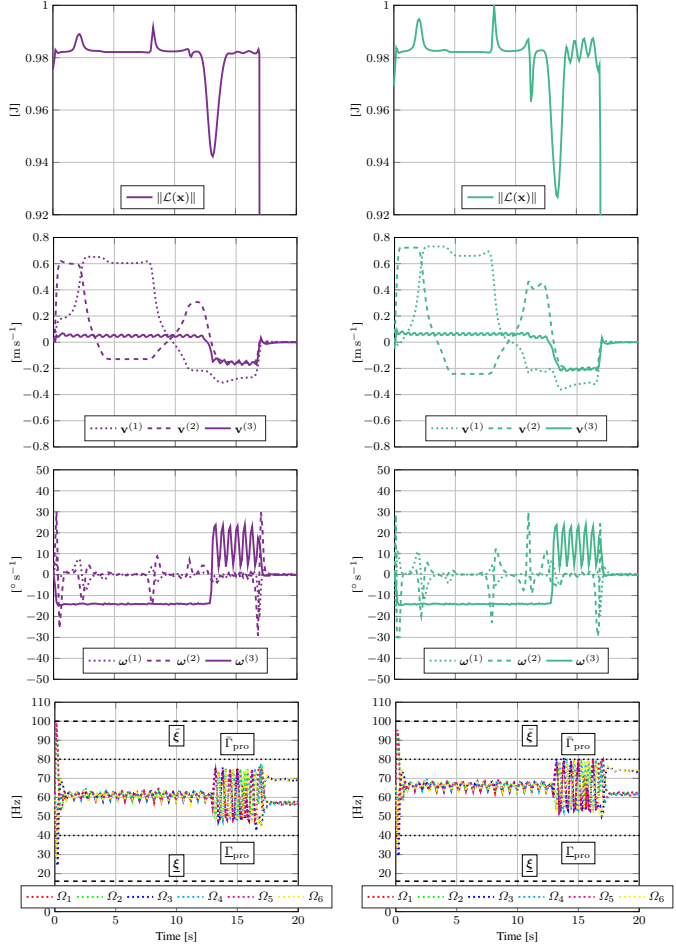}
			\vspace{-1.65mm}
			\caption{Comparison of the normalized energy term, \ac{MRAV} linear and angular velocities, and propeller speed for scenarios where the energy term is either included (left) or excluded (right) in the optimization problem \eqref{eq:optimizationProblemSmooth}.}
			\label{fig:comparisonEnergy}
		\end{figure}
		
		Figure \ref{fig:comparisonEnergy} compares energy consumption values between scenarios with and without the energy term $\mathcal{L}(\mathbf{x})$ in the \ac{STL} optimization problem \eqref{eq:optimizationProblemSmooth}, based on data obtained from Gazebo simulations. The normalized energy term, along with \ac{MRAV} linear and angular velocities and propeller speeds, are plotted for both cases. As shown, trajectories without energy optimization result in higher linear and angular velocities and increased propeller speeds (left plots in Figure \ref{fig:comparisonEnergy}), making them less energy-efficient than energy-aware trajectories (right plots in Figure \ref{fig:comparisonEnergy}). Importantly, in both cases, the mission requirements encoded in the \ac{STL} formula $\pi$ are fully met. The constraint \eqref{eq:robustTresholdOP} ensures a minimum robustness level $\kappa$ to satisfy mission objectives, even when energy optimization might slightly reduce robustness values.
		
		Taken together, Figures~\ref{fig:energyAware3D} and~\ref{fig:comparisonEnergy} illustrate that the energy-related term acts primarily as a regularizer on actuator usage rather than as a constraint-enforcing mechanism: all \ac{STL} predicates remain satisfied in both cases, but the optimized solution distributes thrust and velocity changes more evenly over time when energy is included. This behavior is consistent with the design goal of extending endurance in repetitive collaborative tasks without altering mission logic.
		
		To assess the performance of the robustness-aware replanner, MATLAB simulations were conducted in which unexpected disturbances caused the \ac{MRAV} to deviate from its planned trajectory. These disturbances emulate realistic effects such as wind gusts, hardware malfunctions, or operator-induced variability, as described in Section~\ref{sec:replanner}. Replanning is triggered when either the deviation from the nominal path exceeds a threshold, $\lVert \tilde{\mathbf{p}}_l - \mathbf{p}_l^\star \rVert > \zeta$, or when the predicted robustness margin of the safety-critical \ac{STL} subset falls below a prescribed value, $\hat{\rho}_{\pi_\mathrm{crit}}(\tilde{\mathbf{x}}_l) < \kappa_\mathrm{crit}$. In both cases, a short-horizon online optimization is solved to reconnect the vehicle to the next ``topic'' waypoint, as illustrated in Figure~\ref{fig:replanner3D}.
		During recovery, only the safety-critical subset of the \ac{STL} specification is enforced, while comfort- and ergonomics-related predicates are temporarily relaxed to prioritize feasibility and collision avoidance.The newly computed trajectory is then provided as a reference to the tracking controller (see Figure~\ref{fig:controlScheme}). 
		Across all tested disturbance events, the replanning optimization consistently completed within sub-second to few-second times. Figure~\ref{fig:replanningTimes} reports a box plot of execution times over $50$ replanning runs under varying disturbance realizations, showing tightly clustered runtimes with a median of $\SI{0.745}{\second}$ and a maximum of $\SI{2.65}{\second}$.
		
		% Define the data with explicit row separators (\\)
		\pgfplotstableread[row sep=\\]{
			time \\
			0.34 \\ 0.38 \\ 0.41 \\ 0.42 \\ 0.45 \\ 0.47 \\ 0.48 \\ 0.49 \\ 0.51 \\ 0.52 \\
			0.54 \\ 0.55 \\ 0.56 \\ 0.57 \\ 0.59 \\ 0.60 \\ 0.62 \\ 0.63 \\ 0.64 \\ 0.65 \\
			0.67 \\ 0.68 \\ 0.70 \\ 0.72 \\ 0.74 \\ 0.75 \\ 0.77 \\ 0.79 \\ 0.81 \\ 0.83 \\
			0.86 \\ 0.88 \\ 0.91 \\ 0.94 \\ 0.97 \\ 1.01 \\ 1.05 \\ 1.09 \\ 1.14 \\ 1.20 \\
			1.27 \\ 1.35 \\ 1.44 \\ 1.55 \\ 1.68 \\ 1.82 \\ 1.98 \\ 2.15 \\ 2.38 \\ 2.65 \\
		}\replanningdata
		
		\begin{figure}[h]
			\centering
			\begin{tikzpicture}
				\begin{axis}[
					ymajorgrids,
					ylabel={Execution time [s]},
					xtick={1},
					xticklabels={Replanning [-]},
					width=7cm,
					height=6cm,
					boxplot/draw direction=y, 
					]
					
					\addplot+[
					black,
					thick,
					boxplot prepared={
						draw position=1,
						lower whisker=0.34,
						lower quartile=0.5625,
						median=0.745,
						upper quartile=1.08,
						upper whisker=1.82,
						average=0.881, 
						every box/.style={draw=blue!50!black, fill=blue!15},
						every whisker/.style={draw=black, thick},
						every median/.style={draw=red, ultra thick}, 
					},
					] coordinates {}; 
					
					\addplot[only marks, mark=*, mark size=1.5pt, black] 
					coordinates {(1,1.98) (1,2.15) (1,2.38) (1,2.65)};
					
				\end{axis}
			\end{tikzpicture}
			\vspace{-1em}
			\caption{Distribution of online replanning execution times over 50 runs under varying environmental perturbations. The blue box denotes the interquartile range, the red line indicates the median, and whiskers extend to the non-outlier extrema. Black dots mark outlier executions.}
			\label{fig:replanningTimes}
		\end{figure}
		
		Replanning time is reported as a performance metric because the recovery mechanism is intended to operate during execution; sub-second to few-second computation is therefore required for practical deployment in close-proximity aerial manipulation scenarios.
		
		Overall, the results in this section indicate that the offline energy-aware optimization and the online robustness-aware replanner play complementary roles: the former shapes the nominal solution toward endurance-oriented and smooth behaviors, whereas the latter restores feasibility under disturbances using a computationally lightweight, specification-aware recovery strategy. Together, these two layers support repeated execution of collaborative handover missions under realistic operating conditions.
		
		%%% END SECTION ============================================================
		
		%%% START SECTION ==========================================================
		
		\subsection{Uncertainty-aware risk assessment}
		\label{sec:riskAssessment}
		
		\begin{sloppypar}
			
			As outlined in Section \ref{sec:uncertaintyAwareRiskAnalysis}, this section presents a numerical assessment of the proposed uncertainty-aware risk analysis framework. The objective is not to redesign the trajectory online, but to quantify how uncertainty in human pose degrades the satisfaction margins of the \ac{STL} specification and to provide interpretable risk indicators that can inform whether re-execution of the planning problem is warranted. The analysis therefore focuses on how the distribution of the smooth robustness score evolves as uncertainty increases and how tail-risk metrics summarize this degradation.
			
			The evaluation considers trajectories obtained from the optimization problem under stochastic human pose perturbations and investigates how far the operator can deviate from the nominal handover configuration before mission requirements encoded in the \ac{STL} formula are violated. VaR and CVaR are adopted as performance metrics to characterize worst-case and tail behavior of the robustness distribution, providing more informative risk indicators than mean robustness alone in safety-critical collaborative settings.
			
			The simulations were conducted in MATLAB, where uncertainty in human pose was modeled as a stochastic process, $Y$, following a Gaussian distribution $\mathcal{N}(\mu_z, \sigma_z)$. This  stochastic model captures variability in human position and orientation, which directly influences mission-critical predicates, including the handover location ($\pi_\mathrm{ho}$), approach direction ($\pi_\mathrm{pr}$), and the no-approach-from-behind constraint ($\pi_\mathrm{beh}$). A total of $K=\num{15,000}$ realizations of the stochastic process $Y$ were generated to simulate diverse human poses during the mission. This number balances the computational cost of generating $Y(\cdot, \varepsilon)$ and ensures reliable, non-conservative estimates.
			
			Four random vectors, $^i\mathbf{p}_\mathrm{hum}$, with $i=\{1,2,3,4\}$, were employed to model variations in the human position, while the human attitude, $\bm{\eta}_\mathrm{hum}$, was fixed to maintain consistent metrics for trajectory performance evaluation. Although a similar analysis could involve fixing the position and varying the attitude, human positional uncertainty was prioritized due to its greater impact on trajectory validity, as well as on ergonomic and comfort requirements. The random vectors for positional uncertainty were defined as follows:
			\begin{equation}\label{eq:humanPoses}
				\resizebox{0.92\hsize}{!}{$%
					\begin{split}
						&^1\mathbf{p}_\mathrm{hum} \sim \mathcal{N} 
						\begin{pmatrix}
							\begin{bmatrix} 0 \\ 0 \\ 0 \end{bmatrix},
							\begin{bmatrix} 0.125 & 0 & 0\\ 0 & 0.125 & 0 \\ 0 & 0 & 0 \end{bmatrix}
						\end{pmatrix},
						&^2\mathbf{p}_\mathrm{hum} \sim \mathcal{N} 
						\begin{pmatrix}
							\begin{bmatrix} 0 \\ 0 \\ 0 \end{bmatrix},
							\begin{bmatrix} 0.25 & 0 & 0\\ 0 & 0.25 & 0 \\ 0 & 0 & 0 \end{bmatrix}
						\end{pmatrix},
						\\
						&^3\mathbf{p}_\mathrm{hum} \sim \mathcal{N} 
						\begin{pmatrix}
							\begin{bmatrix} 0 \\ 0 \\ 0 \end{bmatrix},
							\begin{bmatrix} 0.5 & 0 & 0\\ 0 & 0.5 & 0 \\ 0 & 0 & 0 \end{bmatrix}
						\end{pmatrix},
						&^4\mathbf{p}_\mathrm{hum} \sim \mathcal{N} 
						\begin{pmatrix}
							\begin{bmatrix} 0 \\ 0 \\ 0 \end{bmatrix},
							\begin{bmatrix} 0.75 & 0 & 0\\ 0 & 0.75 & 0 \\ 0 & 0 & 0 \end{bmatrix}
						\end{pmatrix}.
					\end{split}
					$}%
			\end{equation}
			
			The human attitude, $\bm{\eta}_\mathrm{hum}$, was modeled using a Gaussian distribution $\mathcal{N}(\mu_z, \sigma_z)$, with mean vector $\bm{\mu}_z = (0.07,0.07,0.09)^\top$ and covariance $\bm{\sigma}_z = (0.09, 0.09, 0.52)^\top$.
			
			This analysis assumes that human pose uncertainty is modeled exogenously through predefined stochastic distributions; modeling human–robot coupling or reactive human behavior is beyond the scope of this work and could affect robustness estimates in more tightly interactive scenarios.
			
\begin{figure}[tb]
				\centering	
				\input{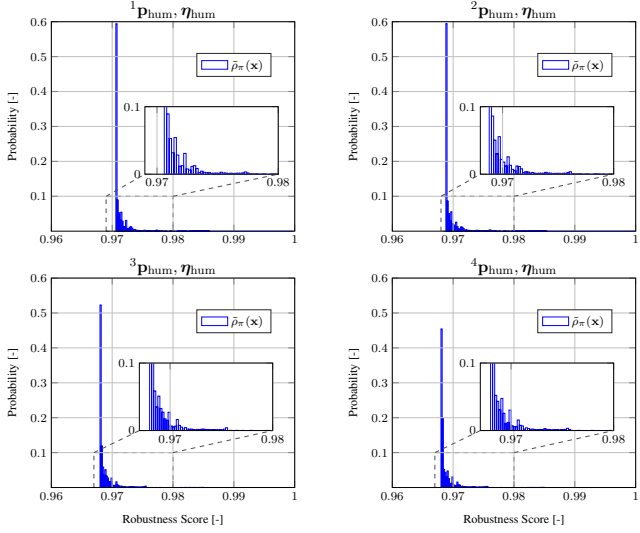}
				\vspace{-1.65mm}
				\caption{Histogram of the smooth robustness score $\tilde{\rho}_\pi(\mathbf{x})$ of the specification $\pi$ in \eqref{eq:stlFormulaProblem}, accounting for uncertainty in human pose ($^i\mathbf{p}_\mathrm{hum}, \bm{\eta}_\mathrm{hum}$) for the $K$ realizations.}
				\label{fig:histogram}
			\end{figure}

			Following the three-step process detailed in Section \ref{sec:uncertaintyAwareRiskAnalysis}, the smooth robustness score $\tilde{\rho}_\pi(\mathbf{x})$ was computed using the trajectory presented in Section \ref{sec:objectHandover} and illustrated in Figure \ref{fig:3Dtrajectory}. This score quantifies the degree to which the \ac{MRAV} satisfies the \ac{STL} formula $\pi$ under a specific realization $Y(\cdot, \varepsilon)$ of the stochastic process, indicating how close the system is to either violating or satisfying the mission objectives. The resulting histograms of $\tilde{\rho}_\pi(\mathbf{x})$ for the $K$ realizations are presented in Figure \ref{fig:histogram}. The data show that for lower covariance values in the human position, the probability of achieving higher robustness scores increases. Conversely, as covariance increases, the likelihood of obtaining lower robustness scores rises.
			
			Risk measures, including $\mathrm{VaR}_\beta$, were applied to quantify the likelihood of not satisfying the \ac{STL} formula $\pi$. Table \ref{tab:VarValues} presents the upper and lower bounds of $\mathrm{VaR}_\beta$ for various risk levels $\beta$. The difference between these bounds, $\lvert \overline{\mathrm{VaR}}_\beta - \underline{\mathrm{VaR}}_\beta \rvert$, remains small across all $\beta$, indicating that the estimates are tight.
			
			To interpret $\mathrm{VaR}_\beta$, consider it as the threshold value below which the robustness score $\tilde{\rho}_\pi(\mathbf{x})$ is unlikely to fall with a probability of at least $\beta$. For example, a $\mathrm{VaR}_{0.80}$ of $0.95811$ for ($^1\mathbf{p}_\mathrm{hum}, \bm{\eta}_\mathrm{hum}$) indicates that there is an 80\% chance the robustness score will not drop below this value. Values lower than this threshold represent a 20\% risk of violating the \ac{STL} specifications $\pi$, providing a quantifiable measure of the system's tolerance to uncertainty.
			
			The histograms in Figure \ref{fig:histogram} and the values in Table \ref{tab:VarValues} collectively suggest that poses with larger covariance, such as ($^3\mathbf{p}_\mathrm{hum}, \bm{\eta}_\mathrm{hum}$) and ($^4\mathbf{p}_\mathrm{hum}, \bm{\eta}_\mathrm{hum}$), are less favorable. This is reflected in the lower robustness scores, indicating higher risks of not meeting mission objectives. These trends indicate that increasing uncertainty primarily shifts probability mass toward lower robustness values rather than uniformly scaling the distribution, which is consistent with violations being driven by localized geometric effects (e.g., approach-direction or handover-location predicates). From a decision-making perspective, this motivates monitoring tail-risk metrics rather than mean robustness alone, since rare but critical realizations dominate safety considerations in collaborative scenarios.
			
			\begin{table}[tb]
				\centering
				\caption{Estimated $\mathrm{VaR}_\beta$ values for different risk levels $\beta$ and $K$ realizations, considering the human poses defined in \eqref{eq:humanPoses}.}
				\label{tab:VarValues}
				\vspace{-1.2em}
				\renewcommand{\arraystretch}{1.15}
				\begin{adjustbox}{max width=0.98\columnwidth}
					\begin{tabular}{l|c|c|c|c|c|c}
						\cmidrule{2-7}
						& $\underline{\mathrm{VaR}}_{0.70}$ & $\underline{\mathrm{VaR}}_{0.80}$ &  $\underline{\mathrm{VaR}}_{0.90}$ & $\overline{\mathrm{VaR}}_{0.70}$ & $\overline{\mathrm{VaR}}_{0.80}$ &  $\overline{\mathrm{VaR}}_{0.90}$ \\
						\cmidrule{2-7} \hline
						($^1\mathbf{p}_\mathrm{hum}, \bm{\eta}_\mathrm{hum}$) & 0.95771 & 0.95811 & 0.95891 & 0.98429 & 0.98469 & 0.98549 \\
						($^2\mathbf{p}_\mathrm{hum}, \bm{\eta}_\mathrm{hum}$) & 0.95591 & 0.95631 & 0.95731 & 0.98249 & 0.98289 & 0.98389 \\
						($^3\mathbf{p}_\mathrm{hum}, \bm{\eta}_\mathrm{hum}$) & 0.95511 & 0.95571 & 0.95651 & 0.98169 & 0.98229 & 0.98309 \\
						($^4\mathbf{p}_\mathrm{hum}, \bm{\eta}_\mathrm{hum}$) & 0.95511 & 0.95571 & 0.95651 & 0.98169 & 0.98229 & 0.98309 \\
						\hline
					\end{tabular}
				\end{adjustbox}
			\end{table}
			
			\begin{table}[tb]
				\centering
				\caption{Estimated $\mathrm{CVaR}_\beta$ values for different risk levels $\beta$ and $K$ realizations, considering the human poses defined in \eqref{eq:humanPoses}.}
				\label{tab:CVarValues}
				\vspace{-1.2em}
				\renewcommand{\arraystretch}{1.15}
				\begin{adjustbox}{max width=0.98\columnwidth}
					\begin{tabular}{l|c|c|c|c|c|c}
						\cmidrule{2-7}
						& $\underline{\mathrm{CVaR}}_{0.70}$ & $\underline{\mathrm{CVaR}}_{0.80}$ &  $\underline{\mathrm{CVaR}}_{0.90}$ & $\overline{\mathrm{CVaR}}_{0.70}$ & $\overline{\mathrm{CVaR}}_{0.80}$ &  $\overline{\mathrm{CVaR}}_{0.90}$ \\
						\cmidrule{2-7} \hline
						($^1\mathbf{p}_\mathrm{hum}, \bm{\eta}_\mathrm{hum}$) & 0.015711 & 0.015311 & 0.014511 & 0.042289 & 0.041889 & 0.041089 \\
						($^2\mathbf{p}_\mathrm{hum}, \bm{\eta}_\mathrm{hum}$) & 0.017511 & 0.017111 & 0.016111 & 0.044089 & 0.043689 & 0.042689 \\
						($^3\mathbf{p}_\mathrm{hum}, \bm{\eta}_\mathrm{hum}$) & 0.018311 & 0.017711 & 0.016911 & 0.044889 & 0.044289 & 0.043489 \\
						($^4\mathbf{p}_\mathrm{hum}, \bm{\eta}_\mathrm{hum}$) & 0.018311 & 0.017711 & 0.016911 & 0.044889 & 0.044289 & 0.043489 \\
						\hline
					\end{tabular}
				\end{adjustbox}
			\end{table}
			
			In Table \ref{tab:CVarValues}, $\mathrm{CVaR}_\beta$ values are estimated for various risk levels $\beta$. Similar observations regarding human poses in \eqref{eq:humanPoses} can be made. However, the values are generally lower because $\mathrm{CVaR}$ considers only the tail of the $F_{\tilde{\rho}_\pi}(\tilde{\rho}_\pi(\mathbf{x}))$ (see Figure \ref{fig:expectedValue}). Notably, the estimates $\underline{\mathrm{CVaR}}_\beta$ of $\mathrm{CVaR}$ are less tight than those for $\mathrm{VaR}_\beta$, with the difference $\lvert \overline{\mathrm{CVaR}}_\beta - \underline{\mathrm{CVaR}}_\beta \rvert$ increasing significantly for larger $\beta$, primarily due to division by $1-\beta$ (see Section \ref{sec:randomVariableStocasticProcess}).
			
			In practice, these risk indicators can be used to define admissible operating envelopes for repeated handovers. For instance, if $\mathrm{VaR}_{0.90}$ (see Table \ref{tab:VarValues}) drops below a prescribed robustness margin, the corresponding human-pose uncertainty level can be interpreted as exceeding the range for which the nominal plan is acceptable, thereby motivating a new offline planning phase or a modification of the mission parameters.
			
			This analysis highlights the extent to which trajectories generated by \eqref{eq:optimizationProblemSmooth} remain compliant with mission requirements under human pose uncertainty. As deviations in human position increase, the smooth robustness score $\tilde{\rho}_\pi(\mathbf{x})$ decreases, providing a quantitative measure of the system’s sensitivity to variability. At the same time, the combined use of robustness distributions and tail-risk metrics offers interpretable indicators of when safety margins erode, thereby supporting principled decisions on whether the nominal plan remains acceptable or should be recomputed. Overall, the uncertainty-aware framework demonstrates that substantial robustness margins can be maintained for moderate levels of uncertainty, while explicitly characterizing the conditions under which replanning or offline re-optimization becomes necessary, complementing the robustness-aware recovery mechanism introduced in Section~\ref{sec:comparativeAnalysis}.
			
		\end{sloppypar}
		
		%%% END SECTION ============================================================
		
		%%% START SECTION ==========================================================
		
		\subsection{Gazebo simulations}
		\label{sec:gazeboSimulations}
		
		Flight tests conducted in the Gazebo simulator (see Figure \ref{fig:experimentsTimeLine}) successfully validated the object handover task encoded by the \ac{STL} formula \eqref{eq:stlFormulaProblem}. The purpose of these simulations is to assess the practical viability of the proposed STL-based planning and robustness-aware replanning framework when deployed in a physics-based simulator with onboard low-level control, rather than to perform a detailed tracking-performance study. In particular, this section examines whether the trajectories generated offline remain feasible under full vehicle dynamics, whether actuator limits are respected in closed loop, and whether the online replanner is able to recover from disturbances while enforcing safety-critical specifications. The simulations demonstrated adherence to critical physical constraints, including motor force limits ($\underline{\bm{\xi}}$ and $\bar{\bm{\xi}}$), while ensuring compliance with \ac{MRAV} dynamics. Ergonomic and comfort features, such as propeller velocity constraints ($\pi_\mathrm{vel}$) and restrictions on the drone's approaching velocity ($\pi_\mathrm{pro}$), were also thoroughly evaluated. Furthermore, the visibility requirement ($\pi_\mathrm{vis}$) was assessed, ensuring the drone consistently aligned its heading with its direction of movement, allowing the operator to maintain continuous visual contact with the small object on the stick. Simulation results are summarized in Figure \ref{fig:simulationResults}, with parameter values for the optimization problem detailed in Table \ref{tab:tableParamters}. The closed-loop executions indicate that the planned trajectories can be followed without violating actuator saturation or stability limits, and that the safety-critical predicates encoded in the \ac{STL} formula, including workspace containment, obstacle avoidance, and no-approach-from-behind, remain satisfied throughout the maneuver. In addition, comfort- and visibility-related specifications are preserved during execution, indicating that the collaboration-oriented objectives enforced at the planning stage are not lost when the trajectories are followed by the onboard controller.
		
		The Gazebo simulation environment was designed to closely replicate the MATLAB scenario, ensuring consistency between both testing frameworks. In the simulation, the designated handover location was represented as a blue region, no-fly zones (including no-approach areas and obstacles) were highlighted in red, and approach regions defined by the operator were marked in yellow. For this test, the approach direction was specified as left-handed, with a top-to-bottom preference, aligned with mission parameters defined prior to execution.
		
		The system architecture, illustrated in Figure \ref{fig:controlScheme}, integrates the \ac{STL}-based motion planner that solves the optimization problem \eqref{eq:optimizationProblemSmooth} offline to generate feasible trajectories ($\mathbf{x}^\star$, $\mathbf{u}^\star$) for the \ac{MRAV}. This planning stage is executed at the initial time $t_0$, and the resulting trajectories are then provided as reference inputs to the onboard tracking controller \cite{Afifi2022ICRA, Corsini2022IROS} for closed loop execution. The implementation assumes availability of standard onboard state estimation and human-pose perception, together with environment geometry obtained from mapping or prior models, and does not rely on specialized hardware beyond what is typically available on aerial robotic platforms.
		
		When disturbances are introduced, the robustness-aware replanner computes corrective trajectory segments that reconnect the vehicle to the nominal plan while maintaining satisfaction of the safety-critical \ac{STL} subset, demonstrating that the proposed online module can be executed in conjunction with realistic vehicle dynamics and control loops.
		
		Videos showcasing the Gazebo simulations are available at \url{https://mrs.felk.cvut.cz/stl-ergonomy-risk-analysis}, providing a comprehensive visualization of the scenarios and the system's capabilities. These simulations are not intended to provide a comprehensive evaluation of tracking accuracy or controller robustness, as they rely on a standard onboard controller provided by the TeleKyb3 framework rather than on a controller designed or tuned specifically for this study. Their purpose is instead to demonstrate the implementability of the proposed planning and replanning architecture and the feasibility of enforcing \ac{STL} specifications in a realistic physics-based environment. A systematic quantitative analysis of tracking errors under controlled disturbances and dedicated controller tuning is left for future work.
		
		%%% END SECTION ============================================================
		
		%%% START SECTION ==========================================================
		
		\section{Conclusions}
		\label{sec:conclusions}
		
		This paper introduced a novel motion planning approach for human-robot collaboration using an \ac{MRAV} equipped with a rigidly attached stick carrying a small object, with a focus on ergonomics, comfort, and efficiency for robotics co-workers. The planner leverages \ac{STL} specifications to encode diverse mission objectives, including safety, temporal constraints, and human preferences, highlighting the flexibility and expressive power of this formal specification language. The approach formulates an optimization problem to generate dynamically feasible trajectories that meet mission requirements while accounting for vehicle dynamics and physical actuation limits. 
		
		Energy efficiency is enhanced through an integrated energy-saving term, minimizing consumption and improving operational efficiency. An event-driven replanning strategy was incorporated, enabling real-time trajectory updates in response to unexpected disturbances, a critical requirement for real-world robotic missions. Additionally, a risk-aware analysis framework was developed to quantify and assess potential violations of \ac{STL} specifications under uncertainties in human pose, providing criteria to determine when re-execution of the planning problem is necessary. The planner was validated through extensive simulations conducted in MATLAB and Gazebo, demonstrating its effectiveness in achieving safe and efficient collaboration.
		
		Future work will include field experiments in realistic mock-up environments to validate the proposed framework beyond Gazebo simulations and to further assess its practical robustness. We also plan to incorporate human operator fatigue into the formulation through adaptive weighting of Boolean and temporal operators, enabling robustness margins to vary with the operator's state. In addition, systematic sensitivity analysis and automatic tuning of \ac{STL} and robustness parameters across different mission profiles and vehicle platforms will be investigated to improve adaptability and deployment readiness.
		
		%%% END SECTION ============================================================
		
		%%% START SECTION ==========================================================
		
		\section*{Declarations}
		
		% This avoids text going beyond the margins
		\begin{sloppypar}
			\noindent\textbf{Acknowledgments} This paper is an extended and substantially revised version of our prior conference publication presented at the 2023 International Conference on Unmanned Aircraft Systems (ICUAS'23) \cite{SilanoICUAS2023}. The present manuscript significantly expands the modeling, specification, and evaluation components of that work, including the integration of full nonlinear \ac{MRAV} dynamics and actuator-level constraints into the \ac{STL} formulation, the introduction of energy-aware optimization and robustness thresholds, a systematic uncertainty-aware risk analysis, and an enhanced robustness-aware replanning strategy. \\ %\url{10.1109/ICUAS54217.2022.9836083} \\
		\end{sloppypar}
		
		\noindent\textbf{Authors' contributions} \textbf{Giuseppe Silano}: Conceptualization, Methodology, Formal analysis, Software, Simulations, Writing -
		original draft. \textbf{Amr Afifi}: Conceptualization, Software, Writing – review \& editing. \textbf{Martin Saska}: Writing – review \& editing, Supervision. \textbf{Antonio Franchi}: Writing – review \& editing, Supervision. \\
		
		\noindent\textbf{Code or data availability} Data sets generated during the current study are available from the corresponding author on reasonable request. \\
		
		\noindent\textbf{Consent for publication} Informed consent was obtained from all the co-authors of this publication. \\
		
		\noindent\textbf{Competing interests} The authors declare that they have no conflict of interest. \\
		
		\noindent\textbf{Ethics approval and consent to participate} All applicable institutional and national guidelines were followed. \\
		
		% This avoids text going beyond the margins
		\begin{sloppypar}
			\noindent\textbf{Funding} This work was partially funded by the CTU grant no. SGS26/077/OHK3/1T/13, by the Czech Science Foundation (GAČR) under research project no. 26-22419S, by the European Union under the project ``Robotics and Advanced Industrial Production'' (reg. no. CZ.02.01.01/00/22 008/0004590), by the European Union's Horizon 2020 research and innovation programme AERIAL-CORE under grant agreement no. 871479, by the European Union's Horizon Europe research and innovation programme AUTOASSESS under grant agreement no. 101120732, and by the research fund for the Italian Electrical System (Ricerca di Sistema) through the decree n. 388 of November 6th, 2024.
		\end{sloppypar}
		
		%%% END SECTION ============================================================
		
		%%% START SECTION ==========================================================
		
		\bibliographystyle{spmpsci}
		\bibliography{bib}% common bib file
		
	\end{document}